\documentclass[11pt]{article}

\makeatletter
\def\input@path{{latex/}}
\makeatother

\usepackage{acl}

\usepackage{times}
\usepackage{latexsym}
\usepackage[T1]{fontenc}
\usepackage[utf8]{inputenc}
\usepackage{microtype}
\usepackage{inconsolata}

\usepackage{graphicx}
\usepackage{grffile}
\usepackage{amsmath}
\usepackage{amssymb}
\usepackage{bm}
\usepackage{booktabs}
\usepackage{multirow}
\usepackage{array}   % extended column definitions (e.g., >{\raggedright\arraybackslash})
\usepackage{tabularx} % tables that automatically fit \columnwidth
\usepackage{xcolor}
\usepackage{colortbl}    % \rowcolor in tables
\usepackage{subcaption}
\usepackage{enumitem}
\usepackage{placeins}

\usepackage[most]{tcolorbox} % framed/colored boxes
\newtcolorbox{promptcard}{
  enhanced,
  breakable,
  colback=white,
  colframe=black!70,
  boxrule=0.9pt,
  arc=3pt,
  left=10pt,right=10pt,top=8pt,bottom=8pt,
  boxsep=0pt,
  before skip=6pt,
  after skip=6pt
}
\usepackage{adjustbox}
\usepackage{float} % for [H] placement
\usepackage{dblfloatfix} % improve placement options for two-column floats (e.g., [!b] for table*)
\usepackage{xurl} % better line breaking for long paths/URLs in \url/\path

\hypersetup{hidelinks}

\usepackage{pifont}
\newcommand{\cmark}{\ding{51}}
\newcommand{\xmark}{\ding{55}}

\newcommand{\tabmain}{\ref{tab:main}}        % Table 1
\newcommand{\tabablation}{\ref{tab:ablation}} % Table 3
\newcommand{\appcases}{\ref{app:cases}}       % Appendix (multi-hop case studies)

\usepackage{CJKutf8}

\title{LEGO: Synergizing ExpertGraphRAG and ExpertCoT for Legal Reasoning}

\author{
\textbf{Qingjing Chen}\textsuperscript{1}\thanks{These authors contributed equally to this work.},
\textbf{Junkai Zhang}\textsuperscript{2}\footnotemark[1],
\textbf{Shaochun Wang}\textsuperscript{5}\thanks{Corresponding Author.},
\textbf{Jiahao Ding}\textsuperscript{3},
\textbf{Siyuan Zheng}\textsuperscript{4},\\
\textbf{Yukun Yan}\textsuperscript{2},
\textbf{Zhi Zheng}\textsuperscript{5},
\textbf{Antonino Rotolo}\textsuperscript{1}\footnotemark[2],
\textbf{Yun Liu}\textsuperscript{2},
\textbf{Weixing Shen}\textsuperscript{2}
\\
\textsuperscript{1}Alma AI, University of Bologna, Italy\\
\textsuperscript{2}Tsinghua University, China\\
\textsuperscript{3}Xiamen University, China\\
\textsuperscript{4}Shanghai Jiao Tong University, China\\
\textsuperscript{5}Modelbest Inc.\\
\texttt{\{qingjing.chen2,antonino.rotolo\}@unibo.it}\\
\texttt{jj-zhang25@mails.tsinghua.edu.cn}\\
\texttt{wangshaochun@modelbest.cn}
}

\begin{document}
\begin{CJK*}{UTF8}{gbsn}
\maketitle

\begin{abstract}
Large language models are increasingly applied to high-risk domains such as law, yet complex legal reasoning remains limited by two structural challenges. First, existing RAG and GraphRAG methods emphasize lexical or semantic similarity while overlooking normative relations among legal provisions. Second, vanilla Chain-of-Thought prompting may generate plausible rationales without enforcing the normative structure of legal reasoning. To deal with the bottleneck of pipelines in the legal reasoning domain, we propose \textbf{LEGO}, a dual-module framework that synergizes \textbf{L}egal \textbf{E}xpert \textbf{G}raphRAG (ExpertGraphRAG) and \textbf{E}xpert \textbf{C}hain-\textbf{o}f-\textbf{T}hought (ExpertCoT) for complex legal reasoning. ExpertGraphRAG uses an expert-annotated Civil Code graph encoding these normative relations with a greedy normative-coverage retrieval algorithm to dynamically extract instance-specific provision subgraphs, while ExpertCoT organizes the retrieved provisions and case facts into structured Provision–Fact–Conclusion reasoning. With a Qwen3-8B backbone, LEGO achieves 40.53\% exact-match accuracy on LawExamQA\_Civil, outperforming the evaluated RAG and CoT baselines and performing comparably to the evaluated larger models, while remaining robust on multi-hop questions. It also achieves the best results among the evaluated baselines on the open-ended benchmarks. Ablation studies confirm the individual and complementary contributions of both modules, demonstrating LEGO’s effectiveness in improving LLMs’ complex legal reasoning ability. Code and dataset can be found in the link: https://github.com/BLK-WHT/LEGO

\end{abstract}

% Place the paper overview figure immediately after the abstract so it lands at the top of page 2 in two-column ACL layout.
\begin{figure*}[!t]
\centering
\includegraphics[width=0.95\textwidth]{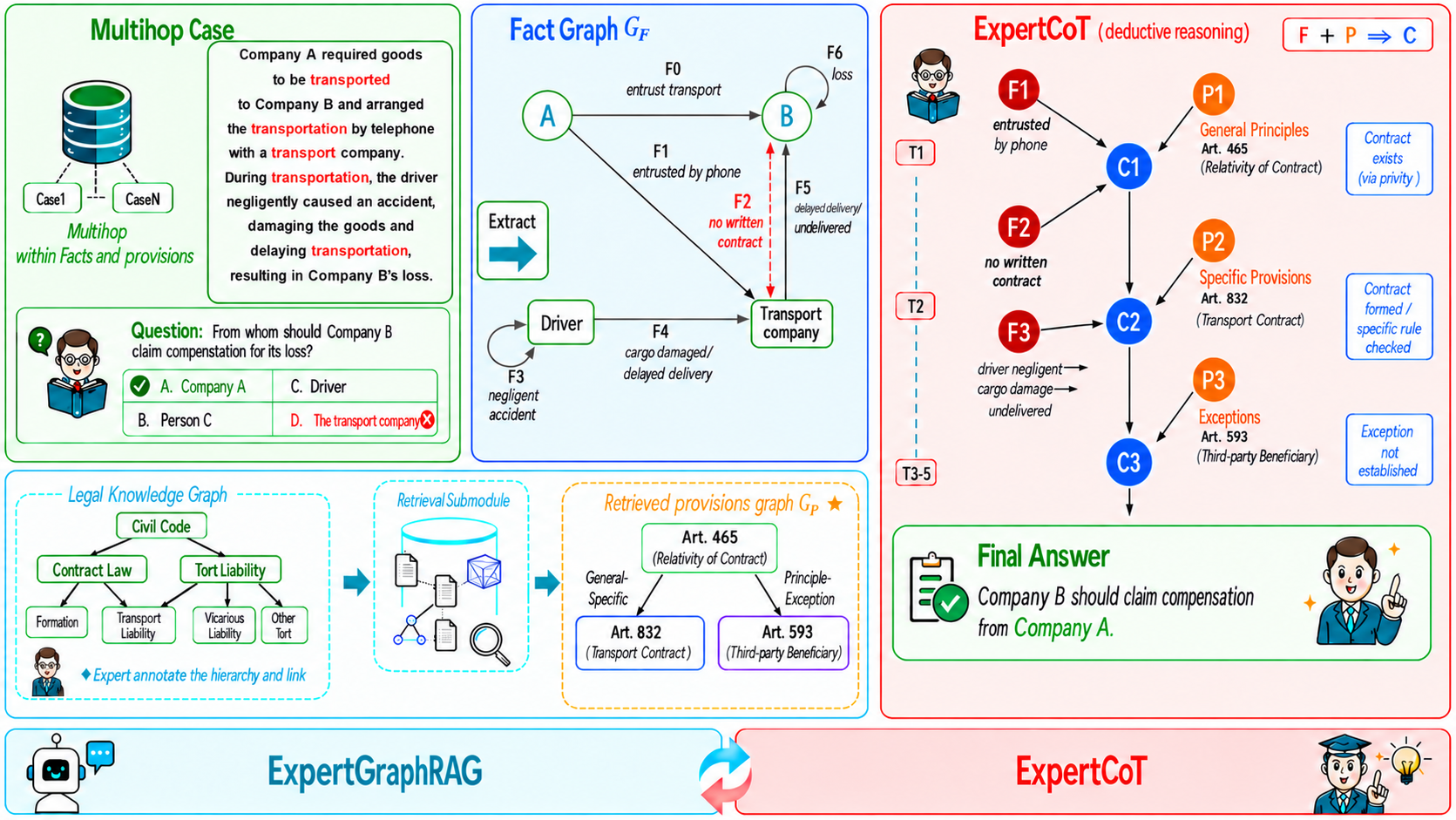}
\caption{LEGO’s architecture. ExpertGraphRAG uses the fact-side instance representation, implemented as a natural-language query containing the case facts, question, and options to retrieve an instance-specific provision subgraph from the static Expert Provision Graph. ExpertCoT then composes the retrieved provisions and case facts into a structured P–F–C analysis. “Fact Graph” and “Conclusion Graph” in the figure denote conceptual representations rather than separately materialized graph objects.
}
\label{fig:pipeline}
\end{figure*}

\section{Introduction}

The application of LLMs to high-risk professional domains has accelerated
rapidly, including law, where dedicated benchmarks now measure legal language
understanding and reasoning
\cite{chalkidis2022lexglue,guha2023legalbench,li2025lexrag,shi-etal-2026-plawbench}. Yet reliable legal reasoning
remains challenging: legal LLM reasoning is not robust and can be easily misled by seemingly relevant adversarial knowledge~\cite{hu2025j}; models that achieve strong performance on training statutes often fail to generalize to unseen statutes~\cite{blair2023can}; they can reach the right answer
without producing a sound reasoning path~\cite{kang-etal-2023-chatgpt}, and
they struggle to recover the provisions a question actually depends on once
several are involved~\cite{koblex2025}. These problems involve two main
pipelines: Retrieval-Augmented Generation (RAG) grounds generation in external
evidence~\cite{NEURIPS2020_6b493230}, while Chain-of-Thought (CoT) prompting has the
model follow intermediate reasoning steps~\cite{wei2022chain}. Although their strengths and
weaknesses have begun to be examined jointly in the general domain
\cite{li2025cotrag} and in medicine~\cite{ge_expert-guided_2025}, legal NLP applications have
not yet combined them in a way that addresses the limitations of each: both are
largely evaluated and improved in isolation~\cite{liu2025jurex,chen-etal-2026-legalgraphrag}.

The first bottleneck is retrieval. Standard RAG and many GraphRAG systems~\cite{microsoftgraphrag2024,Zhang2025ASO,linearrag2026,logicrag2026,hipporag2025,guo-etal-2025-lightrag} %Existing 
%methods 
still rely heavily on lexical or embedding similarity, even with induced graph neighborhoods. Yet legal knowledge is semantically sparse: textually distant provisions may be tightly linked through prerequisites, exceptions, general–special relations, or legal effects, while textually similar ones may distract because they belong to different constituent elements~\cite{blair2023can,koblex2025}. The complex case of Figure~\ref{fig:pipeline} shows this issue: repeated transport-related terms may lead similarity-based models to select the transport company (Option D). Yet the Fact Graph shows that B has no contract with the transport company. Correspondingly, the Provision Graph must capture the general–specific relation between contractual privity under Art. 465 and carrier liability under Art. 832, as well as Art. 593’s exclusion of third-party-caused breach as a basis for bypassing contractual privity. Lexical similarity alone captures neither these factual relations nor the normative links among provisions.

The second bottleneck is reasoning. Vanilla CoT can produce plausible intermediate text, but it does not guarantee that the model follows the normative order required by a professional domain~\cite{wei2022chain,yao2023tree}. CoT is extremely fragile and lacks robustness in high-risk domains such
as law~\cite{yu2025benchmarking,kang-etal-2023-chatgpt}. The reasoning trace can drift by skipping an exception, treating an unfulfilled prerequisite as satisfied, or affirming the consequent from a legal effect back to its condition~\cite{servantez-etal-2024-chain}. As faithfulness and robustness studies of reasoning in RAG also note, retrieval helps only if the retrieved sources are authoritative and structurally relevant; otherwise, each generated step may amplify noise from the previous one~\cite{brink2026}. Research highlights the need to constrain and regulate the models' CoT using knowledge injection or expert-guided legal structures~\cite{liu2025jurex}.

To address these bottlenecks, we propose LEGO, a dual-module framework that
injects expert legal knowledge into both stages of the pipeline: an
expert-annotated knowledge structure into retrieval (ExpertGraphRAG), and the
expert mode of legal reasoning into generation (ExpertCoT). ExpertGraphRAG
retrieves over an expert-annotated Civil Code graph, in which doctrinal rule
cards written by legal experts encode the normative relations among
provisions. Rather than ranking articles by semantic similarity alone, its
Normative Coverage Greedy (NCG) component grows a provision set by marginal
legal utility, so that each article it adds covers a legal issue the case
raises but the current set does not. This yields a compact, instance-specific
provision subgraph $G^{(i)}_P$ instead of a list of lexically similar
articles. ExpertCoT then organizes the retrieved provisions and case facts
into structured Provision--Fact--Conclusion reasoning, with the controlling
rules as the major premise, the case facts as the minor premise, and the
per-option judgments as the conclusion. With a Qwen3-8B backbone, LEGO reaches
40.53\% exact-match accuracy on LawExamQA\_Civil, outperforming the evaluated
RAG and CoT baselines and performing comparably to far larger models such as
GPT-5 and DeepSeek-V3 671B, while remaining robust on multi-hop complex reasoning
involving multiple provisions. It also achieves the best results among the
evaluated baselines on two open-ended benchmarks, and ablations confirm the
individual and complementary contributions of the two modules. The structured
syllogistic output further makes the rationale linking provisions, facts, and
conclusions easier to inspect and audit.

\section{Related Work}
\label{sec:related}

\subsection{GraphRAG and Legal-Domain Retrieval}

Flat-vector RAG can struggle when evidence is distributed across passages because similarity ranking alone does not guarantee retrieval of complete reasoning chains~\cite{han-etal-2026-rag-vs-graphrag}, a challenge isolated by multi-hop QA benchmarks such as HotpotQA, MuSiQue, and 2WikiMultiHopQA~\cite{yang2018hotpotqa,trivedi2022musique,ho2020constructing}. GraphRAG instead structures corpus information as graphs and performs graph-aware retrieval or aggregation~\cite{microsoftgraphrag2024,Zhang2025ASO}. Recent work explores efficient or ontology-guided graph construction~\cite{linearrag2026,wang2026omdgraphragenhancinggraphragontologyguided}, inference-time reasoning structures without pre-built graphs~\cite{logicrag2026}, graph-free triplet retrieval~\cite{gong-etal-2026-beyond}, dependency-aware reranking~\cite{pankrag2025}, memory-based HippoRAG~2~\cite{hipporag2025}, Tree Retrieval RAPTOR~\cite{raptor2024}, subgraph-retrieval-based G-Retriever~\cite{gretriever2024}, and LightRAG incorporating graph structures~\cite{guo-etal-2025-lightrag}. LegalGraphRAG further organizes legal knowledge into Fact, Ontology, and Rule subgraphs~\citep{chen-etal-2026-legalgraphrag}. However, neither these general-purpose baselines nor LegalGraphRAG evaluates retrieval on complex or multi-hop legal reasoning tasks involving normative relations among provisions. Besides, retrieving the right provisions does not ensure reasoning to the correct final answer, as legal conclusions often depend on reasoning over the relational interactions among multiple provisions and performing deductive inference from facts to the applicable provisions.

\subsection{Chain-of-Thought and Legal Reasoning}

Chain-of-Thought~\cite{wei2022chain} and Tree-of-Thought~\cite{yao2023tree} prompting externalise intermediate steps but offer no guarantee of normative correctness and reliability in specific domains. Verifier-driven reasoning, popularised by HuatuoGPT-o1~\cite{huatuogpto12025} for medicine, audits each step with a True/False check and triggers backtracking, path-exploration, or self-correction on failure. Legal prompting studies have directly evaluated IRAC-based reasoning on the COLIEE entailment task~\cite{yu-etal-2023-exploring} and on expert-annotated legal scenarios~\cite{kang-etal-2023-chatgpt}. More recently, MSLR introduced expert-derived IRAC traces for evaluating multi-step legal reasoning~\cite{yu2025benchmarking}. IRAC-style structured legal reasoning has also been explored on top of ChatLaw~\cite{Cui_2026}, Lawformer~\cite{xiao2021lawformer}, and LexGLUE~\cite{chalkidis2022lexglue}. However, these studies lack grounding in reliable retrieved evidence, leaving CoT vulnerable to instability and hallucination. Moreover, existing legal CoT, such as IRAC, provides procedural reasoning guidance but does not model the complex nature of normative reasoning, which requires syllogistic and deductive inference over multiple interrelated legal norms and case facts.

Recent RAG--CoT systems interleave retrieval with reasoning~\cite{trivedi2023ircot}, revise thought steps with retrieved evidence~\cite{wang2024rat}, or use knowledge graphs to constrain CoT generation~\cite{li2025cotrag}; expert-guided medical QA further shows that domain attributes can constrain both retrieval and CoT in high-risk settings~\cite{ge_expert-guided_2025}. However, legal reasoning is uniquely challenging because it requires iterative normative reasoning between facts and legal norms: the reasoner must identify applicable legal provisions, infer normative relations among them, and continuously refine the interpretation of both facts and norms until a justified conclusion is reached.

\section{Methodology}
\label{sec:methodology}
We instantiate LEGO as a pipeline following legal syllogistic reasoning~\cite{patzig2013aristotle}, which decomposes legal inference into the major premise (normative Provision), minor premise (Fact), and conclusion (Figure~\ref{fig:pipeline}).
%We thus instantiate LEGO as a three-step graph-conditioned pipeline following legal syllogistic reasoning, which decomposes legal inference into the major premise, minor premise, and conclusion. 
For each query $q_i$,
comprising the case facts and question, with answer options $O_i$,
LEGO forms a fact-side query $G_F^{(i)}$. ExpertGraphRAG uses this
query to retrieve a set of relevant provisions $P_i$ from the global
Provision Graph $G_P$. ExpertCoT then combines the original query,
answer options, and retrieved provision texts to generate a structured
P--F--C analysis $G_C^{(i)}$ and the final answer $\hat{A}_i$:

\begin{equation}
\begin{aligned}
G_F^{(i)} &:= \operatorname{Serialize}(q_i, O_i), \\
P_i &:= f_{\mathrm{RAG}}\!\left(G_F^{(i)}, G_P\right), \\
\left(\hat{A}_i, G_C^{(i)}\right)
&= f_{\mathrm{CoT}}\!\left(q_i, O_i, P_i\right).
\end{aligned}
\label{eq:lego-pipeline}
\end{equation}

Here, $G_F^{(i)}$ is implemented as the natural-language retrieval
query, $P_i$ denotes the retrieved provision texts, and $G_C^{(i)}$
denotes the generated P--F--C response rather than a separately
materialized graph. LEGO therefore combines graph-structured provision retrieval with structured syllogistic generation. Unlike prior expert-conditioned QA that uses flat attributes such as subject-area tags~\cite{ge_expert-guided_2025}, LEGO represents expert knowledge as a relational legal subgraph and uses CoT as a constrained syllogistic reasoning process over that graph.

\subsection{ExpertGraphRAG}
\label{sec:expertrag}
LEGO retrieves Civil Code articles from an offline \emph{Provision Graph}, rather than from free-form chunks. Let $\mathcal{R}$ be expert-defined legal rule units and $\mathcal{A}$ be Civil Code articles:
\begin{equation}
\begin{aligned}
G_P &= (V_P,E_P;\mathbf{B}),\\
V_P &= \mathcal{R}\cup\mathcal{A},\\
E_P &= E_{\mathcal{RA}}\cup E_{\mathrm{norm}} .
\end{aligned}
\label{eq:provision_graph}
\end{equation}
Here $E_{\mathcal{RA}}$ is the bipartite rule--article layer, $E_{\mathrm{norm}}$ stores expert provision--provision relations, and $B_{r,a}$ is the support weight of rule unit $r$ for article $a$ (see the Civil-law ExpertGraph description in Section~\ref{sec:civil_law_expertgraph} and Appendix~\ref{sec:appendix_graph}). For input $x_i$, let $\mathcal{R}_i$ be the top-8 routed rules ranked by $\mathrm{sim}(x_i,r)$, and let $w_r$ be the corresponding normalized routing similarity over $\mathcal{R}_i$. We define \(d_i(a)=\mathrm{sim}(x_i,a)\), \(g_i(a)=\max_{r\in\mathcal{R}_i} w_r\,B_{r,a}\), and \(\Gamma_i(a)=\{r\in\mathcal{R}_i: B_{r,a}>0\}\).

\paragraph{Normative Coverage Greedy.}
We use \emph{Normative Coverage Greedy} (NCG) to select articles. Given a partial set \(S\), NCG scores a candidate article \(a\) by:
\begin{equation}
\begin{aligned}
\Delta_i(a\mid S)
&=d_i(a)+\lambda_{\mathrm{align}}g_i(a)\\
&\quad+\lambda_{\mathrm{cov}}n_i(a,S)-\lambda_{\mathrm{red}}o_i(a,S),
\end{aligned}
\label{eq:retrieval_gain}
\end{equation}
where \(\Gamma_i(S)=\bigcup_{a\in S}\Gamma_i(a)\), \(n_i(a,S)=|\Gamma_i(a)\setminus\Gamma_i(S)|\) is newly covered legal-rule evidence, and \(o_i(a,S)=|\Gamma_i(a)\cap\Gamma_i(S)|\) is already-covered evidence. We restrict candidates to $\mathcal{C}_i$, the union of the top-50 dense-retrieved articles and the top-30 graph-supported articles. The greedy update is:
\begin{equation}
\begin{aligned}
a_t
&=
\operatorname*{arg\,max}_{a\in\mathcal{C}_i\setminus S_{t-1}}
\Delta_i(a\mid S_{t-1}),\\
S_t&=S_{t-1}\cup\{a_t\},\quad S_i^\ast=S_K .
\end{aligned}
\label{eq:submodular_retrieval}
\end{equation}
The first three terms are the marginal gain of a relevance-weighted coverage objective over expert rule units. This is the same monotone submodular family underlying maximum coverage and facility-location retrieval, for which greedy selection is the standard approximation strategy~\cite{nemhauser1978,lin2011submodular}. The final term is a lightweight redundancy regularizer: it discourages repeatedly selecting provisions that explain the same legal issue, but we do not use it to claim a new approximation bound. In effect, NCG optimizes the \emph{marginal legal utility} of each provision: it keeps provisions on topic, expands normative coverage, and yields a compact provision subgraph that serves as the legal major-premise context for downstream ExpertCoT. The weights $\lambda_{\mathrm{align}},\lambda_{\mathrm{cov}},\lambda_{\mathrm{red}}$ and the article budget $K$ used in our experiments are listed in Appendix~\ref{sec:appendix_hyperparams}.

The selected set induces the instance-level provision subgraph:
\begin{equation}
\begin{aligned}
G_P^{(i)} &= (V_P^{(i)},E_P^{(i)}),\\
V_P^{(i)} &= S_i^\ast\cup\mathcal{R}_i^\ast,\\
E_P^{(i)} &= E_P\cap\bigl(V_P^{(i)}\times V_P^{(i)}\bigr).
\end{aligned}
\end{equation}
where $\mathcal{R}_i^\ast=\{r\in\mathcal{R}:\max_{a\in S_i^\ast}B_{r,a}>0\}$.

\subsection{ExpertCoT}
\label{sec:syllogistic_cot}

ExpertCoT implements $f_{\mathrm{CoT}}$ in Eq.~\ref{eq:lego-pipeline}. It
receives the query $q_i$, the answer options $O_i$, and the texts $P_i$ of
the articles selected by ExpertGraphRAG (Sec.~\ref{sec:expertrag}); no
graph structure is passed to the model. In a single generation pass, the
model writes the P--F--C analysis $G_C^{(i)}$ followed by the final answer
$\hat{A}_i$:
\begin{itemize}[leftmargin=*,topsep=2pt,itemsep=1pt]
  \item \textbf{P (major premise).} The controlling rules, drawn only from
        $P_i$, together with the relations among them that govern the
        outcome (general vs.\ special rule, proviso or exception, limiting
        condition, remedy).
  \item \textbf{F (minor premise).} The legally relevant facts in $q_i$,
        stated without drawing legal conclusions.
  \item \textbf{C (conclusion).} A check of the provision relations stated
        in P, then a judgment on each option in $O_i$ obtained by applying
        those rules to the facts, and finally the option set $\hat{A}_i$
        required by the question.
\end{itemize}
The P and F sections are written once for the whole instance, and the C
section judges every option against them, so an option-level judgment may
rest on several provisions. The relations among provisions are therefore
stated by the model in its P section rather than supplied as input.

To reduce reasoning errors and unsupported assertions, ExpertCoT augments the P–F–C reasoning framework with eleven explicit instructions. The model is required to first identify the legal object under examination and the question requirement. It must distinguish different legal effects, including contract validity, transfer of real rights, opposability, liability, damages, and priority rights or defenses, without incorrectly extending one legal effect to another. In addition, limiting clauses, such as “shall not be held invalid solely because ...” or “does not affect the validity of ...”, must be interpreted as restrictions on the scope of application rather than as grounds for invalidity or the extinction of rights. The model also verifies the consistency of each option’s subject, legal predicate, required elements, and asserted legal effect, thereby avoiding over-selection and correctly handling questions that ask for either correct or incorrect statements. These instructions were iteratively refined through expert-in-the-loop error analysis. The full prompt and an end-to-end example are reproduced in Appendix~\ref{sec:appendix_prompt_card}.

\section{Experiments}
\label{sec:experiments}

\subsection{Setup}
\label{sec:setup}
\paragraph{Datasets.} We choose the civil-law domain because it contains rich legal knowledge and complex reasoning patterns involving provision dependencies, exceptions, and diverse legal effects, making accurate and reliable legal reasoning particularly challenging. We evaluate LEGO on three open datasets in the civil‑law domain, ensuring consistency with the expert‑constructed civil‑law knowledge graph used in our pipeline.  (1) \textbf{LawExamQA\_Civil}, a subset of the Chinese questions from the National Judicial Examination of China (NJEC), where we categorize questions by the number of provisions referenced in the official rationale. This hop count serves as a proxy for the complexity of multi‑hop legal reasoning. (2) \textbf{LexRAG\_Civil}~\cite{li2025lexrag}, a civil‑law subset of \textit{LexRAG} used to assess multi‑turn legal consultation performance. \textbf{LexRAG\_Civil} evaluates open-ended, multi-turn legal consultation. Models generate free-form responses to successive user queries rather than selecting from predefined answer options. (3)~\textbf{PLawBench\_Civil}~\cite{shi-etal-2026-plawbench}, a rubric‑based benchmark for evaluating LLMs in real‑world legal practice. We use the \textit{Practical Case Analysis} section, which consists of complex civil‑law cases annotated with reasoning rubrics, enabling assessment of legal reasoning ability in realistic scenarios. Our evaluation is not limited to multiple-choice question answering. Only LawExamQA\_Civil uses a multiple-choice format; LexRAG\_Civil evaluates free-form responses in multi-turn legal consultations, and PLawBench\_Civil evaluates generated practical legal analyses using task-specific rubrics.
All of them are evaluated on complex legal‑reasoning tasks involving multiple provisions and long case texts. The data details can be found in Appendix~\ref{sec:appendix_datacard}.

\textbf{Civil Law ExpertGraph.}\label{sec:civil_law_expertgraph} Our expert-annotated Civil Law graph was built over 7 months by trained legal experts; it serves solely to enhance RAG signal over the Civil Code corpus (Appendix~\ref{sec:appendix_graph}). It encodes the PRC Civil Code as
318 concepts over 409 rule cards, linked to 1{,}211 of its 1{,}260
articles by 4{,}152 support edges. A card states one doctrinal rule — its issue frames, positive and negative conditions, legal effects and exceptions — and every field is embedded, so a query routes to concepts rather than to article text. Edges carry the support strength of a concept for an article, learned by a six-view alignment and smoothed by entropic optimal transport. The six views are semantic, field-level, structural, hierarchical, and two reference views (an LLM view and a rule/regex view). A concept may carry several cards; at retrieval it competes on its best-matching one.

\paragraph{Language model Baselines} 
(1) Closed‑source and large general LMs. We choose GPT‑5~\cite{openai2025gpt5systemcard} (closed-source) and DeepSeek‑V3 671B~\cite{deepseek2024v3} as large general‑purpose LMs to test the performance ceiling achievable through purely parametric reasoning without external retrieval. (2) Open Larger LMs. We choose Qwen3-30B-A3B~\cite{qwen3techreport2025} and GLM-4.7-Flash~\cite{zai2026glm47flash} as Open Larger LMs, which offer a strong balance between computational cost and performance at the 30B scale. (3) Small Open-source LMs. We use Qwen3‑8B~\cite{qwen3techreport2025} and GLM‑4‑9B‑Chat~\cite{zai2024glm49bchat} as base models whose 8B–9B parameter scale matches our RAG generators for direct comparison. (4) Legal-domain LMs. We select DISC‑LawLLM (7B)~\cite{10.1007/978-981-97-5569-1_19} and LegalOne (8B)~\cite{Li2026LegalOneAF}, two legal-domain fine-tuned open‑source models tailored for legal reasoning.

\paragraph{RAG baselines} 
We compare LEGO with several representative RAG architectures: flat top-k dense retrieval with Qwen3-Embedding-8B; auto-induced semantic graph (LightRAG~\cite{guo-etal-2025-lightrag}); hierarchical summary tree (RAPTOR~\cite{raptor2024}); Steiner-tree graph retrieval (G-Retriever~\cite{gretriever2024}); long-range memory / associative graph (HippoRAG~2~\cite{hipporag2025}). All RAG baselines use the same Qwen3-8B backbone and the same Civil Code retrieval corpus as LEGO, so accuracy differences are attributable to retrieval / reasoning design rather than to scale or external knowledge.

The experiments use the method-specific configurations documented in Appendix \ref{sec:appendix_baseline} because heterogeneous retrievers define and expand retrieval units differently. These units include chunks, statute articles, PCST seed nodes, tree nodes, and entity–relation candidates, followed by method-specific expansion and serialization procedures. Therefore, no single measure, such as top-$k$, article count, node count, or token count can fully equalize their retrieval budgets. We accordingly compare complete pipelines under a shared reader, corpus, evaluation set, and decoding configuration, rather than retrieval algorithms using numerically identical but semantically different retrieval units. Details for each baseline are provided in Appendix~\ref{sec:appendix_baseline}.

\textbf{CoT baselines}
Zero‑shot‑CoT~\cite{kojima2022large} induces models to generate reasoning steps by adding “Let’s think step by step,” while IRAC‑CoT is a typical legal reasoning method with an Issue–Rule–Application–Conclusion loop demonstrated by MSLR~\cite{yu2025benchmarking}.

\paragraph{Evaluation metrics.}
For LawExamQA\_Civil, we report exact-match accuracy (Acc.)
and set-level F1 over predicted and gold option sets.
For each item $i$, accuracy is 1 if the predicted set $P_i$
equals the gold set $G_i$, and 0 otherwise; set-level F1 is
$\mathrm{F1}_i = 2|P_i \cap G_i|/(|P_i| + |G_i|)$.
Both metrics are averaged over items rather than classes.
Empty predictions or outputs from which no answer can be
parsed receive zero for both metrics. For LexRAG\_Civil, we report Factuality, Satisfaction,
Clarity, Coherence, Completeness, and Overall scores.
For PLawBench\_Civil, we report Reasoning and Overall scores; judge models and scoring details are given in Appendix~\ref{sec:appendix_datacard}.

For provision retrieval, we report Recall@$k$ at
$k \in \{8,10,20\}$ to measure coverage of gold provisions
cited in the official rationales. We also report the
Gap Closing Rate (GCR), defined in Sec.~\ref{sec:retrieval_recall}, to quantify
the fraction of the accuracy gap between the zero-shot
and gold-article settings recovered by each method.

We also compare LEGO with all 13 systems in Table~\tabmain{} using two-sided exact McNemar tests on paired per-item predictions, percentile 95\% bootstrap confidence intervals from 10,000 item-level resamples, and Holm correction across the complete comparison family (Appendix~\ref{app:stats}).

\begin{table*}[t]
\centering
\small
\renewcommand{\arraystretch}{1.2}
\begin{tabular*}{\textwidth}{@{\extracolsep{\fill}} l l c c c c c @{}}
\toprule
\multirow{2}{*}{\textbf{Method}} & \multirow{2}{*}{\textbf{Size}} & \multicolumn{5}{c}{\textbf{LawExamQA\_Civil Accuracy (\%) by Hop Count}} \\
\cmidrule(lr){3-7} 
& & \textbf{Overall} & \textbf{1-hop} & \textbf{2-hop} & \textbf{3-hop} & \textbf{$\ge$4-hop} \\
\midrule
\multicolumn{7}{@{}l}{\textit{Closed-source \& Large General LMs}} \\
GPT-5 & API & 37.48 & \underline{41.23} & 37.67 & 29.81 & 29.03 \\
DeepSeek-V3 & 671B & \underline{39.97} & \underline{41.23} & \underline{39.53} & \textbf{41.35} & \underline{32.26} \\
\addlinespace 
\multicolumn{7}{@{}l}{\textit{Open Larger LMs}} \\
Qwen3-30B-A3B & 30B & 36.93 & 39.18 & 38.14 & 30.77 & 30.65 \\
GLM-4.7-Flash & 30B & 30.98 & 36.26 & 27.44 & 24.04 & 25.81 \\
\addlinespace
\multicolumn{7}{@{}l}{\textit{Small Open-source LMs}} \\
Qwen3-8B & 8B & 25.45 & 25.44 & 27.44 & 26.92 & 16.13 \\
GLM-4-9B-chat & 9B & 29.18 & 30.41 & 27.44 & 30.77 & 25.81 \\
\addlinespace
\multicolumn{7}{@{}l}{\textit{Legal Domain LMs}} \\
DISC-LawLLM & 7B & 18.12 & 17.84 & 19.53 & 15.38 & 19.35 \\
LegalOne &  8B & 30.29 & 32.16 & 31.63 & 20.19 & \underline{32.26} \\
\addlinespace
\multicolumn{7}{@{}l}{\textit{RAG Baselines }} \\
Naive RAG & 8B & 28.91 & 30.41 & 29.30 & 26.92 & 22.58 \\
HippoRAG 2 & 8B & 29.88 & 30.70 & 29.30 & 29.81 & 27.42 \\
RAPTOR & 8B & 29.32 & 28.95 & 31.16 & 30.77 & 22.58 \\
G-Retriever & 8B & 31.67 & 31.58 & 35.81 & 26.92 & 25.81 \\
LightRAG & 8B & 28.77 & 28.65 & 30.70 & 26.92 & 25.81 \\
\midrule
\textbf{LEGO(ours)} & \textbf{8B} & \textbf{40.53} & \textbf{41.81} & \textbf{40.47} & \underline{37.50} & \textbf{38.71} \\
\bottomrule
\end{tabular*}
\caption{\textbf{Main results stratified by multi-hop reasoning complexity on LawExamQA\_Civil.} Performance is measured by exact-match accuracy (\%). \textbf{Bold} indicates the absolute best performance across all evaluated models, while \underline{underline} indicates the second-best. By integrating ExpertGraphRAG and ExpertCoT, the 8B-parameter LEGO achieves the highest observed overall accuracy and exhibits remarkable robustness in deep reasoning chains ($\ge$4-hop).}
\label{tab:main_results}
\label{tab:main} % alias for appendix_statistical_validation.tex
\end{table*}

\begin{figure}[t]
\centering
\includegraphics[width=\columnwidth]{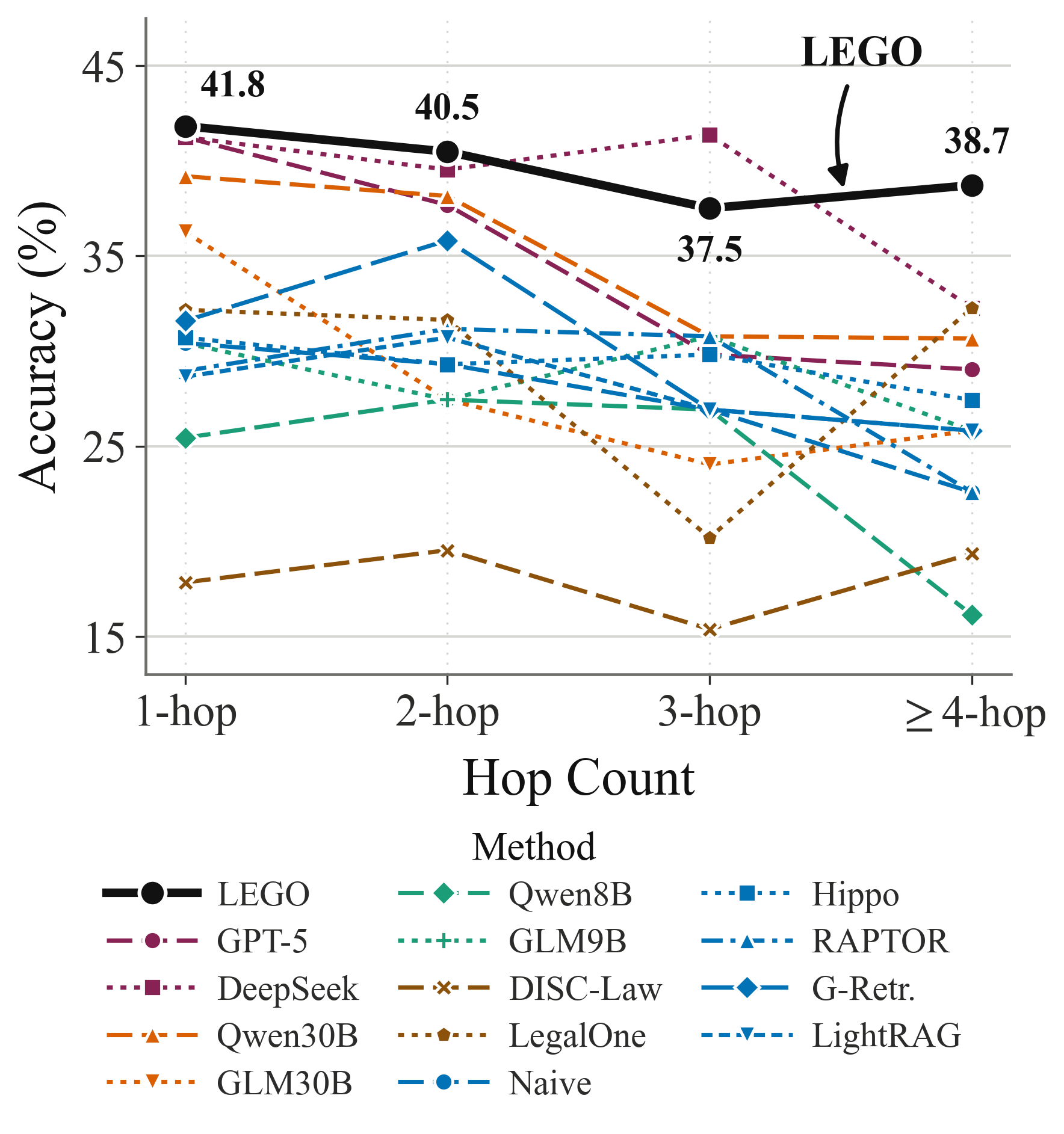}
\caption{\textbf{Multi-hop performance trend.} Accuracy breakdown across hop counts on LawExamQA\_Civil.}
\label{fig:multihop_trend}
\end{figure}

\subsection{Main Results }
Table~\ref{tab:main_results} reports the main results on LawExamQA\_Civil. LEGO achieves the best overall exact-match accuracy among the evaluated systems, reaching 40.53\% with an 8B backbone. This point estimate is numerically higher than those of DeepSeek-V3 (39.97\%) and GPT-5 (37.48\%), although paired tests do not establish statistically significant superiority over these larger models. By contrast, LEGO significantly outperforms all five evaluated same-backbone RAG systems after Holm correction, including G-Retriever (31.67\%; $\Delta=8.86$ pp, adjusted $p=4.4\times10^{-6}$). The results therefore suggest that external expert structure can compensate for parameter scale, while statistically supported superiority is limited to comparisons under the shared Qwen3-8B backbone and Civil Code corpus. Complete per-comparison statistics are reported in Appendix~\ref{app:stats}.

\paragraph{Hop-robustness.}As shown in Figure \ref{fig:multihop_trend}, LEGO’s advantage is most pronounced on deeper reasoning cases. On ≥4-hop questions, LEGO reaches 38.71\%, outperforming the second-best result of 32.26\% by 6.45 points. It also shows stronger hop-resilience: accuracy remains relatively stable across 1/2/3/4+-hop items (41.8 → 40.5 → 37.5 → 38.7), while GPT-5 drops by 12.2 points from 1-hop to 4+-hop cases; all five RAG baselines decline from 3-hop to ≥4-hop questions (by 1.1–8.2 points), whereas LEGO remains stable and even rises slightly. To better understand this robustness, we analyze items correctly
answered by LEGO but missed by all 13 baselines. The qualitative
analysis highlights four reasoning patterns: resolving general
and special rules, selecting among closely related liability
regimes, checking constitutive elements before recognizing a
claim, and identifying the relevant right holder in multi-party
settings. These patterns align with the priority, prerequisite,
exception, and entitlement relations represented in the Civil
Law ExpertGraph, suggesting that  legal knowledge structure helps
LEGO identify and apply the controlling provisions.
 Appendix~\appcases{} provides detailed analyses of representative cases.

\textbf{LexRAG\_Civil and PLawBench\_Civil.} As shown in Table~\ref{tab:cross_benchmark}, LEGO outperforms all RAG baselines on both benchmarks. On LexRAG\_Civil, LEGO achieves the best score in every evaluated dimension, including factuality (4.486), satisfaction (4.600), clarity (6.879), coherence (5.700), completeness (5.093), and overall quality (5.164). Compared with the strongest baseline, HippoRAG~2, LEGO improves the overall score by 0.335 points and completeness by 0.407 points. On PLawBench\_Civil, LEGO obtains the highest reasoning score (47.40) and overall score (59.68), suggesting that its expert-structured retrieval and reasoning framework benefits long-form legal reasoning. These results
demonstrate that LEGO improves not only subjective answer quality, but also
the robustness and practical usability of legal reasoning in multi-turn and
long-form civil-law QA. Appendix~\ref{app:lexrag_cases} and
Appendix~\ref{app:plawbench_cases} expand these two rows into case-level
analyses of multi-turn consultations and practical case analyses,
respectively.

% Keep Section 4.2 text together by preventing later floats (e.g., Fig. 3) from
% drifting upward and splitting the paragraph.
\FloatBarrier

% --- Table 2 (wide float; will appear at the top of the next available page) ---
\begin{table*}[!t]
\centering
\scriptsize
\setlength{\tabcolsep}{4.8pt}
\renewcommand{\arraystretch}{1.05}

\begin{tabular}{@{}llrrrrrr@{}}
\toprule
\textbf{Benchmark} & \textbf{Metric}
& \textbf{Naive RAG}
& \textbf{RAPTOR}
& \textbf{G-Retriever}
& \textbf{HippoRAG 2}
& \textbf{LightRAG}
& \textbf{LEGO} \\
\midrule
\multirow{6}{*}{LexRAG\_Civil}
& Factuality
& 3.707 & 3.564 & 3.829 & \underline{4.329} & 4.121 & \textbf{4.486} \\
& Satisfaction
& 3.829 & 3.814 & 3.986 & \underline{4.300} & 4.043 & \textbf{4.600} \\
& Clarity
& 6.029 & 6.029 & 6.136 & \underline{6.586} & 6.286 & \textbf{6.879} \\
& Coherence
& 4.814 & 4.743 & 4.921 & \underline{5.329} & 5.257 & \textbf{5.700} \\
& Completeness
& 4.193 & 4.129 & 4.307 & \underline{4.686} & 4.279 & \textbf{5.093} \\
& Overall
& 4.314 & 4.293 & 4.450 & \underline{4.829} & 4.543 & \textbf{5.164} \\
\midrule
\multirow{2}{*}{PLawBench\_Civil}
& Reasoning
& \underline{46.50} & 44.70 & 43.00 & 45.30 & 43.53 & \textbf{47.40} \\
& Overall
& \underline{59.05} & 58.02 & 57.22 & 57.93 & 58.43 & \textbf{59.68} \\
\bottomrule
\end{tabular}

\caption{\textbf{Cross-benchmark evaluation on LexRAG\_Civil and PLawBench\_Civil.}}
\label{tab:cross_benchmark}
\end{table*}

% --- Table 4 (wide float; will appear at the top of the next available page) ---
\begingroup
\setcounter{table}{3}
\begin{table*}[!t]
\centering
\scriptsize
\begin{adjustbox}{max width=\textwidth}
\begin{tabular}{lcccccc}
\toprule
\multirow{2}{*}{\textbf{Method}} & \multicolumn{3}{c}{\textbf{Context Coverage} (\%)} & \multicolumn{2}{c}{\textbf{QA}} & \textbf{GCR} \\
\cmidrule(lr){2-4}\cmidrule(lr){5-6}
& Recall@8 & Recall@10 & Recall@20 & Acc (\%) & F1 & \\
\midrule
Naive RAG                         & 27.99& 34.70 &50.52 &28.91 &52.16 &0.61 \\
BM25 (CivilCode, plain)           & 45.55 & 48.21 & 57.90 & 29.05 &52.96 & 0.64 \\
Dense (CivilCode, plain)          & 68.36 & 71.93 & 79.19
& 29.60 &55.07 & 0.73 \\
\textbf{LEGO (CivilCode,w/ ExpertGraph)} & 72.53 & 74.58 & 81.07 & 30.29 & 54.78 & \textbf{0.85} \\
Gold Article            & 100 & 100 & 100 & 31.12       & 55.41 & 1.00 \\
\bottomrule
\end{tabular}
\end{adjustbox}
\caption{\textbf{Gold Context Approximation on LawExamQA\_Civil.} Coverage is reported as Recall@$k$ at $k\in\{8,10,20\}$. QA: Qwen3-8B zero-shot reader (no CoT), top-8 provisions. GCR = Gap Closing Rate on QA accuracy.  }
\label{tab:gold_coverage}
\end{table*}
\endgroup

% Restore the natural table counter after forcing Table 4.
\setcounter{table}{2}

\subsection{Ablation Study}
\label{sec:layered_ablation_section}

\begin{table}[t]\centering
\scriptsize
\renewcommand{\arraystretch}{1.12}
\setlength{\tabcolsep}{4pt}

\definecolor{tablegreen}{HTML}{7AC373}

\resizebox{\columnwidth}{!}{%
\begin{tabular}{ll cccc}
\toprule
\multirow{2}{*}{\textbf{Model}} & \multirow{2}{*}{\textbf{Description}} & \multicolumn{2}{c}{\textbf{LEGO}} & \multicolumn{2}{c}{\textbf{Improvement ($\Delta$)}} \\
\cmidrule(lr){3-4} \cmidrule(lr){5-6}
 & & \textbf{Acc} & \textbf{F1} & $\Delta$\textbf{Acc} & $\Delta$\textbf{F1} \\
\midrule

\multicolumn{6}{l}{\textit{No-RAG Baselines}} \\
Qwen3-8B & zero-shot & 0.2545 & 0.5382 & \cellcolor{tablegreen!90}$\uparrow$ 0.1508 & \cellcolor{tablegreen!25}$\uparrow$ 0.0305 \\
Qwen3-8B & CoT & 0.2918 & 0.4728 & \cellcolor{tablegreen!70}$\uparrow$ 0.1135 & \cellcolor{tablegreen!90}$\uparrow$ 0.0959 \\
Qwen3-8B & IRAC-CoT & 0.3112 & 0.4973 & \cellcolor{tablegreen!60}$\uparrow$ 0.0941 & \cellcolor{tablegreen!70}$\uparrow$ 0.0714 \\
\midrule

\multicolumn{6}{l}{\textit{RAG + Standard CoT}} \\
Qwen3-8B & Naive RAG + CoT & 0.3430 & 0.5159 & \cellcolor{tablegreen!35}$\uparrow$ 0.0623 & \cellcolor{tablegreen!50}$\uparrow$ 0.0528 \\
Qwen3-8B & Naive RAG + IRAC-CoT & 0.3223 & 0.5312 & \cellcolor{tablegreen!50}$\uparrow$ 0.0830 & \cellcolor{tablegreen!35}$\uparrow$ 0.0375 \\
Qwen3-8B & G-Retriever + CoT & 0.3375 & 0.5074 & \cellcolor{tablegreen!40}$\uparrow$ 0.0678 & \cellcolor{tablegreen!60}$\uparrow$ 0.0613 \\
Qwen3-8B & G-Retriever + IRAC-CoT & 0.3347 & 0.5331 & \cellcolor{tablegreen!45}$\uparrow$ 0.0706 & \cellcolor{tablegreen!30}$\uparrow$ 0.0356 \\
\midrule

\multicolumn{6}{l}{\textit{RAG + ExpertCoT}} \\
Qwen3-8B & Naive RAG + ExpertCoT & 0.3582 & 0.5280 & \cellcolor{tablegreen!20}$\uparrow$ 0.0471 & \cellcolor{tablegreen!40}$\uparrow$ 0.0407 \\
Qwen3-8B & G-Retriever + ExpertCoT & 0.3555 & 0.5415 & \cellcolor{tablegreen!25}$\uparrow$ 0.0498 & \cellcolor{tablegreen!20}$\uparrow$ 0.0272 \\
\midrule

\multicolumn{6}{l}{\textit{ExpertGraphRAG + Standard CoT}} \\
LEGO & CoT & 0.3555 & 0.5312 & \cellcolor{tablegreen!25}$\uparrow$ 0.0498& \cellcolor{tablegreen!35}$\uparrow$ 0.0375 \\
LEGO & IRAC-CoT & 0.3734 & 0.5456 & \cellcolor{tablegreen!15}$\uparrow$ 0.0319 & \cellcolor{tablegreen!15}$\uparrow$ 0.0231 \\
\midrule

\multicolumn{6}{l}{\textit{ExpertGraphRAG + ExpertCoT}} \\
\textbf{LEGO} & \textbf{Full System} & \textbf{0.4053} & \textbf{0.5687} & - & - \\
\midrule

\multicolumn{6}{l}{\textit{Gold Article}} \\
Gold Article & + CoT & 0.3472 & 0.5257 & - & - \\
Gold Article & + IRAC-CoT & 0.3790 & 0.5468 & - & - \\
Gold Article & + ExpertCoT & 0.4122 & 0.5894 & - & - \\
\bottomrule
\end{tabular}%
}

\vspace{1mm}
\caption{\textbf{End-to-end RAG performance and ablation study.} Each row reports the full LEGO system's $\Delta$Acc/$\Delta$F1 over that configuration.}

\label{tab:layered_ablation}
\label{tab:ablation} % alias for appendix_statistical_validation.tex
\end{table}

\begin{figure}[t]
\centering
\includegraphics[draft=false,width=0.90\columnwidth]{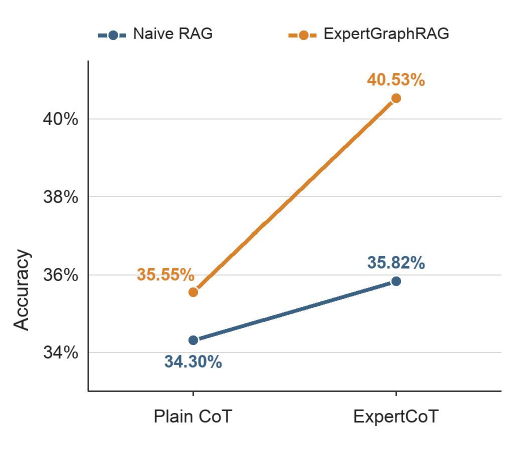}
\caption{\textbf{Interaction between ExpertGraphRAG and \mbox{ExpertCoT}.} Accuracy improves most when both components are enabled.}
\label{fig:synergy_waterfall_acc}
\vspace{-2mm}
\end{figure}

Table~\ref{tab:layered_ablation} shows that the full LEGO system consistently outperforms all baselines and variants in both Accuracy and F1, validating the contribution of each core component. When the RAG module is removed, or when the model relies only on standard prompting strategies such as Qwen3-8B zero-shot, CoT, and IRAC-CoT, performance drops substantially. Compared with these No-RAG baselines, LEGO improves Accuracy by 9.41--15.08 percentage points (pp) and F1 by 3.05--9.59 pp, highlighting the difficulty of complex legal reasoning without external legal knowledge retrieval. Even with a retrieval module, general RAG frameworks paired with standard prompts, such as Naive RAG and G-Retriever with CoT/IRAC-CoT, still lag behind the full system by 6.23--8.30 pp in Accuracy and 3.56--6.13 pp in F1. This suggests that generic retrieval pipelines struggle to capture the rigorous reasoning structure required for statutory application. Moreover, although RAG + ExpertCoT variants benefit from expert reasoning guidance, they remain 4.71--4.98 pp lower in Accuracy and 2.72--4.07 pp lower in F1 than LEGO, underscoring the importance of our structurally optimized ExpertGraphRAG for reducing semantic noise and supporting deeper reasoning chains. Figure~\ref{fig:synergy_waterfall_acc} further shows that the benefit of each component is larger when the other is present: ExpertGraphRAG improves Accuracy by 1.25 pp with Plain CoT but by 4.71 pp with ExpertCoT, while ExpertCoT yields gains of 1.52 and 4.98 pp with Naive RAG and ExpertGraphRAG, respectively. The corresponding interaction
analysis is reported in Appendix~\ref{sec:appendix_synergy_case_study}.

\subsection{Recall of Legal Retrieval }
\label{sec:retrieval_recall}

Table \ref{tab:gold_coverage} shows that ExpertGraph substantially improves gold-context approximation over plain retrieval. Compared with Naive RAG, CivilCode-RAG + ExpertGraph raises Recall@8 from 27.99\% to 72.53\%, Recall@10 from 34.70\% to 74.58\%, and Recall@20 from 50.52\% to 81.07\%. This improved provision coverage also translates into stronger downstream QA performance, increasing accuracy from 25.45\% in the zero-shot setting to 30.29\%. Notably, ExpertGraph achieves substantially higher coverage of gold provisions while relying entirely on automatically retrieved context, indicating that structured legal-graph expansion helps recover relevant provisions missed by plain BM25 or dense retrieval.
To diagnose why ExpertGraph helps, Table~\ref{tab:gold_coverage} decomposes retrieval quality into four facets and reports the \textbf{Gap Closing Rate}
\begin{equation}
\textsc{GCR}(M) \;=\; \frac{\textsc{Acc}(M) - \textsc{Acc}_{\text{zeroshot}}}{\textsc{Acc}_{\text{gold}} - \textsc{Acc}_{\text{zeroshot}}}\,.
\end{equation}

\textbf{LEGO performs close to the gold-article reference.} The ExpertGraph is built
once, at corpus level, independently of the benchmark questions, answer labels and
official rationales, and is then reused across downstream legal reasoning tasks without
any task-specific annotation of gold provisions. Under that setting LEGO reaches
$40.53\%$ accuracy, against $41.22\%$ for the gold-article reference in
Table~\ref{tab:layered_ablation} (Gold Article + ExpertCoT), which is handed the
controlling provisions directly. The $0.69$~pp gap corresponds to only five items out of
723, suggesting that automatic retrieval recovers most of the benefit provided by manually annotated provisions. Although constructing the ExpertGraph requires a non-trivial upfront investment, this cost is incurred only once at the corpus level and can be amortized across multiple downstream datasets and tasks, rather than repeated for each application.

\subsection{Error Analysis}
\label{sec:error_analysis}

Beyond aggregate scores, comparing LEGO's outputs with those of the baselines shows that it mainly reduces three recurring failures of generic RAG systems: retrieving lexically similar but legally irrelevant provisions, producing partial answers based on a single provision, and giving hedged conclusions where the legal issue requires a determinate judgment. By combining ExpertGraphRAG with ExpertCoT, LEGO retrieves structurally relevant provisions, composes them into more complete reasoning chains, and reaches determinate final judgments. The errors that remain are concentrated in the same structural steps: Appendix~\ref{sec:appendix_error_analysis} codes the items LEGO still mis-answers by the earliest reasoning step that makes the answer unrecoverable; they arise mainly from incorrect provision selection and prerequisite violation, with priority misapplication, concept conflation, and exception bypass accounting for most of the rest.

\section{Conclusion}

This paper presents LEGO, a legal-domain, expert-guided dual-module
framework that injects expert legal knowledge into retrieval through
ExpertGraphRAG and into reasoning through ExpertCoT to solve the
challenge of complex legal reasoning tasks. ExpertGraphRAG selects
provisions from an expert-annotated Civil Code graph using the Normative
Coverage Greedy (NCG) algorithm, and ExpertCoT organizes the retrieved
provisions and case facts into structured Provision--Fact--Conclusion
reasoning. Experiments show that, with an 8B backbone, LEGO reaches
40.53\% accuracy on LawExamQA\_Civil, outperforming the evaluated RAG
and CoT baselines and performing comparably to the evaluated larger
models, while remaining robust on multi-hop questions and leading the
evaluated baselines on two open-ended benchmarks. Ablation studies confirm the
individual and complementary contributions of both modules. LEGO thus
improves the complex legal reasoning ability of LLMs and makes the
rationales they generate traceable.

\section*{Limitations}

This paper focuses on a methodological framework rather than proposing a new legal reasoning benchmark. LawExamQA\_Civil is used only as an evaluation set to validate expert-structured retrieval and syllogistic reasoning, not as a standalone public multi-hop benchmark. Its hop count is approximated by the number of Civil Code provisions cited in official rationales, which does not yet capture fine-grained inter-provision transitions such as prerequisite, exception, priority, condition–consequence, or general–special relations. Our experiments evaluate answer accuracy, but do not establish either the faithfulness of the generated analyses to the model’s internal reasoning process or their correctness as post-hoc explanations. Moreover, although the Civil Law ExpertGraph is constructed independently of benchmark questions, answer labels, and official rationales, building and maintaining such a normative graph requires substantial expert effort and may limit scalability across jurisdictions and legal domains. The Fact Graph is a conceptual representation; the implementation passes the case text to the model and does not construct an explicit fact graph.

\section*{Ethics Statement}

LEGO is a methodological research framework for evaluating expert-structured retrieval and syllogistic reasoning in legal question answering. It is not intended to provide legal advice, replace qualified legal professionals, or support automated legal decision-making. All evaluation questions are derived from publicly available judicial-examination materials and are used only for controlled experimental evaluation. This work does not propose LawExamQA\_Civil as a new public benchmark, nor does the evaluation encode or promote any particular legal interpretation. The Civil Law ExpertGraph is constructed independently of benchmark questions, answer labels, and official rationales. Expert annotators were compensated at fair market rates.

\section*{Use of AI Assistants}
We acknowledge the use of several large language models, including Claude, Gemini, GPT, and Qwen, as well as GitHub Copilot, to assist with text editing and code development during this research. All outputs were thoroughly verified by the authors, who maintain full accountability for the paper's contents.

\section*{Acknowledgements}

We thank Modelbest for their technical support. We also acknowledge the Tsinghua University Initiative Scientific Research Program (20255080016) for its support. In addition, we express our gratitude to the legal experts who collected the data and analyzed the cases.

\sloppy Antonino Rotolo was supported by the projects EUSAiR “EU Regulatory Sandboxes for AI” (DIGITAL-2024-AI-ACT-06-SANDBOX), IT4LIA “Italy for Artificial Intelligence”(Grant agreement ID: 101234224), and AISHA “Artificial Intelligence Skills Hub Academy”  (DIGITAL-2025-SKILLS-08-GENAI-ACADEMY-STEP).

% \bibliographystyle{acl_natbib} % style already set in latex/acl.sty; keep only one to avoid BibTeX "Illegal, another \\bibstyle"
% arXiv sometimes runs BibTeX from the project root; make the path explicit.
\bibliography{latex/lego}

\appendix

\section*{Appendix Contents}
\noindent
\begin{itemize}[leftmargin=*,topsep=2pt,itemsep=1pt]
\item \textbf{Appendix~\ref{sec:appendix_prompt_card}:} \hyperref[sec:appendix_prompt_card]{ExpertCoT Prompt Card} (p.~\pageref{sec:appendix_prompt_card}).
\item \textbf{Appendix~\ref{sec:appendix_baseline}:} \hyperref[sec:appendix_baseline]{Baseline Details and Hyperparameters} (p.~\pageref{sec:appendix_baseline}).
\item \textbf{Appendix~\ref{sec:appendix_pfc_case_study_detailed}:} \hyperref[sec:appendix_pfc_case_study_detailed]{P/F/C Case Studies with Pipeline-Level Reasoning} (p.~\pageref{sec:appendix_pfc_case_study_detailed}).
\item \textbf{Appendix~\ref{sec:appendix_datacard}:} \hyperref[sec:appendix_datacard]{Dataset Card} (p.~\pageref{sec:appendix_datacard}).
\item \textbf{Appendix~\ref{sec:appendix_graph}:} \hyperref[sec:appendix_graph]{Legal Expert-Domain Graph Construction} (p.~\pageref{sec:appendix_graph}).
\item \textbf{Appendix~\ref{sec:appendix_graph_statistics}:} \hyperref[sec:appendix_graph_statistics]{Civil Law ExpertGraph: Source Statistics} (p.~\pageref{sec:appendix_graph_statistics}).
\item \textbf{Appendix~\ref{sec:appendix_error_analysis}:} \hyperref[sec:appendix_error_analysis]{Detailed Error Analysis} (p.~\pageref{sec:appendix_error_analysis}).
\item \textbf{Appendix~\ref{sec:appendix_ablation_case_studies}:} \hyperref[sec:appendix_ablation_case_studies]{Case Studies for Ablation Observations} (p.~\pageref{sec:appendix_ablation_case_studies}).
\item \textbf{Appendix~\ref{sec:appendix_multihop_case_study}:} \hyperref[sec:appendix_multihop_case_study]{Multi-Hop Case Studies on LawExamQA\_Civil} (p.~\pageref{sec:appendix_multihop_case_study}).
\item \textbf{Appendix~\ref{sec:appendix_synergy_case_study}:} \hyperref[sec:appendix_synergy_case_study]{Component-Synergy Case Studies on LawExamQA\_Civil} (p.~\pageref{sec:appendix_synergy_case_study}).
\item \textbf{Appendix~\ref{sec:appendix_plawbench_case_study}:} \hyperref[sec:appendix_plawbench_case_study]{PLawBench\_Civil Case Studies} (p.~\pageref{sec:appendix_plawbench_case_study}).
\item \textbf{Appendix~\ref{sec:appendix_lexrag_case_study}:} \hyperref[sec:appendix_lexrag_case_study]{LexRAG\_Civil Case Studies} (p.~\pageref{sec:appendix_lexrag_case_study}).
\item \textbf{Appendix~\ref{app:stats}:} \hyperref[app:stats]{Statistical Validation of the Main Results} (p.~\pageref{app:stats}).
\end{itemize}

\section{ExpertCoT Prompt Card}
\label{sec:appendix_prompt_card}
\label{app:prompt}

This appendix reproduces the full prompts used by LEGO's ExpertCoT step (\S\ref{sec:syllogistic_cot}) on the production run that obtains $0.4053 / 0.5687$ on LawExamQA\_Civil. Both the \textsc{system} and \textsc{user} prompts are passed to the same Qwen3-8B backbone in a single call with greedy decoding (\texttt{temperature} = 0.0, \texttt{max\_tokens} = 1800).

\subsection*{A.1.1\quad System prompt}
\begin{promptcard}
\textbf{PROMPT:} \textit{You are a careful multiple-choice assistant for PRC civil-law questions. You must reason strictly from the case facts, the options, and the provided statutes. Do not invent facts or statutes that are not given.}
\end{promptcard}

\subsubsection{User prompt template}
\noindent The user prompt is a fixed template into which three blocks are filled at run-time: \texttt{\{law\_context\}} (the top-$N$ retrieved Civil Code articles, each rendered as ``\texttt{[k] PRC Civil Code Art.~\textit{c}: ...}''); \texttt{\{case\}} (the case body); and \texttt{\{question\_with\_options\}} (the question stem followed by the four options A--D).

\begin{promptcard}
\textbf{PROMPT:}

Please solve this PRC civil-law multiple-choice question using the P/F/C (Provision--Fact--Conclusion) method.

\textbf{Role:} write as a neutral adjudicator / exam grader.

\textbf{Requirements:}
\begin{enumerate}[leftmargin=*,topsep=2pt,itemsep=0pt]
\item In one sentence, state the legal object to be decided and what the question asks.
\item P-Provision: list only the controlling rules among the \textbf{[Provided Statutes]}; do not add any new statutes.
\item In P, explicitly state the provision relations (general vs. special rule, proviso/exception, limiting condition, remedy) and which one controls the conclusion.
\item Carefully distinguish: contract validity, real-rights transfer, opposability, liability allocation, damages, priority rights / defenses; do not expand one legal effect into another.
\item When you see limiting clauses (e.g., ``shall not be deemed invalid solely because...'', ``does not affect the validity of...'', ``may claim compensation''), interpret them as limitations; do not reverse-infer invalidity or automatic extinction of rights.
\item F-Fact: list only the legally relevant facts; do not write the final legal conclusion.
\item C-Conclusion: first perform a ``provision-relation self-check'', then judge options A/B/C/D. Use at most one sentence per option.
\item Select options according to the question polarity. If it asks for ``incorrect / unlawful / not established'', choose the incorrect option(s).
\item Do not multi-select for coverage. Select an option only when its subject, legal predicate, required elements, and claimed legal effect all match the Provision--Fact mapping.
\item Keep each section to 1--3 sentences; focus on comparing options rather than repeating the case.
\item The last line must be exactly \texttt{[FINAL\_ANSWER]X<eoa>}, e.g., \texttt{[FINAL\_ANSWER]A<eoa>} or \texttt{[FINAL\_ANSWER]BD<eoa>}.
\end{enumerate}

\textbf{[Provided Statutes]}\\
\texttt{\{law\_context\}}

\textbf{[Case]}\\
\texttt{\{case\}}

\textbf{[Question/Options]}\\
\texttt{\{question\_with\_options\}}

\textbf{Output format:}\\
Object:\\[2pt]
P-Provision:\\
Provision relations:\\[2pt]
F-Fact:\\[2pt]
C-Conclusion:\\
Provision-relation self-check:\\[2pt]
\texttt{[FINAL\_ANSWER]...<eoa>}
\end{promptcard}

\subsubsection{End-to-end example (qid 0)}
The following is the user prompt as actually sent for question \texttt{qid 0} (2002 Paper~III No.\,1, the transport-delay / cargo-damage case used as the worked example in Appendix~\ref{sec:appendix_pfc_case_study_detailed}). Only the three slot fillers (\texttt{\{law\_context\}}, \texttt{\{case\}}, \texttt{\{question\_with\_options\}}) differ from the template above.

\paragraph{Slot filler: \texttt{\{law\_context\}}.}
\begin{promptcard}
\setlength{\parindent}{0pt}
\textbf{\{law\_context\}:}

$[$1$]$ \textit{PRC Civil Code} Art.~832: The carrier shall be liable for compensation for any damage to or loss of the goods during transport. However, where the carrier proves that such damage or loss was caused by force majeure, the natural properties of the goods or reasonable wear and tear, or the fault of the consignor or consignee, the carrier shall not be liable for compensation.

\smallskip
$[$2$]$ \textit{PRC Civil Code} Art.~834: Where two or more carriers undertake through-transport by the same mode of transport, the carrier that entered into the contract with the consignor shall be liable for the whole course of transport; where the loss occurs on a particular segment, the carrier that entered into the contract with the consignor and the carrier responsible for that segment shall be jointly and severally liable.

\smallskip
$[$3$]$ \textit{PRC Civil Code} Art.~825: Where the consignor handles the carriage of goods, it shall accurately state to the carrier the name of the consignee, the name or the consignee as instructed, and the necessary information about the goods for carriage such as the name, nature, weight, quantity, and place of delivery. Where the consignor makes a false declaration or omits important information and thereby causes losses to the carrier, the consignor shall be liable for compensation.

\smallskip
$[$4--8$]$ \textit{PRC Civil Code} Arts.~841 / 842 / 824 / 607 / 837 (multimodal transport, passenger transport, risk allocation in sales contracts, deposit/consignation, etc.; full text omitted; not on the main reasoning path of this question).
\end{promptcard}

\paragraph{Slot filler: \texttt{\{case\}}.}
\begin{promptcard}
\textbf{\{case\}:}

Company A needs to ship a batch of goods to Company B (the consignee). The parties agreed that Company A would arrange the logistics. Company A's legal representative, C, contacted by phone and commissioned a trucking company holding a road-transport business license to transport the goods. The trucking company assigned its full-time employee driver, Liu, to drive a dedicated vehicle. During transport, a traffic accident occurred due to Liu's negligence; the traffic police accident report found Liu fully at fault, and the goods were damaged. Company B had previously queried the trucking company directly about the location of the goods. Company B suffered losses because it failed to receive the goods in time.
\end{promptcard}

\paragraph{Slot filler: \texttt{\{question\_with\_options\}}.}
\begin{promptcard}
\setlength{\parindent}{0pt}
\textbf{\{question\_with\_options\}:}

Since the damage was directly caused by the transport side, who should Company B claim compensation from?

\smallskip
Options:\\
A. Company A\\
B. C\\
C. Liu\\
D. The trucking company, as the actual carrier that directly caused the damage; holding it liable for damages accords with tort-law principles.
\end{promptcard}
\paragraph{Expected output skeleton.} The model returns the five labelled sections of \S\ref{sec:syllogistic_cot} populated against this input, ending with the token \texttt{[FINAL\_ANSWER]A<eoa>} (the gold answer for this item). The full P / F / C body is stored in the prediction log and is the source for the worked example in Appendix~\ref{sec:appendix_pfc_case_study_detailed}.

\clearpage
\section{Baseline Details and Hyperparameters}
\label{sec:appendix_baseline}

This section lists the configurations used for baselines reported in Table~\ref{tab:main_results}, Table~\ref{tab:cross_benchmark}, and Table~\ref{tab:layered_ablation}.

\subsection{RAG Baselines}

\paragraph{Naive RAG.}
Naive RAG~\cite{NEURIPS2020_6b493230} retrieves the top-$k$ Civil Code chunks by dense similarity and concatenates them as context for the generator.

\begin{center}
\small
\fbox{\begin{minipage}{0.92\columnwidth}
\textbf{Naive RAG Configuration}\\[3pt]
\rule{\linewidth}{0.3pt}\\[2pt]
\texttt{\{}\\
\hspace*{1.5em}\texttt{embedding\_model:\ Qwen3-Embedding-8B,}\\
\hspace*{1.5em}\texttt{retrieval\_topk:\ 5,}\\
\hspace*{1.5em}\texttt{chunk\_token\_size:\ 1000,}\\
\hspace*{1.5em}\texttt{chunk\_overlap\_token\_size:\ 200}\\
\texttt{\}}
\end{minipage}}
\end{center}

\paragraph{G-Retriever.}
G-Retriever~\cite{gretriever2024} solves a Prize-Collecting Steiner Tree over the Civil Code knowledge graph, returning a connected sub-tree spanning the top-$k$ prize nodes and their relations.

\begin{center}
\small
\fbox{\begin{minipage}{0.92\columnwidth}
\textbf{G-Retriever Configuration}\\[3pt]
\rule{\linewidth}{0.3pt}\\[2pt]
\texttt{\{}\\
\hspace*{1.5em}\texttt{embedding\_model:\ Qwen3-Embedding-8B,}\\
\hspace*{1.5em}\texttt{retrieval\_topk:\ 10,}\\
\hspace*{1.5em}\texttt{chunk\_token\_size:\ 1200,}\\
\hspace*{1.5em}\texttt{chunk\_overlap\_token\_size:\ 100,}\\
\hspace*{1.5em}\texttt{entities\_max\_tokens:\ 3000,}\\
\hspace*{1.5em}\texttt{relationships\_max\_tokens:\ 2000}\\
\texttt{\}}
\end{minipage}}
\end{center}

\paragraph{LightRAG.}
LightRAG~\cite{guo-etal-2025-lightrag} performs hybrid retrieval over an auto-induced entity--relation graph, scoring queries against a global-relation channel and a local-entity channel with separate token budgets.

\begin{center}
\small
\fbox{\begin{minipage}{0.92\columnwidth}
\textbf{LightRAG Configuration}\\[3pt]
\rule{\linewidth}{0.3pt}\\[2pt]
\texttt{\{}\\
\hspace*{1.5em}\texttt{embedding\_model:\ Qwen3-Embedding-8B,}\\
\hspace*{1.5em}\texttt{query\_type:\ hybrid,}\\
\hspace*{1.5em}\texttt{retrieval\_topk:\ 20,}\\
\hspace*{1.5em}\texttt{chunk\_token\_size:\ 1200,}\\
\hspace*{1.5em}\texttt{chunk\_overlap\_token\_size:\ 100,}\\
\hspace*{1.5em}\texttt{max\_token\_global\_context:\ 2000,}\\
\hspace*{1.5em}\texttt{max\_token\_local\_context:\ 2000,}\\
\hspace*{1.5em}\texttt{max\_token\_text\_unit:\ 2000}\\
\texttt{\}}
\end{minipage}}
\end{center}

\paragraph{HippoRAG 2.}
HippoRAG~2~\cite{hipporag2025} links query mentions into a pre-built fact / passage hybrid graph and accumulates evidence via multi-step traversal before returning the final passages.

\begin{center}
\small
\fbox{\begin{minipage}{0.92\columnwidth}
\textbf{HippoRAG 2 Configuration}\\[3pt]
\rule{\linewidth}{0.3pt}\\[2pt]
\texttt{\{}\\
\hspace*{1.5em}\texttt{embedding\_model:\ Qwen3-Embedding-8B,}\\
\hspace*{1.5em}\texttt{retrieval\_top\_k:\ 10,}\\
\hspace*{1.5em}\texttt{linking\_top\_k:\ 7,}\\
\hspace*{1.5em}\texttt{qa\_top\_k:\ 10,}\\
\hspace*{1.5em}\texttt{max\_qa\_steps:\ 3,}\\
\hspace*{1.5em}\texttt{graph\_type:\ facts\_and\_sim\_passage\_}\\
\hspace*{3em}\texttt{node\_unidirectional}\\
\texttt{\}}
\end{minipage}}
\end{center}

\paragraph{RAPTOR.}
RAPTOR~\cite{raptor2024} constructs a hierarchical summary tree by recursively clustering and summarising chunks, enabling retrieval at multiple levels of abstraction; we use the collapsed-tree variant.

\begin{center}
\small
\fbox{\begin{minipage}{0.92\columnwidth}
\textbf{RAPTOR Configuration}\\[3pt]
\rule{\linewidth}{0.3pt}\\[2pt]
\texttt{\{}\\
\hspace*{1.5em}\texttt{embedding\_model:\ Qwen3-Embedding-8B,}\\
\hspace*{1.5em}\texttt{retrieval\_topk:\ 10,}\\
\hspace*{1.5em}\texttt{chunk\_token\_size:\ 1200,}\\
\hspace*{1.5em}\texttt{chunk\_overlap\_token\_size:\ 100,}\\
\hspace*{1.5em}\texttt{num\_layers:\ 5,}\\
\hspace*{1.5em}\texttt{max\_length\_in\_cluster:\ 3500,}\\
\hspace*{1.5em}\texttt{threshold:\ 0.1,}\\
\hspace*{1.5em}\texttt{cluster\_metric:\ cosine,}\\
\hspace*{1.5em}\texttt{threshold\_cluster\_num:\ 5000,}\\
\hspace*{1.5em}\texttt{max\_context\_tokens:\ 4500}\\
\texttt{\}}
\end{minipage}}
\end{center}

\subsection{CoT Baselines}

CoT baselines pair with any retriever in the ablation: the retrieved articles fill the \texttt{law\_context} block while the scaffold controls how the generator consumes them.

\paragraph{Zero-shot CoT.}
Zero-shot CoT~\cite{kojima2022large} prepends ``Let's think step by step'' and lets the model freely generate a reasoning chain before committing an answer.

\begin{center}
\small
\fbox{\begin{minipage}{0.92\columnwidth}
\textbf{Zero-shot CoT Configuration}\\[3pt]
\rule{\linewidth}{0.3pt}\\[2pt]
\texttt{\{}\\
\hspace*{1.5em}\texttt{backbone:\ Qwen3-8B,}\\
\hspace*{1.5em}\texttt{sampling:\ greedy,}\\
\hspace*{1.5em}\texttt{max\_tokens:\ 1800,}\\
\hspace*{1.5em}\texttt{top\_n\_articles:\ 8,}\\
\hspace*{1.5em}\texttt{scaffold:\ "Let's think step by step",}\\
\hspace*{1.5em}\texttt{output\_skeleton:\ [Reasoning] \(\to\)}\\
\hspace*{3em}\texttt{[Option check] \(\to\)}\\
\hspace*{3em}\texttt{[Answer]X\textless eoa\textgreater}\\
\texttt{\}}
\end{minipage}}
\end{center}

\paragraph{IRAC-CoT.}
IRAC-CoT~\cite{yu2025benchmarking} forces a four-section Issue--Rule--Application--Conclusion trace before the answer token.

\begin{center}
\small
\fbox{\begin{minipage}{0.92\columnwidth}
\textbf{IRAC-CoT Configuration}\\[3pt]
\rule{\linewidth}{0.3pt}\\[2pt]
\texttt{\{}\\
\hspace*{1.5em}\texttt{backbone:\ Qwen3-8B,}\\
\hspace*{1.5em}\texttt{sampling:\ greedy,}\\
\hspace*{1.5em}\texttt{max\_tokens:\ 1800,}\\
\hspace*{1.5em}\texttt{top\_n\_articles:\ 8,}\\
\hspace*{1.5em}\texttt{scaffold:\ Issue \(\to\) Rule \(\to\)}\\
\hspace*{3em}\texttt{Application \(\to\) Conclusion,}\\
\hspace*{1.5em}\texttt{output\_skeleton:\ [Issue] \(\to\) [Rule] \(\to\)}\\
\hspace*{3em}\texttt{[Application] \(\to\)}\\
\hspace*{3em}\texttt{[Conclusion] \(\to\)}\\
\hspace*{3em}\texttt{[Answer]X\textless eoa\textgreater}\\
\texttt{\}}
\end{minipage}}
\end{center}

\subsection{Experiment Settings}
\label{sec:appendix_hyperparams}
For LEGO and all retrieval-based baselines, we use Qwen3-8B as the generator to ensure a fair backbone-controlled comparison. During generation, all open-source experiments use greedy decoding with \texttt{temperature}=0 and \texttt{max\_tokens}=1800 (512 for the closed-book zero-shot setting, whose output is a bare answer tag); the Qwen3-8B reader is run with thinking mode disabled (\texttt{enable\_thinking=False}) in all configurations; API-based experiments use the providers' default decoding configurations. For LEGO's provision retrieval on LawExamQA\_Civil, we instantiate the Normative Coverage Greedy (NCG) objective of Eq.~\ref{eq:retrieval_gain}--\ref{eq:submodular_retrieval} with the weights listed below; all \(d_i\) / \(g_i\) similarities are min--max normalized to \([0,1]\) per query (\(d_i\)) before the greedy selection. To guarantee reproducibility, we fix the random seed at 42. All open-source model inferences and evaluations are conducted on a single server equipped with 8 \(\times\) NVIDIA H100 (80GB HBM3) GPUs.

\begin{center}
\small
\fbox{\begin{minipage}{0.92\columnwidth}
\textbf{LEGO Configuration}\\[3pt]
\rule{\linewidth}{0.3pt}\\[2pt]
\texttt{\{}\\
\hspace*{1.5em}\texttt{backbone:\ Qwen3-8B,}\\
\hspace*{1.5em}\texttt{embedding\_model:\ Qwen3-Embedding-8B,}\\
\hspace*{1.5em}\texttt{sampling:\ greedy,}\\
\hspace*{1.5em}\texttt{temperature:\ 0.0,}\\
\hspace*{1.5em}\texttt{max\_tokens:\ 1800,}\\
\hspace*{1.5em}\texttt{seed:\ 42,}\\
\hspace*{1.5em}\texttt{\(\lambda_{\mathrm{align}}\):\ 0.55,\quad // Eq.~\ref{eq:retrieval_gain}}\\
\hspace*{1.5em}\texttt{\(\lambda_{\mathrm{cov}}\):\ 0.18,\quad // Eq.~\ref{eq:retrieval_gain}}\\
\hspace*{1.5em}\texttt{\(\lambda_{\mathrm{red}}\):\ 0.06,\quad // Eq.~\ref{eq:retrieval_gain}}\\
\hspace*{1.5em}\texttt{article\_budget\ K:\ 8,\quad // Eq.~\ref{eq:submodular_retrieval}}\\
\hspace*{1.5em}\texttt{top\_n\_articles\ N:\ 8}\\
\texttt{\}}
\end{minipage}}
\end{center}

\noindent The three weights are lightly tuned on a held-out development split, giving the largest weight to query--article alignment (\(\lambda_{\mathrm{align}}=0.55\)), a moderate weight to marginal normative-rule coverage (\(\lambda_{\mathrm{cov}}=0.18\)), and a small redundancy penalty (\(\lambda_{\mathrm{red}}=0.06\)). The article budget \(K=N=8\) means NCG selects eight articles per query and all eight are passed into the ExpertCoT prompt as \texttt{\{law\_context\}}.

% ---- BEGIN inlined appendix_pfc_case_study_detailed.tex ----
% Appendix-ready snippet.
% Requires: booktabs, adjustbox, enumitem, xcolor/colortbl.

\section{P/F/C Case Studies with Pipeline-Level Reasoning}
\label{sec:appendix_pfc_case_study_detailed}

\providecommand{\pfcgood}[1]{\cellcolor{green!13}#1}
\providecommand{\pfcbad}[1]{\cellcolor{red!13}#1}
\providecommand{\pfcnote}[1]{\cellcolor{yellow!18}#1}

This appendix expands six representative ablation examples into case-level
P/F/C (Provision--Fact--Conclusion) comparisons. The original benchmark cases
are Chinese civil-law multiple-choice questions; to avoid encoding problems
in the appendix, the case facts and options below are English renderings of the
original items. For each case, green cells mark legally correct P/F/C steps,
red cells mark the step that causes or directly contributes to the wrong
answer, and yellow cells mark partially correct reasoning that still misses the
decisive legal boundary.

\begin{table*}[t]
\centering
\small
\begin{adjustbox}{max width=\textwidth}
\begin{tabular}{p{2.0cm}p{0.7cm}p{0.7cm}p{4.8cm}p{5.2cm}}
\toprule
\textbf{Observation} & \textbf{QID} & \textbf{Gold} & \textbf{Prediction contrast} & \textbf{Dominant P/F/C difference} \\
\midrule
Obs.~1 retrieval helps & 19 & A &
zero-shot B vs.\ RAG/LEGO A &
P: retrieved transfer-consent rules block implied-consent intuition. \\
Obs.~1 retrieval insufficient & 4 & D &
plain RAG B vs.\ LEGO Full D &
P: apparent agency must override bare no-authority reasoning. \\
Obs.~2 ExpertCoT at fixed retrieval & 1 & C &
CoT CD vs.\ ExpertCoT C &
C: termination damages are not full agreed remuneration. \\
Obs.~3 ExpertGraph alone & 36 & B &
plain RAG/RAG+ExpertCoT C vs.\ LEGO CoT B &
P: graph supplies the negative rule that no remuneration claim exists. \\
Obs.~4 synergy & 7 & D &
single components A/B vs.\ LEGO Full D &
P+F: the good-faith, no-present-benefit exception controls the result. \\
Obs.~5 strongest baseline & 16 & C &
G-Retriever+ExpertCoT D vs.\ LEGO Full C &
C: the private return claim is limited to principal plus legal fruits. \\
\bottomrule
\end{tabular}
\end{adjustbox}
\caption{\textbf{Six representative P/F/C case studies.}}
\label{tab:pfc_case_overview}
\end{table*}

\subsection{QID 19: Contract Rights and Obligations Transfer}

\paragraph{Case.}
Company A and Korean Company B formed a Chinese-foreign joint venture, and the
joint-venture contract had already been approved by the competent authority.
Company C is a large state-owned enterprise with the relevant qualifications.
Company A wrote to Company B: ``We plan to transfer our equity to Company C. If
you object, please raise a written objection within 30 days; otherwise you will
be deemed to have consented.'' Company A then transferred its contractual rights
and obligations to Company C without Company B's consent. Company B did not
reply in writing for three months after receiving the notice. Company C had
already participated in the joint venture's operations for one year, with good
business results, and Company B had once signed a board resolution together
with Company C.

\paragraph{Question and options.}
Given that Company C has actually performed contractual obligations and
Company B did not object in time, is the transfer of the contract illegal?
\begin{enumerate}[label=\Alph*.]
\item Yes.
\item No. Based on commercial efficiency and actual performance, the transfer
should be treated as valid, and Company B impliedly recognized it by joining
board actions.
\item It is illegal only if Company B expressly objects.
\item It depends on whether the approval authority later ratifies the transfer.
\end{enumerate}

\begin{table*}[t]
\centering
\small
\setlength{\tabcolsep}{4pt}
\renewcommand{\arraystretch}{1.1}
\begin{adjustbox}{max width=\textwidth}
\begin{tabular}{p{2.8cm}p{0.9cm}p{3.7cm}p{3.7cm}p{3.9cm}}
\toprule
\textbf{Pipeline} & \textbf{Pred.} & \textbf{P-Provision} & \textbf{F-Fact} & \textbf{C-Conclusion} \\
\midrule
Qwen3-8B zero-shot &
\pfcbad{B} &
\pfcbad{Does not anchor the answer in Articles 551, 555, and 556.} &
\pfcbad{Overweights actual operation, silence, and board participation.} &
\pfcbad{Treats commercial efficiency and silence as implied consent.} \\
Naive RAG + CoT &
\pfcgood{A} &
\pfcgood{Identifies consent as required for transfer of rights and obligations.} &
\pfcgood{Makes the lack of Company B's consent decisive.} &
\pfcgood{Finds the transfer illegal and selects A.} \\
G-Retriever + ExpertCoT &
\pfcbad{D} &
\pfcgood{Lists the transfer-consent rules.} &
\pfcgood{Extracts the key facts correctly.} &
\pfcbad{Shifts the answer from consent to later administrative ratification.} \\
LEGO Full &
\pfcgood{A} &
\pfcgood{Combines Articles 551, 555, and 556 as validity conditions.} &
\pfcgood{Treats silence and performance as insufficient substitutes for consent.} &
\pfcgood{Concludes that the missing consent makes the transfer illegal.} \\
\bottomrule
\end{tabular}
\end{adjustbox}
\caption{P/F/C comparison for QID 19.}
\label{tab:pfc_qid19}
\end{table*}

\paragraph{Difference analysis.}
The legal core is consent for a combined transfer of contractual rights and
obligations. Civil Code Articles 551, 555, and 556 require consent by the other
party; Article 140 permits implied expression but does not make silence equal
consent by default. RAG pipelines improve because retrieval supplies this major
premise. LEGO Full adds a relation check in the P stage, preventing later
performance or business efficiency from being converted into statutory consent.

\subsection{QID 4: Stamped Blank Contract and Apparent Agency}

\paragraph{Case.}
Zhang was a salesperson of an enterprise and carried blank contract forms
bearing the enterprise's official seal for external contracting. Because Zhang
accepted kickbacks, the enterprise formally removed him and circulated an
internal email about the removal, but it did not recover the stamped blank
contracts. On the day after leaving the enterprise, Zhang concealed his
departure and used one of the stamped blank contracts to sign a purchase and
sale agreement with a counterparty. The counterparty only checked the
authenticity of the seal and did not ask Zhang to produce proof of employment.
The enterprise's internal rules state that former employees may not use company
contract forms. The enterprise also copied Zhang's removal notice to all
long-term business partners, including the counterparty.

\paragraph{Question and options.}
Because Zhang had been removed and had no current authorization, and because he
did not produce any updated authorization document, how should the purchase and
sale agreement be characterized?
\begin{enumerate}[label=\Alph*.]
\item It was not formed.
\item It is invalid because Zhang lacked agency authority, and the counterparty
was clearly negligent in ignoring the enterprise's formal notice and relying
only on past impressions.
\item It is voidable.
\item It was formed and became effective.
\end{enumerate}

\begin{table*}[t]
\centering
\small
\begin{adjustbox}{max width=\textwidth}
\begin{tabular}{p{2.8cm}p{0.9cm}p{3.7cm}p{3.7cm}p{3.9cm}}
\toprule
\textbf{Pipeline} & \textbf{Pred.} & \textbf{P-Provision} & \textbf{F-Fact} & \textbf{C-Conclusion} \\
\midrule
Naive RAG + CoT &
\pfcbad{B} &
\pfcbad{Stops at the general no-authority rule and misses apparent agency.} &
\pfcbad{Overweights removal, no updated authorization, and lack of checking.} &
\pfcbad{Infers invalidity directly from lack of authority.} \\
Naive RAG + ExpertCoT &
\pfcgood{D} &
\pfcgood{Treats Article 172 apparent agency as the special effectiveness rule.} &
\pfcgood{Uses the stamped blank contract as an appearance attributable to the enterprise.} &
\pfcgood{Finds apparent agency and selects D.} \\
LEGO CoT &
\pfcbad{B} &
\pfcbad{Mixes Articles 171, 172, and 504 without stable priority.} &
\pfcbad{Treats failure to verify current employment as decisive fault.} &
\pfcbad{Incorrectly classifies the agreement as invalid.} \\
LEGO Full &
\pfcgood{D} &
\pfcgood{Gives Article 172 priority over the general no-authority rule.} &
\pfcgood{Focuses on the unrecovered stamped blank contract as apparent authority.} &
\pfcgood{Finds the agreement formed and effective.} \\
\bottomrule
\end{tabular}
\end{adjustbox}
\caption{P/F/C comparison for QID 4.}
\label{tab:pfc_qid4}
\end{table*}

\paragraph{Difference analysis.}
Article 172 provides that an unauthorized agency act is effective if the
counterparty has reason to believe authority exists. The civil-law graph lists
stamped blank contracts as a typical apparent-authority appearance. Correct
pipelines succeed because the P stage treats apparent agency as a special rule
that can override bare no-authority reasoning. Plain RAG and LEGO CoT fail by
overweighting absence of current authority.

\paragraph{Doctrinal caveat.}
The rewritten facts include notice of Zhang's removal to the counterparty. If
that notice is treated as effectively received and understood, reasonable
reliance may be weakened. The benchmark gold and source explanation preserve
the original apparent-agency conclusion, so this case should be presented as
LEGO Full matching the benchmark's rule hierarchy, not as a universal
conclusion for all notice variants.

\subsection{QID 1: Mandate Termination and Loss Compensation}

\paragraph{Case.}
Party A orally authorized Party B to purchase timber on A's behalf, and the
parties agreed that B would receive remuneration for the service. B spent time
and effort on the matter, repeatedly visited timber markets, and contacted
several suppliers. B had already selected specific timber specifications and
reported quotations and stock information to A, who raised no objection. Later,
A no longer wanted the timber and telephoned B to cancel the mandate without
mentioning compensation. B objected.

\paragraph{Question and options.}
A orally mandated B to purchase timber and agreed to pay remuneration. A now
cancels the mandate. Given that B has performed substantial work and completed
the main entrusted affairs, which statements are correct?
\begin{enumerate}[label=\Alph*.]
\item A has no right to unilaterally cancel the mandate; otherwise A must
compensate B's losses.
\item A may unilaterally cancel the mandate, but only in writing.
\item A may unilaterally cancel the mandate, but must compensate B's losses.
\item Because B has performed substantial work, A must still pay the agreed
remuneration so that B receives consideration for the work product.
\end{enumerate}

\begin{table*}[t]
\centering
\small
\begin{adjustbox}{max width=\textwidth}
\begin{tabular}{p{2.8cm}p{0.9cm}p{3.7cm}p{3.7cm}p{3.9cm}}
\toprule
\textbf{Pipeline} & \textbf{Pred.} & \textbf{P-Provision} & \textbf{F-Fact} & \textbf{C-Conclusion} \\
\midrule
Naive RAG + CoT &
\pfcbad{CD} &
\pfcgood{Finds Article 933 on discretionary termination and loss compensation.} &
\pfcnote{Correctly identifies paid mandate and work already done.} &
\pfcbad{Expands loss compensation into full agreed remuneration.} \\
G-Retriever + CoT &
\pfcbad{CD} &
\pfcgood{Identifies the mandate termination rule.} &
\pfcnote{Uses work-performance facts too strongly to support D.} &
\pfcbad{Overselects D and broadens the legal effect.} \\
Naive RAG + ExpertCoT &
\pfcgood{C} &
\pfcgood{Separates the termination right from the compensation duty.} &
\pfcgood{Uses the work facts only to support loss, not full remuneration.} &
\pfcgood{Selects only C.} \\
LEGO Full &
\pfcgood{C} &
\pfcgood{Keeps the paid-mandate loss rule clear.} &
\pfcgood{Does not convert substantial work into a full remuneration claim.} &
\pfcgood{Matches the option predicate and rejects D.} \\
\bottomrule
\end{tabular}
\end{adjustbox}
\caption{P/F/C comparison for QID 1.}
\label{tab:pfc_qid1}
\end{table*}

\paragraph{Difference analysis.}
Article 933 gives both parties a discretionary termination right and then
allocates loss compensation. The improvement is not retrieval but C-stage
legal-effect matching: ``compensate loss'' and ``pay agreed remuneration'' are
distinct consequences. ExpertCoT succeeds because it checks the option
predicate against the exact statutory consequence.

\subsection{QID 36: Emergency Rescue and Negotiorum Gestio}

\paragraph{Case.}
Zhang was travelling in a scenic area when he saw a woman standing alone near a
cliff with an unusual expression. When the woman jumped, Zhang urgently grabbed
her clothing and pulled her back. During the rescue, Zhang's camera was
damaged, his arm was scratched, and he advanced medical, bandaging, food, and
lodging expenses. The woman's family later expressed gratitude and said they
were willing to compensate lost work time. Zhang was a medical professional who
encountered the incident on his way home from work.

\paragraph{Question and options.}
Given that Zhang spent time and money on the rescue, may Zhang ask the woman to
pay a certain remuneration?
\begin{enumerate}[label=\Alph*.]
\item He may request full remuneration.
\item He may not request remuneration.
\item Based on fairness, he may request appropriate remuneration.
\item He may request remuneration from the woman's family.
\end{enumerate}

\begin{table*}[t]
\centering
\small
\begin{adjustbox}{max width=\textwidth}
\begin{tabular}{p{2.8cm}p{0.9cm}p{3.7cm}p{3.7cm}p{3.9cm}}
\toprule
\textbf{Pipeline} & \textbf{Pred.} & \textbf{P-Provision} & \textbf{F-Fact} & \textbf{C-Conclusion} \\
\midrule
Naive RAG + CoT &
\pfcbad{C} &
\pfcbad{Conflates necessary expenses, appropriate compensation, and remuneration.} &
\pfcbad{Overweights advanced expenses, lost work time, and family gratitude.} &
\pfcbad{Uses fairness to create a remuneration claim.} \\
Naive RAG + ExpertCoT &
\pfcbad{C} &
\pfcbad{Uses P/F/C but misses the no-remuneration legal effect.} &
\pfcbad{Treats compensation facts as a remuneration basis.} &
\pfcbad{Selects C.} \\
LEGO CoT &
\pfcgood{B} &
\pfcgood{Graph expansion supplies the negative rule: no remuneration claim.} &
\pfcgood{Treats family gratitude as neither mandate nor public reward.} &
\pfcgood{Because the question asks about remuneration, selects B.} \\
LEGO Full &
\pfcgood{B} &
\pfcgood{Separates rescue immunity, expenses, compensation, and remuneration.} &
\pfcgood{Expense facts do not change the legal nature of remuneration.} &
\pfcgood{Rejects C and D, selects B.} \\
\bottomrule
\end{tabular}
\end{adjustbox}
\caption{P/F/C comparison for QID 36.}
\label{tab:pfc_qid36}
\end{table*}

\paragraph{Difference analysis.}
Article 979 allows reimbursement of necessary expenses and appropriate
compensation for losses; it does not create remuneration. The civil-law graph
explicitly states that the manager has no remuneration claim. LEGO CoT already
succeeds because ExpertGraph provides this negative legal effect. Plain RAG and
RAG+ExpertCoT fail because they see expense or compensation facts but miss the
legal boundary of remuneration.

\subsection{QID 7: Misdelivered Milk and Good-Faith Recipient}

\paragraph{Case.}
Because of a delivery worker's negligence, milk ordered by Wang was mistakenly
placed in the milk box of Wang's neighbor Zhang. Zhang did not understand why
the milk was there and, without checking, directly took it out and discarded
it. The milk package bore Wang's name, address, and order number. Wang
discovered the missing milk the next day and demanded compensation from Zhang.
The milk was indeed Wang's lawful property.

\paragraph{Question and options.}
After Zhang discarded another person's milk that had been placed in his milk
box, and given that Wang did suffer property loss and Zhang did discard the
milk, how should Zhang's conduct be characterized?
\begin{enumerate}[label=\Alph*.]
\item It constitutes unjust enrichment because Zhang obtained a benefit without
legal basis and caused loss.
\item It constitutes tort liability, either as intentional destruction of
another's property or as negligent disposal causing loss.
\item It constitutes unauthorized agency.
\item It involves no legal wrong.
\end{enumerate}

\begin{table*}[t]
\centering
\small
\begin{adjustbox}{max width=\textwidth}
\begin{tabular}{p{2.8cm}p{0.9cm}p{3.7cm}p{3.7cm}p{3.9cm}}
\toprule
\textbf{Pipeline} & \textbf{Pred.} & \textbf{P-Provision} & \textbf{F-Fact} & \textbf{C-Conclusion} \\
\midrule
Naive RAG + ExpertCoT &
\pfcbad{A} &
\pfcbad{Uses only the general unjust-enrichment elements and misses the good-faith limitation.} &
\pfcbad{Emphasizes name, address, and Wang's loss while weakening Zhang's good faith.} &
\pfcbad{Incorrectly finds unjust enrichment.} \\
G-Retriever + ExpertCoT &
\pfcbad{B} &
\pfcbad{Shifts to tort fault without stabilizing the good-faith recipient rule.} &
\pfcbad{Treats failure to check as sufficient legal fault.} &
\pfcbad{Incorrectly finds tort liability.} \\
LEGO CoT &
\pfcbad{B} &
\pfcnote{Rejects unjust enrichment but drifts to tort.} &
\pfcbad{Still treats non-checking as enough fault.} &
\pfcbad{Selects B.} \\
LEGO Full &
\pfcgood{D} &
\pfcgood{Checks both the general unjust-enrichment rule and the good-faith exception.} &
\pfcgood{Centers good faith, no retained benefit, and no intentional destruction.} &
\pfcgood{Rejects A and B, selects D.} \\
\bottomrule
\end{tabular}
\end{adjustbox}
\caption{P/F/C comparison for QID 7.}
\label{tab:pfc_qid7}
\end{table*}

\paragraph{Difference analysis.}
The gold explanation treats Zhang's lack of awareness as decisive. Zhang was a
good-faith recipient; the milk no longer existed; no present benefit remained.
The graph lists unjust-enrichment elements and the good-faith recipient's
present-benefit limitation. This case demonstrates synergy: ExpertCoT alone is
pulled to A by the general unjust-enrichment rule, while graph-alone/free CoT
drifts to B by over-reading negligence. LEGO Full combines the exception and
the fact filter.

\subsection{QID 16: Bank Overpayment and Business Profit}

\paragraph{Case.}
When Party A withdrew money from a bank, a bank employee mistakenly overpaid
A by 10,000 yuan because of a counting error. A knew the extra money was not
owed, but used the 10,000 yuan as capital for business, quickly bought goods
and resold them, and made a profit of 5,000 yuan, of which 2,000 yuan was cost
for labor and management. After discovering a cash shortage, the bank traced
the flow of funds and confirmed that the profit was directly generated from
the overpaid money. One month later the bank discovered the overpayment and
asked A to return the principal and corresponding gains. A refused.

\paragraph{Question and options.}
Given that A used the bank's funds for profit and that the bank suffered an
interest loss, which statement is correct?
\begin{enumerate}[label=\Alph*.]
\item A does not need to return anything because the problem was entirely
caused by the bank's own mistake.
\item A should return the overpaid 10,000 yuan, with no other liability.
\item A should return the overpaid 10,000 yuan and one month's interest.
\item A should return the overpaid 10,000 yuan, one month's interest, and the
entire 3,000 yuan net profit generated by using the money.
\end{enumerate}

\begin{table*}[t]
\centering
\small
\begin{adjustbox}{max width=\textwidth}
\begin{tabular}{p{2.8cm}p{0.9cm}p{3.7cm}p{3.7cm}p{3.9cm}}
\toprule
\textbf{Pipeline} & \textbf{Pred.} & \textbf{P-Provision} & \textbf{F-Fact} & \textbf{C-Conclusion} \\
\midrule
G-Retriever + ExpertCoT &
\pfcbad{D} &
\pfcbad{Uses an overbroad rule requiring return of all gains from the object.} &
\pfcbad{Overweights that the business profit was produced by the overpaid money.} &
\pfcbad{Awards the business profit to the bank.} \\
LEGO CoT &
\pfcbad{D} &
\pfcbad{Expands damages under Article 987 into full return of net profit.} &
\pfcbad{Underweights labor, management, and business intervention.} &
\pfcbad{Selects D.} \\
Naive RAG + ExpertCoT &
\pfcgood{C} &
\pfcgood{Limits the return scope to principal and legal fruits.} &
\pfcgood{Distinguishes the bank's use-of-money loss from business profit.} &
\pfcgood{Selects C.} \\
LEGO Full &
\pfcgood{C} &
\pfcgood{Uses Articles 122 and 985 for return, without expanding Article 987 to D.} &
\pfcgood{Treats principal and interest as the private return scope.} &
\pfcgood{Rejects D, selects C.} \\
\bottomrule
\end{tabular}
\end{adjustbox}
\caption{P/F/C comparison for QID 16.}
\label{tab:pfc_qid16}
\end{table*}

\paragraph{Difference analysis.}
The gold explanation gives the bank principal plus one month's interest. The
interest is the legal fruit of money; downstream business profit is not awarded
to the bank under the original exam's private return claim. Wrong traces have
the right broad domain, unjust enrichment, but over-extend the legal
consequence. LEGO Full succeeds by checking whether option D's consequence is
strictly authorized by the controlling rule and the benchmark explanation.

\paragraph{Doctrinal caveat.}
The civil-law graph contains a broad note that returnable benefit may include
gains obtained by using the original object. Some modern unjust-enrichment
analyses may therefore support a broader answer. The benchmark gold follows
the original exam explanation; this case should be used to illustrate
legal-effect boundary control, not a universal profit-return rule.

\subsection{Cross-Case Lessons}

Across the six cases, the observed failures are mostly P/F/C mismatches rather
than answer-format errors. Retrieval helps when the missing major premise is
the bottleneck (QID 19). ExpertCoT helps when a retrieved rule must be mapped
to the exact option predicate (QID 1). ExpertGraph helps when the missing
information is a negative legal effect or exception (QID 36). LEGO Full is
strongest when both are needed: the graph supplies the right rule
neighborhood, and P/F/C prevents the C step from expanding one legal
consequence into another (QID 7 and QID 16).

% ---- END inlined appendix_pfc_case_study_detailed.tex ----

\section{Dataset Card}
\label{sec:appendix_datacard}

\textbf{LawExamQA\_Civil} 
\begin{table}[H]
\centering
\small
\setlength{\tabcolsep}{4pt}
\begin{tabular}{lc}
\toprule
\textbf{Statistic} & \textbf{Value} \\
\midrule
Total items                                  & 723 \\
\midrule
\multicolumn{2}{l}{\emph{Question type distribution}} \\
~~Single-select                              & 365 \\
~~Multi-select                               & 254 \\
~~True/False                                 & 104 \\
~~Subjective (converted to options)          &  26\\
\midrule
\multicolumn{2}{l}{\emph{Hop count distribution}} \\
1-hop & 342 \\
2-hop & 215 \\
3-hop & 104 \\
$\geq$4-hop & 62 \\

\midrule
\multicolumn{2}{l}{\emph{Subject area distribution}} \\
~~Contracts                                  & 291 \\
~~Property                                   & 167 \\
~~Torts                                      & 84 \\
~~Marriage \& Family                         & 48 \\
~~Succession                                 & 25 \\
~~Other                                      & 108 \\
\midrule
Avg.\ tokens                        & 158.06 \\
Avg.\ option tokens                          & 14.82 \\

Vocabulary Hit Rate (median)                 & 0.095 \\
\bottomrule
\end{tabular}
\caption{\textbf{LawExamQA\_Civil Dataset Card.} Detailed statistics for the 723-item release; the 26 subjective questions converted to options form an overlapping subset.}
\label{tab:datacard}
\end{table}

We construct LawExamQA\_Civil from publicly available Chinese National Judicial Examination questions in the civil‑law domain. The dataset contains both objective multiple‑choice items and subjective questions. To enable unified automatic evaluation, subjective questions are converted into option‑style items while preserving their original legal‑reasoning requirements.

For each question, we use the accompanying official explanation to identify the statutory provisions involved in the reasoning chain. The number of distinct provisions cited or required by the explanation is treated as the \textit{hop count}, providing a direct measure of statutory reasoning complexity. 

\textbf{LexRAG\_Civil}

\begin{table}[H]
\centering
\small
\setlength{\tabcolsep}{4pt}
\begin{tabular}{lc}
\toprule
\textbf{Statistic} & \textbf{Value} \\
\midrule
Total items          & 140 \\
Avg.\ question chars & 21.0 \\
\bottomrule
\end{tabular}
\caption{\textbf{LexRAG\_Civil Dataset Card.} Statistics for the LexRAG\_Civil subset used in our cross-benchmark evaluation.}
\label{tab:datacard_lexrag}
\end{table}

\noindent LexRAG~\cite{li2025lexrag} is the first benchmark targeting RAG systems in multi-turn legal consultation, containing 1{,}013 expert-annotated five-round dialogues paired with a 17{,}228-article candidate pool; each generated response is scored by an LLM-as-judge along five rubrics---\textit{factuality}, \textit{user satisfaction}, \textit{clarity}, \textit{logical coherence}, and \textit{completeness}---with an expert-response anchor of $8/10$. LexRAG\_Civil is the civil-law subset obtained by filtering LexRAG for civil-law items and randomly sampling.

\textbf{PLawBench\_Civil}

\begin{table}[H]
\centering
\small
\setlength{\tabcolsep}{4pt}
\begin{tabular}{lc}
\toprule
\textbf{Statistic} & \textbf{Value} \\
\midrule
Total items          & 114 \\
Avg.\ question chars & 67.4 \\
\bottomrule
\end{tabular}
\caption{\textbf{PLawBench\_Civil Dataset Card.} Statistics for the PLawBench\_Civil subset used in our cross-benchmark evaluation.}
\label{tab:datacard_plawbench}
\end{table}

\noindent PLawBench~\cite{shi-etal-2026-plawbench} is a rubric-based benchmark with $\sim$850 expert-curated questions across 13 practical legal scenarios, spanning three task families (public legal consultation, practical case analysis, legal document generation) and grounded in $\sim$12{,}500 fine-grained expert rubric items. We use its \textit{Practical Case Analysis} section, where each item is structured into four components---\textit{Fact}, \textit{Law/Provision}, \textit{Reasoning}, and \textit{Conclusion}---and each component is scored against dedicated rubrics; PLawBench\_Civil is the civil-law subset obtained by filtering and randomly sampling.

\textbf{Judge models and scoring protocol.} For LexRAG, we used Qwen3-30B-A3B as the evaluator and followed the original LexRAG five-dimensional LLM-as-a-judge rubric and scoring prompt. The same evaluator configuration and prompt were applied to all compared systems. For PLawBench, we used Gemini-3.0-Pro-Preview as the evaluator and followed the scoring protocol specified in the original PLawBench paper. We did not conduct a human-agreement calibration of either judge; the reported margins on these two benchmarks should therefore be read as evidence that LEGO is competitive with the strongest RAG baselines under each benchmark's native protocol, not as a calibrated measure of superiority.

\section{Legal Expert-Domain Graph Construction}
\label{sec:appendix_graph}

\textbf{Expert-annotated legal graph}

Expert annotators were compensated at fair market rates. They were licensed legal professionals who had passed the national judicial examination, graduated from reputable law schools, and were paid at the standard hourly rate for legally trained research assistants at the authors’ institution.

The following protocol reproduces the complete instructions provided to annotators. Annotators were instructed to identify the applicable Civil Code provisions and doctrinal concepts; annotate constitutive elements, factual conditions, exceptions, defenses, and legal effects; connect them using the predefined relation schema; consult judicial interpretations and representative cases only when statutory provisions were underspecified; use mainstream doctrinal views as the default while recording disputed alternatives in notes; and resolve disagreements through expert discussion. No task-related risks beyond those ordinarily associated with scholarly legal annotation were anticipated.

\begin{enumerate}[leftmargin=*]
\item \textbf{Node extraction.} Chinese Civil Code parsed into article nodes with metadata (Part, Chapter, Section, Article number, full text); legal-concept nodes were extracted from a curated civil-law taxonomy and linked to the relevant articles.
\item \textbf{Edge annotation.} Our expert‑curated legal graph organizes all civil code concepts in a hierarchical structure (Part → Chapter → Section → Article). Each node is annotated with its definition, constitutive elements, comparisons with related provisions, and temporally ordered legal conditions and consequences. Beyond structural organization, we further label legally operative relations among concepts, including order of application, rule priority, hierarchical dependencies, prerequisites, statutory exceptions, and other doctrinal links, each accompanied by a brief legal justification.

\end{enumerate}

Civil Law ExpertGraph Construction

To construct the Civil Law ExpertGraph, we follow a hierarchical legal interpretation process inspired by source validity and doctrinal legal reasoning. The goal is not to annotate benchmark-specific answers, but to build a general-purpose normative graph that captures the structure of PRC civil law. The graph encodes legal concepts, statutory provisions, constitutive elements, exceptions, priority rules, condition–consequence relations, and legal effects. These annotations are constructed independently of LawExamQA\_Civil questions, answer labels, and official rationales.

\textbf{Hierarchical Legal Interpretation Framework}

Civil-law knowledge is organized according to the validity and interpretive function of legal sources. We use the following source hierarchy:

Civil Code Articles → Judicial Interpretations → Guiding / Reference Cases → Civil-law Textbooks and Academic Doctrines.

Civil Code articles provide the primary statutory basis. Judicial interpretations clarify the scope and application of statutory provisions. Guiding and reference cases illustrate how abstract norms operate in concrete factual settings. Civil-law textbooks and academic doctrines are used to refine conceptual boundaries, resolve doctrinal ambiguities, and standardize expert annotations.

Correspondingly, different interpretive methods are applied at different stages. Literal interpretation is used to extract concepts and elements directly from statutory text. Systematic interpretation is used to locate each provision within the structure of the Civil Code and to identify relations among provisions. Purposive and sociological interpretation are used to clarify how rules function in practice. Doctrinal and comparative interpretation are used to refine disputed or abstract civil-law concepts.

\textbf{Annotation Process}

\textbf{Stage One: Article-level extraction through literal interpretation.}

 Annotators first examine Civil Code provisions and extract basic legal units from the statutory text. These units include legal concepts, subjects, objects, constitutive elements, legal acts, conditions, exceptions, and legal effects. For example, a provision on contract validity may be decomposed into nodes such as legal act, expression of intent, capacity, legality, validity, invalidity, and revocability. At this stage, the annotation remains close to the statutory language and avoids adding case-specific reasoning.

\textbf{Stage Two: Systematic interpretation within the Civil Code.}

 Annotators then situate each provision within the broader structure of the Civil Code, including Book, Chapter, Section, and neighboring provisions. This stage identifies inter-provision relations such as general–special rule, prerequisite, limitation, exception, priority, parallel application, and condition–consequence relation. For example, rules on contract formation, contract validity, real-right transfer, liability allocation, and remedies are linked according to their doctrinal order of application rather than surface textual similarity.

\textbf{Stage Three: Refinement using judicial interpretations and representative cases.}

 Where statutory provisions are abstract or underspecified, annotators consult judicial interpretations and representative cases to clarify the practical scope of legal elements. This stage is used to refine borderline concepts, exceptions, defenses, and legal consequences. For example, judicial interpretations may clarify when a third party is protected, when a contract-related remedy is available, or when a general civil-law rule is displaced by a more specific rule.

\textbf{Stage Four: Doctrinal consolidation through civil-law theory.}

 Finally, annotators consult civil-law textbooks and academic doctrines to standardize concept boundaries and relation types. This stage is especially important for distinguishing doctrinally close concepts, such as validity versus effectiveness, contract obligation versus real-right transfer, termination versus rescission, liability for breach versus tort liability, and invalidity versus revocability. When multiple doctrinal views exist, annotators record the mainstream view as the default graph relation and preserve disputed views as notes or alternative annotations.

\textbf{Quality Control}

To ensure consistency, all annotations follow a predefined schema of node types and edge types. Node types include legal concept, statutory provision, constitutive element, factual condition, exception, defense, and legal effect. Edge types include hierarchy, prerequisite, specification, exception, priority, parallel application, condition–consequence, and remedy relation. Disagreements are resolved through expert discussion, with emphasis on whether the annotated relation reflects a general civil-law structure rather than a benchmark-specific answer path.

\textbf{Independence from Evaluation Data}

The Civil Law ExpertGraph is constructed independently from benchmark questions, answer labels, options, and official explanations. Public civil-law examination questions are used only for evaluating whether the graph-enhanced retrieval and reasoning framework improves multi-provision legal reasoning. They are not used as annotation sources for graph nodes or edges. Thus, the graph provides general normative signals for retrieval and reasoning, rather than item-level supervision.

\begin{figure*}[!tbp]
    \centering
    % Trim extra whitespace so the tree renders larger on the page.
    \includegraphics[width=\textwidth,trim=140 220 70 120,clip]{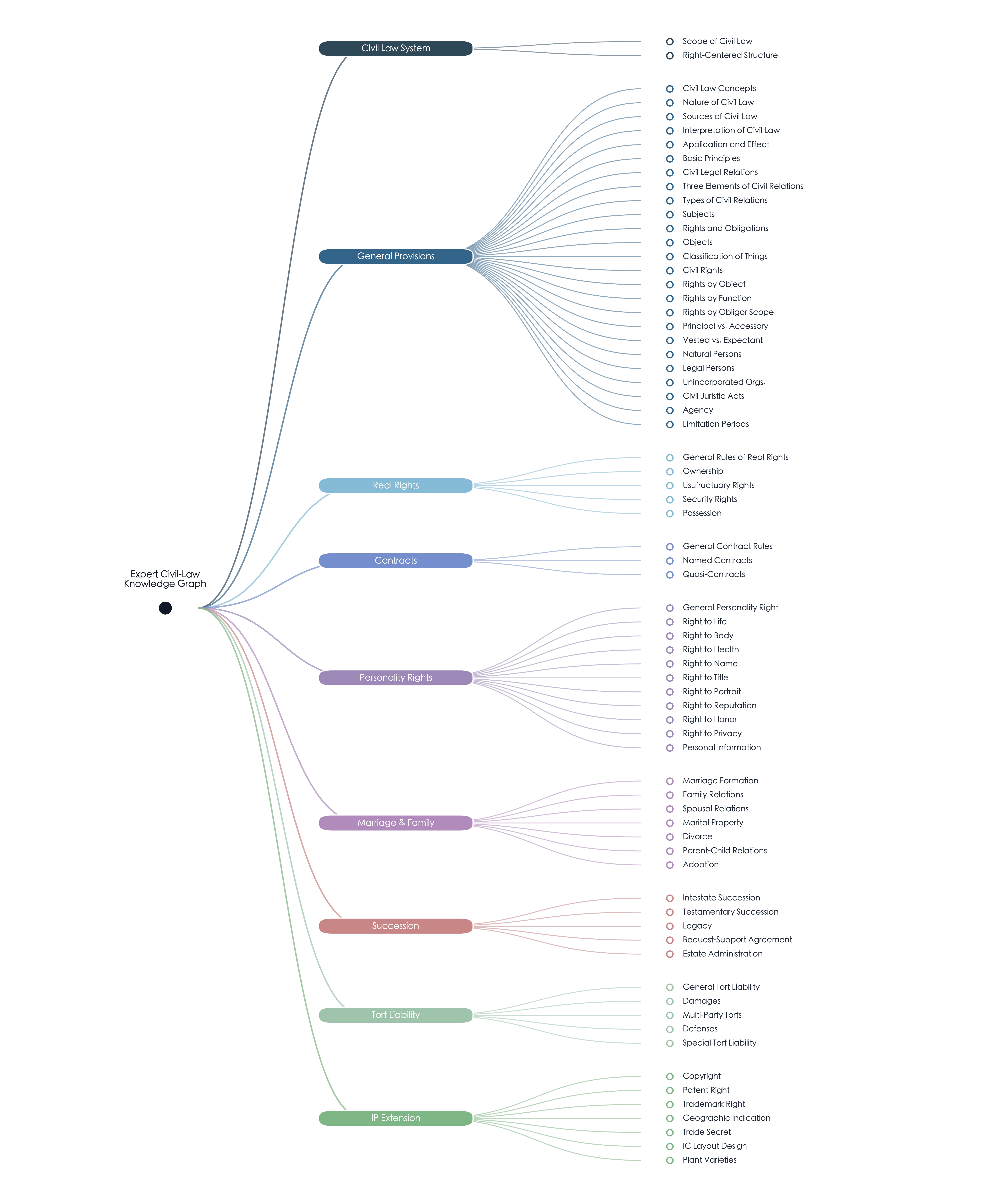}
    \caption{Legal Concept Tree of Civillaw KnowledgeGraph}
    \label{fig:legal_concept_tree}
\end{figure*}

\paragraph{Node types and error-oriented semantic links.}
Our Civil Law Legal Knowledge Graph encodes expert legal knowledge through typed nodes and semantic links. These links are represented by parent--child structures, sibling structures, and explicit semantic annotations such as prerequisite, distinction, exception, priority, and legal effect.
\begin{figure*}[!tbp]
    \centering
    \includegraphics[width=1\textwidth]{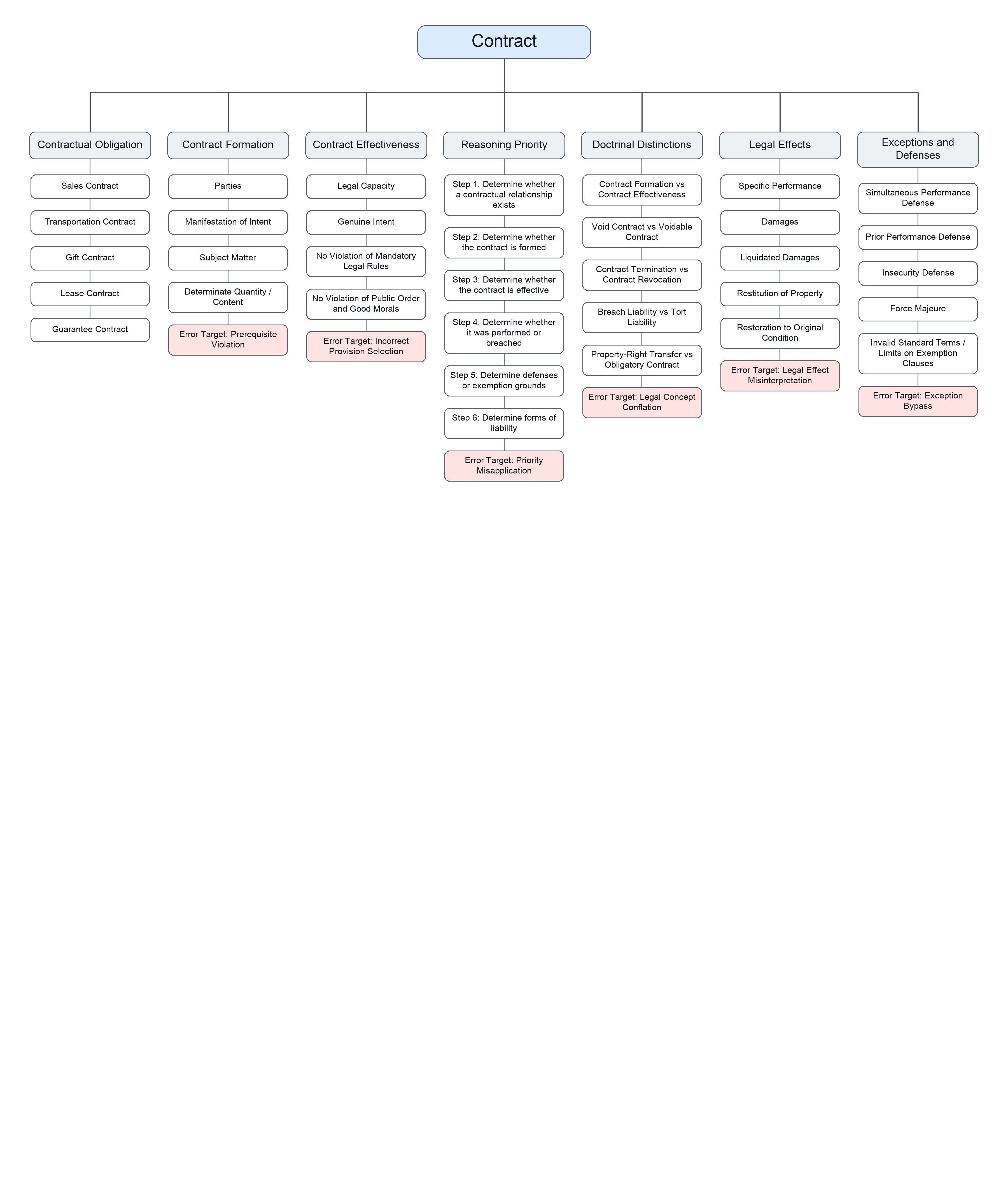}
    \caption{Legal relation in legal concept}
    \label{fig:legal_relation_concept}
\end{figure*}
\begin{table}[!htbp]
\centering
\small
\begin{tabular}{p{0.16\linewidth} p{0.34\linewidth} p{0.38\linewidth}}
\toprule
\textbf{Node Type} & \textbf{Function in the Graph} & \textbf{Targeted Error Types} \\
\midrule
\textit{Concept} 
& Identifies the legal domain or institution, e.g., contract, tort liability, property right, agency, marital property. 
& Legal Concept Conflation; Incorrect Provision Selection; Subordination Confusion. \\

\textit{Element} 
& Decomposes a rule into required legal elements or prerequisites, e.g., valid claim, manifestation of intent, registration, damage, causation. 
& Prerequisite Violation; Fact Subsumption; Reasoning Consistency; Incorrect Provision Selection. \\

\textit{Distinction} 
& Marks easily confused doctrines, e.g., voidness vs. voidability, termination vs. revocation, property right vs. creditor right. 
& Legal Concept Conflation; Legal Effect Misinterpretation; Incorrect Provision Selection. \\

\textit{Effect} 
& Encodes legal consequences after rule application, e.g., restitution, damages, revocation, invalidity, specific performance. 
& Legal Effect Misinterpretation; Reasoning Consistency; Fact Subsumption. \\

\textit{Exception} 
& Represents statutory provisos, defenses, and limiting conditions, e.g., good-faith third party, limitation period, force majeure. 
& Exception Bypass; Incorrect Provision Selection; Legal Effect Misinterpretation. \\

\textit{priority} 
& Specifies the normative order of legal analysis, e.g., relationship before claim basis, prerequisites before effects, special rules before general rules. 
& Priority Misapplication; Prerequisite Violation; Subordination Confusion; Reasoning Consistency. \\
\bottomrule
\end{tabular}
\caption{Node types in the Civil Law Legal Knowledge Graph and their corresponding error targets.}
\label{tab:node_types_errors}
\end{table}

In this design, \textit{Concept} nodes guide the model to the correct legal institution; \textit{Element} nodes force prerequisite checking; \textit{Distinction} nodes prevent confusion between similar doctrines; \textit{Exception} nodes expose provisos and defenses; \textit{Effect} nodes constrain the interpretation of legal consequences; and \textit{priority} nodes impose the proper order of legal reasoning.
\begin{figure}[H]
    \centering
    % NOTE: arXiv/pdflatex is unreliable with graphic filenames that contain spaces.
    % Rename the files to remove spaces, e.g.:
    %   Semantic allocation.jpg  -> Semantic_allocation.jpg
    %   Bar chart of legal relationtype.jpg -> Bar_chart_of_legal_relationtype.jpg
    \includegraphics[width=0.9\linewidth]{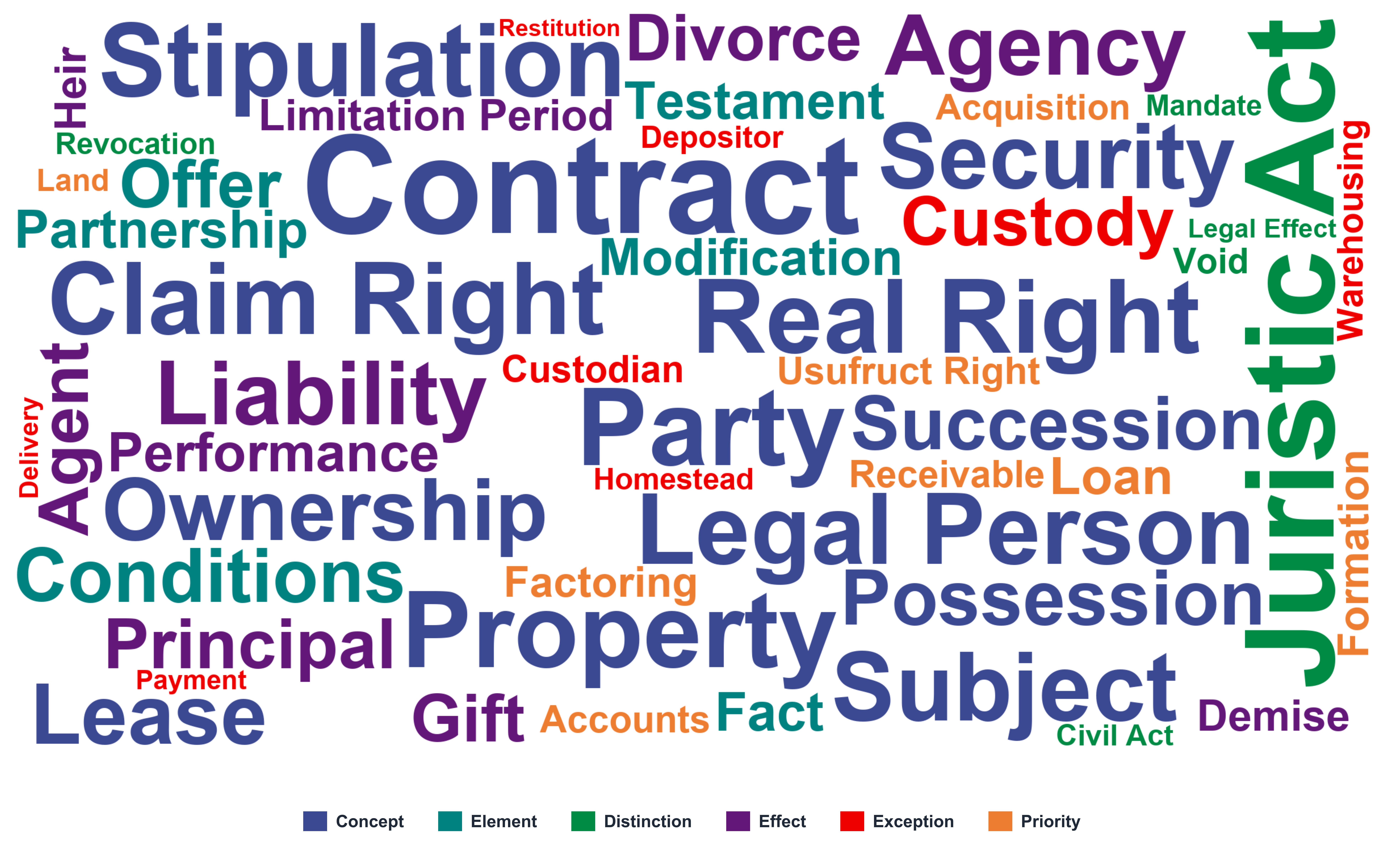}
    \caption{Semantic Cloud of legal knowledge graph}
    \label{fig:semantic_cloud}
\end{figure}
\begin{figure}[H]
    \centering
    \includegraphics[width=0.9\linewidth]{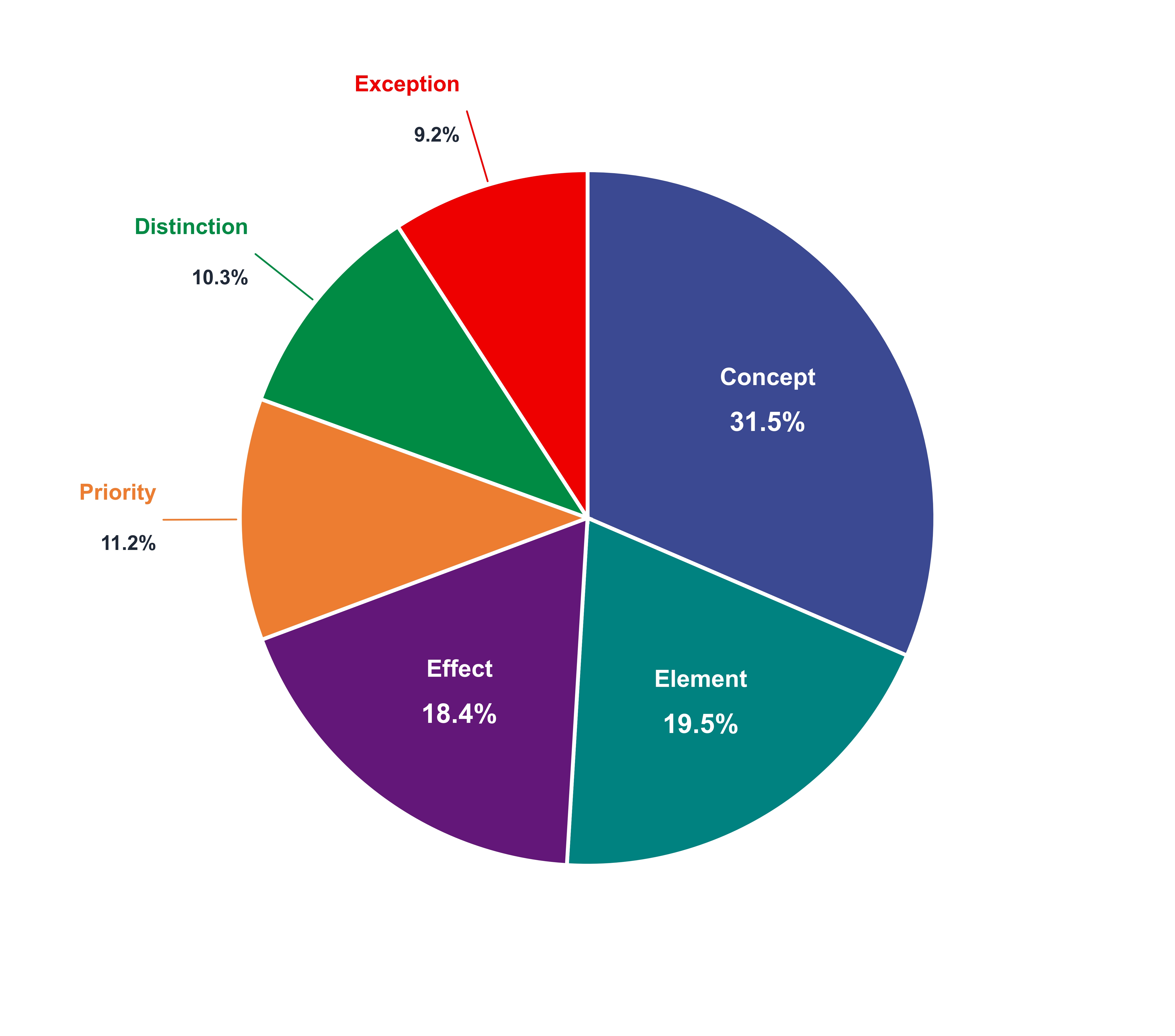}
    \caption{legal relation types of civillaw knowledgegraph}
    \label{fig:legal_relation_types}
\end{figure}

\clearpage
% Graph-source statistics appendix.
\section{Civil Law ExpertGraph: Source Statistics}
\label{sec:appendix_graph_statistics}

This appendix describes the structured Markdown corpus from which the
Civil Law ExpertGraph is constructed. These figures characterise the upstream source and are not directly comparable with the deployed-graph statistics, because the downstream construction pipeline changes the unit of analysis through rule-unit typing, vocabulary normalisation, and article linking.

\paragraph{Overall.}
The source is a single hierarchical Markdown document of roughly
391K characters over 17{,}190 lines (8{,}717 non-empty), organised as a
heading tree of nine depth levels. Its 8{,}211 heading nodes---one
document title, ten book- or major-section headings, and 8{,}200
chapter-level and deeper nodes---form the candidate knowledge units
consumed by the construction pipeline. The corpus resolves to
217 distinct Civil Code articles spanning the full range of the code
(articles 1--1260).

\paragraph{Heading-tree depth.}
Heading depth is the primary structural cue. As Table~\ref{tab:depth}
shows, the mass of the tree sits at H6--H9, where the corpus records
legal-effect statements, exception clauses, and prerequisite
enumerations---the material the pipeline maps onto normative edges.

\begin{table}[t]
\centering
\scriptsize
\setlength{\tabcolsep}{3.2pt}
\begin{adjustbox}{max width=\columnwidth}
\begin{tabular}{lrrrrrrrrr}
\toprule
Level & H1 & H2 & H3 & H4 & H5 & H6 & H7 & H8 & H9 \\
\midrule
Count & 1 & 10 & 44 & 162 & 558 & 1{,}490 & 2{,}682 & 2{,}155 & 1{,}109 \\
\bottomrule
\end{tabular}
\end{adjustbox}
\caption{Heading-depth distribution of the ExpertGraph source corpus.}
\label{tab:depth}
\end{table}

\paragraph{Major-section breakdown.}
Table~\ref{tab:sections} reports heading counts for the ten top-level
sections: a preface on the civil-law system, the seven books of the
Civil Code, a supplementary chapter on intellectual property, and a
notes section. Contracts, General Provisions, and Property Rights
together account for roughly 71.8\% of all hierarchical heading nodes,
reflecting the doctrinal density of these three books.

\begin{table}[t]
\centering
\scriptsize
\setlength{\tabcolsep}{3.2pt}
\begin{adjustbox}{max width=\columnwidth}
\begin{tabular}{lr}
\toprule
Section & Headings (H3+) \\
\midrule
Civil Law System & 44 \\
Book I General Provisions & 1{,}912 \\
Book II Property Rights & 1{,}615 \\
Book III Contracts & 2{,}362 \\
Book IV Personality Rights & 376 \\
Book V Marriage and Family & 490 \\
Book VI Succession & 410 \\
Book VII Tort Liability & 607 \\
Ch.~17 Intellectual Property & 378 \\
Notes & 6 \\
\midrule
Total & 8{,}200 \\
\bottomrule
\end{tabular}
\end{adjustbox}
\caption{Per-section structural footprint of the source corpus.}
\label{tab:sections}
\end{table}

\paragraph{Priority annotations.}
The corpus carries 357 in-text \texttt{Priority:\,N} annotations marking
doctrinally salient nodes (208 at Priority~1, 149 at Priority~2). These
annotations are recorded in the source corpus.

\paragraph{High-frequency legal concepts.}
Table~\ref{tab:concepts} reports frequencies for 22 core civil-law
concepts. The distribution makes the doctrinal focus explicit: the
corpus is anchored on contract, creditor's right, succession, property
right, agency, and tort, with substantial coverage of the machinery
connecting them---validity, transfer, formation, damages, rescission,
termination, and the validity-modality vocabulary. The presence of
procedural and defensive vocabulary (limitation period, claim right,
lien, defence) indicates that the source captures not only first-order
legal concepts but the operational apparatus used in reasoning.

\begin{table}[t]
\centering
\scriptsize
\setlength{\tabcolsep}{3.0pt}
\begin{adjustbox}{max width=\columnwidth}
\begin{tabular}{p{0.36\columnwidth}r p{0.36\columnwidth}r}
\toprule
Term & Freq. & Term & Freq. \\
\midrule
Contract & 1{,}514 & Formation & 261 \\
Creditor's right & 713 & Damages & 260 \\
Succession & 708 & Rescission & 204 \\
Property right & 568 & Termination & 202 \\
Agency & 502 & Void & 196 \\
Tort & 385 & Take effect & 180 \\
Security & 365 & Extinction & 174 \\
Mortgage & 359 & Constitutive element & 143 \\
Validity & 351 & Limitation period & 121 \\
Transfer & 276 & Claim right & 113 \\
 & & Lien & 106 \\
 & & Defence & 83 \\
\bottomrule
\end{tabular}
\end{adjustbox}
\caption{Top-22 legal concept frequencies. Counts are over the original
Chinese terms; English glosses are given for readability.}
\label{tab:concepts}
\end{table}

\paragraph{From source to deployed graph.}
The corpus above is the structured input to the construction pipeline.
The deployed ExpertGraph is produced by typing each candidate heading
node (concept / element / effect / exception-defence / relation /
article), normalising vocabulary across hierarchical levels, resolving
article references against the Civil Code text, and adding the normative
edges---general/special, principle/exception, prerequisite/consequence,
and related types---used to route queries to rule cards. The figures in this
appendix therefore characterise the corpus from which the graph is
built, not the graph itself.

\section{Detailed Error Analysis}
\label{sec:appendix_error_analysis}
\label{app:error_analysis}

Table~\ref{tab:error_analysis} reports the primary error labels for the 430 items mis-answered by the full LEGO setting. We code each item by the earliest reasoning step that makes the final answer unrecoverable. This convention is important because legal reasoning errors often cascade: a wrong provision may later produce an apparent priority conflict, or a missed prerequisite may later look like an erroneous legal effect. The analysis below therefore focuses on root causes rather than surface symptoms.

\begin{figure}[t]
\centering
\includegraphics[width=\columnwidth]{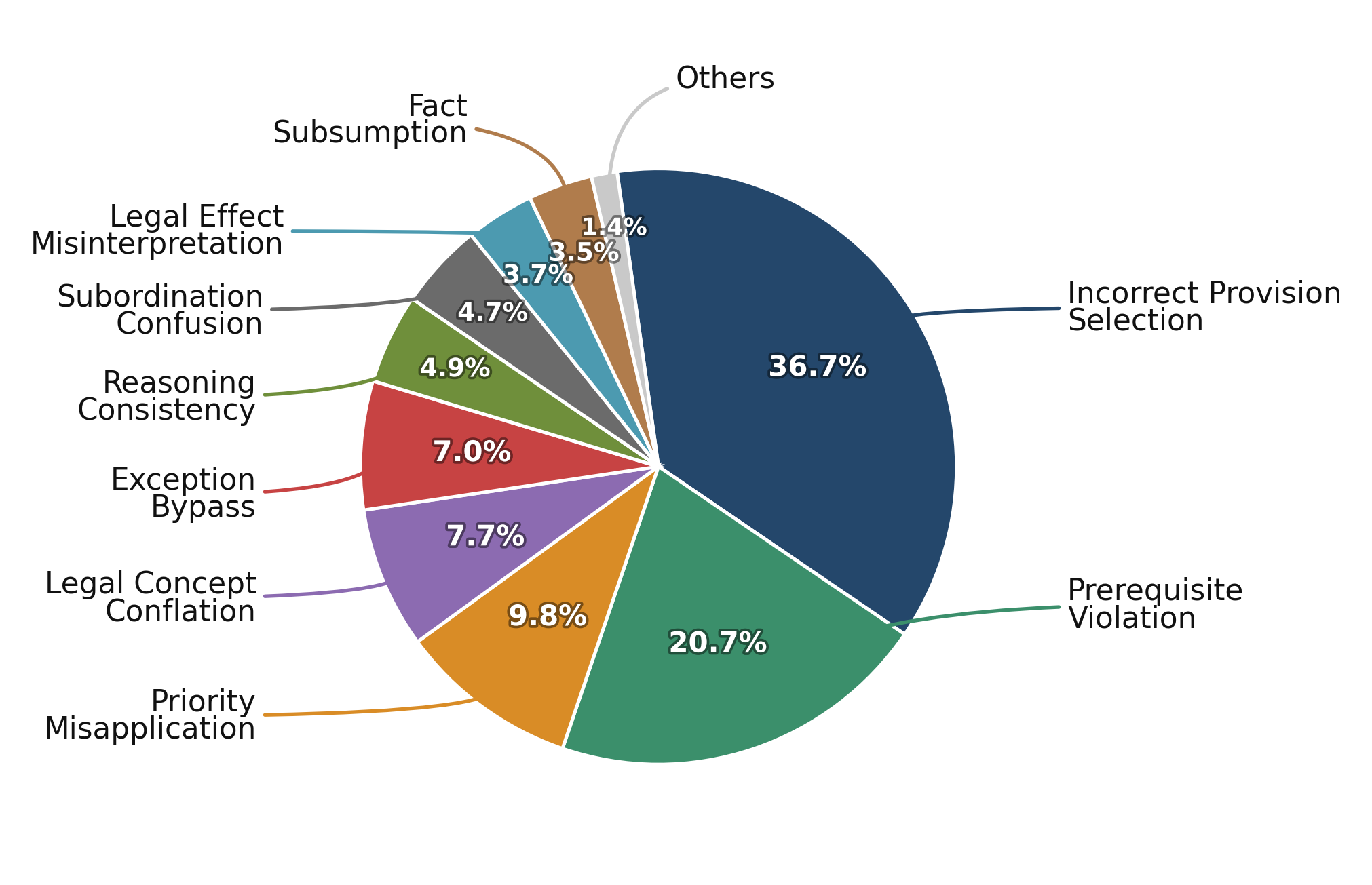}
\caption{\textbf{Error-type distribution} on items mis-answered by full LEGO. These seven failure modes motivate the ExpertCoT design choices in Section~\ref{sec:syllogistic_cot}.}
\label{fig:error_type_barchart}
\end{figure}

Figure~\ref{fig:error_type_barchart} summarises the error-type distribution. Most failures occur before final answer selection: incorrect provision selection (36.74\%) and prerequisite violation (20.70\%) together account for 57.44\% of all errors. The top seven categories account for 91.4\%, indicating that wrong answers are concentrated in a small set of structural reasoning failures rather than diffuse factual uncertainty.

\paragraph{Incorrect Provision Selection.}
This is the largest category, covering 158 errors (36.74\%). These cases usually arise when the model anchors the fact pattern to provisions that are lexically or topically similar but legally inapplicable. For example, a question involving carrier liability may trigger a nearby contractual damages provision while omitting the special transportation rule that controls the case. This error shows that retrieval is not merely a semantic matching problem: the selected provision must occupy the correct position in the legal topology and must be licensed by the facts.

\paragraph{Prerequisite Violation.}
Prerequisite violations account for 89 errors (20.70\%). In these cases, the model applies a downstream legal consequence before verifying an upstream legal state, such as discussing breach liability before establishing contract validity, or assigning tort liability before checking the required causal relation. These errors directly motivate explicit prerequisite gates in ExpertCoT: before deriving a conclusion $Q$, the model must verify every necessary premise $P_i$ encoded by prerequisite edges.

\paragraph{Priority Misapplication.}
Priority misapplication appears in 42 errors (9.77\%). The model often cites both a general rule and a special rule but applies them in the wrong order, or treats them as parallel rather than hierarchical. Such errors are especially common when civil law provisions contain nested exceptions, special chapters, or mandatory rules that override default contractual autonomy. This category motivates an explicit priority-resolution step before answer commitment.

\paragraph{Legal Concept Conflation.}
Legal concept conflation covers 33 errors (7.67\%). These errors occur when the model merges formally distinct concepts, such as prerequisite versus trigger, presumption versus legal fiction, or solidarity versus supplementary liability. Although the generated explanation may sound legally plausible, the logical operator is wrong, which changes the validity of the inference. This supports our use of the 22-concept taxonomy as typed reasoning operations rather than loose textual labels.

\paragraph{Exception Bypass.}
Exception bypass accounts for 30 errors (6.98\%). Here the model correctly identifies a general rule but fails to test whether an exception defeats it. This is a non-monotonic reasoning failure: adding an exception-bearing fact can reverse a conclusion that would otherwise follow. This motivates the exception-checking micro-constraint in ExpertCoT (\S\ref{sec:syllogistic_cot}), which requires the model to test the retrieved provisos and defences before accepting any conclusion derived from a general rule.

\paragraph{Reasoning Consistency.}
Reasoning consistency errors cover 21 cases (4.88\%). These include contradictions between intermediate conclusions and the final answer, switching parties mid-chain, or using one interpretation in the explanation and another in option selection. These errors motivate a final trace-consistency check that verifies whether the selected option is actually entailed by the stated premises.

\paragraph{Subordination Confusion.}
Subordination confusion appears in 20 errors (4.65\%). The model mistakes dependent legal relations for independent ones, such as treating accessory obligations, derivative rights, or supplementary liabilities as if they could exist without their principal legal relation. This category shows why the graph must encode structural dependence, not only topical relatedness among provisions.

\paragraph{Legal Effect Misinterpretation and Fact Subsumption.}
The remaining substantive categories are legal effect misinterpretation (16 errors, 3.72\%) and fact subsumption (15 errors, 3.49\%). The former involves deriving the wrong consequence from an otherwise correct rule, while the latter involves mapping facts to the wrong constitutive elements. These are smaller but still important categories because they occur after provision retrieval succeeds, indicating that closed-set legal reasoning still requires careful rule application.

\paragraph{Design Implications.}
The distribution suggests that ExpertCoT should be organized around four safeguards. First, a provision-grounding step reduces incorrect provision selection by forcing the model to state the controlling rule as the major premise. Second, a prerequisite-verification step blocks downstream conclusions until all necessary legal states are established. Third, a priority-and-exception step handles special rules, mandatory rules, and defeaters before the final inference. Fourth, a consistency check aligns the intermediate legal trace with the selected answer. Together, these checks target the concentrated failure mass: the top seven categories cover 91.4\% of all observed errors.

\begin{table}[t]
\centering
\small
\begin{tabular}{lcc}
\toprule
\textbf{Error Type} & \textbf{Count} & \textbf{\%} \\
\midrule
Incorrect Provision Selection  & 158 & 36.74 \\
Prerequisite Violation         & 89 & 20.70 \\
Priority Misapplication        & 42 &  9.77 \\
Legal Concept Conflation       & 33  &  7.67 \\
Exception Bypass               & 30 & 6.98 \\
Reasoning Consistency          & 21 & 4.88 \\
Subordination Confusion        & 20 & 4.65 \\
Legal Effect Misinterpretation & 16 & 3.72   \\
Fact Subsumption               & 15  &  3.49  \\
Others                         & 6 &  1.40 \\
\bottomrule
\end{tabular}
\caption{\textbf{Error-type distribution} on items mis-answered by full LEGO. These failure modes motivate the ExpertCoT design choices in Section~\ref{sec:syllogistic_cot}.}
\label{tab:error_analysis}
\end{table}

% Ablation case studies on LawExamQA\_Civil.
% Appendix-ready snippet. Requires: booktabs, adjustbox, enumitem.

\section{Case Studies for Ablation Observations}
\label{sec:appendix_ablation_case_studies}

Table~\ref{tab:ablation_case_studies} selects six representative questions
from the ablation CSVs. Each case is chosen by a deterministic correctness
pattern: LexGo Full System answers the item correctly, while the compared
pipeline(s) fail. Together, the cases illustrate why statute retrieval helps
but saturates, why ExpertCoT helps at fixed retrieval, why
ExpertGraph context helps even without ExpertCoT, and why the two components
compound in the full system.

\begin{table*}[t]
\centering
\small
\begin{adjustbox}{max width=\textwidth}
\begin{tabular}{p{2.2cm}p{0.8cm}p{1.0cm}p{4.2cm}p{4.8cm}}
\toprule
\textbf{Observation} & \textbf{QID} & \textbf{Gold} & \textbf{Key Prediction Pattern} & \textbf{What the Case Shows} \\
\midrule
Obs.~1 retrieval helps & 19 & A &
Zero-shot: B; Naive RAG + CoT: A; G-Retriever + CoT: A; LexGo Full: A &
Adding statute context corrects a no-RAG error on contract transfer legality. \\

Obs.~1 retrieval insufficient & 4 & D &
Naive RAG + CoT: B; G-Retriever + CoT: B; LexGo Full: D &
Flat retrieved statutes still over-focus on missing authority; the full system recovers the gold effect. \\

Obs.~2 ExpertCoT helps fixed retrieval & 1 & C &
Naive RAG + CoT: CD; Naive RAG + ExpertCoT: C; G-Retriever + CoT: CD; G-Retriever + ExpertCoT: C &
At the same retrieval setting, syllogistic constraints suppress an over-selected compensation option. \\

Obs.~3 ExpertGraph alone helps & 36 & B &
Naive RAG + CoT: C; G-Retriever + CoT: C; LexGo CoT: B; LexGo IRAC-CoT: B &
ExpertGraph context separates reimbursement-style legal effects from remuneration. \\

Obs.~4 component synergy & 7 & D &
Naive RAG + CoT: A; Naive RAG + ExpertCoT: A; LexGo CoT: B; LexGo Full: D &
Neither graph context alone nor ExpertCoT alone suffices; the full combination resolves the legal category. \\

Obs.~5 full system beats strongest baseline & 16 & C &
G-Retriever + ExpertCoT: D; LexGo Full: C &
The strongest non-oracle baseline over-extends unjust-enrichment return to business profits; LexGo Full limits the legal effect. \\
\bottomrule
\end{tabular}
\end{adjustbox}
\caption{\textbf{Representative ablation case studies.} Predictions are exact answer-letter outputs from the pipeline CSVs.}
\label{tab:ablation_case_studies}
\end{table*}

\paragraph{Case 1: Retrieval helps (Obs.~1, QID 19).}
\textbf{Scenario.} A Chinese company and a Korean company formed a Chinese-foreign equity joint venture. The Chinese party later attempted to transfer its contractual rights and obligations to a qualified state-owned company. The foreign party did not object in writing within three months, and the transferee participated in the venture's operation.
\textbf{Question.} Whether the transfer is illegal.
\textbf{Gold answer.} A (yes).
\textbf{Prediction pattern.} Qwen3-8B zero-shot predicts B, while Naive RAG + CoT, G-Retriever + CoT, LexGo CoT, LexGo IRAC-CoT, and LexGo Full all predict A.
\textbf{Interpretation.} The contrast supports Obs.~1's first half: statute retrieval supplies the controlling contract-transfer rules and prevents the no-RAG model from relying on commercial-efficiency or implied-consent intuition.

\paragraph{Case 2: Plain retrieval is insufficient (Obs.~1, QID 4).}
\textbf{Scenario.} A salesperson carried blank contracts stamped with the company's seal. After being removed from the company, he concealed that fact and signed a sales contract with a counterparty using the stamped form.
\textbf{Question.} How should the sales agreement be characterised when no updated authority document was shown?
\textbf{Gold answer.} D (formed and effective).
\textbf{Prediction pattern.} Naive RAG + CoT and G-Retriever + CoT both predict B, while LexGo Full predicts D.
\textbf{Interpretation.} The case illustrates Obs.~1's second half: a flat top-$k$ statute list can still lead the model to stop at ``lack of authority'', whereas the full system better tracks the legal effect attached to apparent-authority facts.

\paragraph{Case 3: ExpertCoT helps at fixed retrieval (Obs.~2, QID 1).}
\textbf{Scenario.} A principal orally retained an agent to buy timber for remuneration. After the agent spent time contacting suppliers and reporting the quotation and stock information, the principal cancelled the mandate without mentioning compensation.
\textbf{Question.} Which statement about cancellation of the mandate is correct?
\textbf{Gold answer.} C (the principal may cancel unilaterally but must compensate the agent's loss).
\textbf{Prediction pattern.} Under both Naive RAG and G-Retriever, Plain CoT predicts CD, but ExpertCoT predicts C.
\textbf{Interpretation.} With retrieval fixed, ExpertCoT removes the tempting but overbroad option D, which converts compensation for loss into full contractual remuneration. This is a direct item-level witness for Obs.~2.

\paragraph{Case 4: ExpertGraph alone contributes (Obs.~3, QID 36).}
\textbf{Scenario.} A tourist rescued a woman attempting to jump from a cliff, paid medical and incidental expenses, suffered damage to his camera and minor injury, and delayed his own schedule.
\textbf{Question.} Whether the rescuer may request remuneration from the rescued woman.
\textbf{Gold answer.} B (no remuneration may be requested).
\textbf{Prediction pattern.} Naive RAG + CoT, G-Retriever + CoT, Naive RAG + ExpertCoT, and G-Retriever + ExpertCoT predict C, while LexGo CoT, LexGo IRAC-CoT, and LexGo Full predict B.
\textbf{Interpretation.} The graph-expanded context helps distinguish reimbursement or necessary-expense consequences from a claim for remuneration. The fact that LexGo CoT succeeds already, before adding ExpertCoT, supports Obs.~3.

\paragraph{Case 5: ExpertGraph and ExpertCoT compound (Obs.~4, QID 7).}\label{app:c-qid7}
\textbf{Scenario.} A milk deliverer mistakenly put Wang's milk into Zhang's milk box. Zhang, not knowing the situation and without inspection, took it and discarded it. Wang later claimed compensation.
\textbf{Question.} How should Zhang's disposal of the milk be characterised?
\textbf{Gold answer.} D (no improper legal characterisation among the listed claims).
\textbf{Prediction pattern.} Naive RAG + CoT predicts A, Naive RAG + ExpertCoT predicts A, G-Retriever + ExpertCoT predicts B, and LexGo CoT predicts B; only LexGo Full predicts D.
\textbf{Interpretation.} This is a strict synergy case: the ExpertCoT-only branch and the ExpertGraph-only branch each fail in a different way, while the full system avoids both the unjust-enrichment and tort over-classifications.

\paragraph{Case 6: LexGo beats the strongest non-oracle baseline (Obs.~5, QID 16).}
\textbf{Scenario.} A bank teller mistakenly overpaid 10,000 yuan. The recipient knowingly used the money as business capital and earned a net profit after labour and management costs. The bank later requested return of the principal and corresponding gains.
\textbf{Question.} What should the recipient return?
\textbf{Gold answer.} C (principal plus one month's interest).
\textbf{Prediction pattern.} G-Retriever + ExpertCoT predicts D, while LexGo Full predicts C.
\textbf{Interpretation.} The strongest non-oracle baseline over-extends the unjust-enrichment consequence to all business profits. LexGo Full separates the return of unjust enrichment and interest from profits generated through subsequent business activity, supporting Obs.~5.

\clearpage

\section{Multi-Hop Case Studies on LawExamQA\_Civil}
\label{sec:appendix_multihop_case_study}
\label{app:cases} % alias label used by \appcases

This appendix complements the aggregate error-type taxonomy of Appendix~\ref{sec:appendix_error_analysis}. There we coded LEGO's $430$ mis-answered items by failure mode; here we examine the opposite population---items on which LEGO answers correctly while every one of $13$ strong baselines fails---to identify the specific reasoning patterns that drive LEGO's hop-resilience.

\paragraph{Selection.} We isolate the $20$ items on which LEGO is correct while all $13$ baselines are wrong, and select four cases spanning the four hop strata and four distinct reasoning failure patterns. For each case we report the case facts, the question, the answer distribution across the $14$ systems, and the controlling provisions.

\begin{table}[t]
\centering
\small
\begin{tabular}{lllc}
\toprule
\textbf{QID} & \textbf{Hop} & \textbf{Gold} & \textbf{LEGO-unique} \\
\midrule
627& 1-hop& ABD& \cmark \\
716 & 2-hop  & ABD & \cmark \\
544 & 3-hop  & CD  & \cmark \\
22  & 4+-hop & A   & \cmark \\
\bottomrule
\end{tabular}
\caption{Four LawExamQA\_Civil cases on which LEGO is correct and \emph{all} $13$ baselines (GPT-5, DeepSeek-V3, Qwen3-30B-A3B, GLM-4.7-Flash, Qwen3-8B, GLM-4-9B-chat, DISC-LawLLM, LegalOne, Naive RAG, HippoRAG~2, RAPTOR, G-Retriever, LightRAG) are wrong.}
\label{tab:multihop_cases}
\end{table}
\subsection{QID 7 (1-hop): Good-Faith Recipient Exception to Unjust Enrichment and Tort}
\label{app:h-qid7}

\paragraph{Case.} Because of a delivery worker's negligence, milk ordered by Wang was mistakenly placed in the milk box of Wang's neighbor Zhang. Zhang did not understand why the milk was there and, without checking, took it out and discarded it. The package bore Wang's name, address and order number. Wang discovered the missing milk the next day and demanded compensation from Zhang. (The same item is analyzed against LEGO's own ablations in Appendix~\ref{app:c-qid7}; here we compare it with the 13 external baselines.)

\paragraph{Question.} After Zhang discarded another person's milk that had been placed in his milk box, and given that Wang did suffer a property loss, how should Zhang's conduct be characterized?

\paragraph{Options.}
\begin{itemize}
  \item \textbf{A.} It constitutes unjust enrichment, because Zhang obtained a benefit without legal basis and caused a loss.
  \item \textbf{B.} It constitutes a tort, either as intentional destruction of another's property or as negligent disposal causing loss.
  \item \textbf{C.} It constitutes unauthorised agency.
  \item \textbf{D.} It involves no legal wrong.
\end{itemize}

\paragraph{Gold.} D. Civil Code Art.~\textbf{122} (unjust enrichment). Zhang did not know, and had no reason to know, that the milk was not his; he retained no benefit, because the milk no longer exists; and he neither intended nor foresaw harm to Wang. Under the good-faith-recipient limitation, a recipient who is unaware of the lack of legal basis and whose benefit has ceased to exist owes no restitution, so neither unjust enrichment nor tort liability arises.

\paragraph{Answer distribution.}
\begin{center}
\small
\begin{tabular}{@{}lp{0.70\columnwidth}@{}}
\hline
Answer & Systems \\
\hline
B  & GPT-5, DeepSeek-V3, GLM-4.7-Flash, Qwen3-8B, GLM-4-9B-chat, DISC-LawLLM, LegalOne, G-Retriever, LightRAG \\
AB & Naive RAG, HippoRAG~2, RAPTOR \\
A  & Qwen3-30B-A3B \\
D  & LEGO \checkmark \\
\hline
\end{tabular}
\end{center}

\paragraph{Baseline failure mode.} Every baseline attaches liability to Zhang. Nine choose tort alone (B), one chooses unjust enrichment alone (A), and three RAG baselines choose both (AB). The facts supply everything needed to trigger a general liability frame---another person's property, a loss, and an act by Zhang---and the baselines stop there, without testing whether the good-faith-recipient limitation removes liability.

\paragraph{LEGO's contribution.} LEGO's retrieved set contains the unjust-enrichment provisions (Arts.~122, 985--988), and its conclusion tests the general rule against the good-faith limitation before judging the options (translated from the Chinese output):
\begin{quote}
\small
Art.~987 does not apply, because he was unaware and the benefit has ceased to exist; Art.~986 applies, so Zhang bears no duty of restitution. \dots\ A is incorrect: Zhang obtained no benefit, so there is no unjust enrichment. B is incorrect: Zhang did not intentionally destroy the property. \dots\ D is correct: Zhang's conduct constitutes neither a tort nor unjust enrichment.
\end{quote}

\paragraph{Trace caveat.} LEGO rejects option~B only on the ground that there was no intentional destruction; it does not separately address the negligent-disposal branch of B.

\paragraph{Pattern.} \textsc{Exception that defeats a general liability rule}: a 1-hop item can still require testing whether a statutory limitation defeats the general rule that the facts appear to trigger. Here the general unjust-enrichment and tort frames are displaced by the good-faith-recipient limitation, which the ExpertGraph represents as an exception to restitution.

\subsection{QID 716 (2-hop): Threshold-Gated Mental-Distress Claim}
\label{ssec:lq716}

\paragraph{Case.} A and B are neighbors. A's house leaks chronically, flooding the corridor B uses daily. B repeatedly demands repair; A refuses, and later publicly mocks B (``you can't even walk past a puddle''). B develops anxiety symptoms (medical record on file); the village committee fronts RMB~1{,}000 to clear the water.

\paragraph{Question.} Among the following claims, which are correct?

\paragraph{Options.}
\begin{itemize}[leftmargin=*,topsep=2pt,itemsep=0pt]
  \item \textbf{A.} B may demand removal of nuisance.
  \item \textbf{B.} B may demand elimination of the hazard.
  \item \textbf{C.} B may demand a public apology, because the prolonged flooding caused obvious mental pressure.
  \item \textbf{D.} B may demand reimbursement of the RMB~1{,}000.
\end{itemize}

\paragraph{Gold.} \textbf{ABD}. Civil Code Arts.~\textbf{236} (real-right protection: removal of nuisance / hazard elimination) and \textbf{980} (necessary expenses incurred by a third party may be recovered). Option C requires the additional threshold of \emph{serious} mental harm under Art.~1183; the facts do not meet that bar.

\paragraph{Answer distribution.}
\noindent\begin{minipage}{\columnwidth}
\centering\small
\setlength{\tabcolsep}{3.6pt}
\renewcommand{\arraystretch}{1.05}
\begin{tabular}{@{}l p{0.78\columnwidth}@{}}
\toprule
\textbf{Answer} & \textbf{Systems} \\
\midrule
ABCD & GPT-5, DeepSeek-V3, Qwen3-30B-A3B, GLM-4.7-Flash, Qwen3-8B, DISC-LawLLM, Naive RAG, HippoRAG~2, RAPTOR, G-Retriever, LightRAG \\
ABC  & GLM-4-9B-chat \\
AB   & LegalOne \\
\textbf{ABD} & \textbf{LEGO} \\
\bottomrule
\end{tabular}
\end{minipage}

\paragraph{Baseline failure mode.} Eleven of thirteen baselines pick \emph{everything}---ABCD---failing to test option C against the constitutive-element gate of Art.~1183. ``Obvious mental pressure'' (the wording in option C) is below the statute's threshold of ``serious mental harm''; granting an apology remedy without checking this element is a classical \emph{exception bypass} (cf.\ Appendix~\ref{sec:appendix_error_analysis}, error type ``Exception Bypass'').

\paragraph{LEGO's contribution.} LEGO explicitly verifies the Art.~1183 threshold before accepting C, and rejects it:
``B's claim of prolonged mental pressure does not, on the facts, satisfy the `serious mental harm' requirement of Art.~1183, so the apology remedy is not available even though the underlying tort is established.'' Options A, B, and D each map cleanly onto a controlling provision (Art.~236 for A and B; Art.~980 for D), so LEGO accepts them.

\paragraph{Pattern.} \textsc{Threshold-gated remedy}: when a remedy depends on satisfying a constitutive element (here, ``serious'' mental harm), explicitly check the element against the facts before granting the remedy. The ExpertCoT major-premise expansion makes the threshold visible; baselines that skip the check end up over-claiming.

\subsection{QID 544 (3-hop): Joint Tort vs.\ Separate Tort with Divisibility}
\label{ssec:lq544}

\paragraph{Case.} A and B are neighbors, each keeping one goat. The two goats are known to roam together. One morning both pens are left unlocked; the goats together eat C's rare medicinal herbs bare. The damage scene shows interleaved hoofprints and mixed eating marks; the share consumed by each goat cannot be distinguished. A and B had a verbal agreement to take turns supervising, but neither honored it on the day.

\paragraph{Question.} Regarding A's and B's liability:

\paragraph{Options.}
\begin{itemize}[leftmargin=*,topsep=2pt,itemsep=0pt]
  \item \textbf{A.} A and B may each escape liability by proving they discharged the duty of care.
  \item \textbf{B.} Under Civil Code Art.~1168, A and B bear joint and several liability for the jointly-committed tort.
  \item \textbf{C.} If the quantity eaten by each goat can be determined, A and B bear corresponding several liability.
  \item \textbf{D.} If the quantity cannot be determined, A and B bear liability in equal shares.
\end{itemize}

\paragraph{Gold.} \textbf{CD}. Civil Code Arts.~\textbf{1171, 1172, 1245}. The two goat-owners did not jointly commit a tort under Art.~1168; their tortious acts are \emph{separate}, with damage that happens to combine (Art.~1172). When the divisible share can be proved, each bears proportional liability; when it cannot, the default is equal apportionment.

\paragraph{Answer distribution.}
\noindent\begin{minipage}{\columnwidth}
\centering\small
\setlength{\tabcolsep}{3.6pt}
\renewcommand{\arraystretch}{1.05}
\begin{tabular}{@{}l p{0.78\columnwidth}@{}}
\toprule
\textbf{Answer} & \textbf{Systems} \\
\midrule
BCD & GPT-5, DeepSeek-V3, Qwen3-8B, LegalOne, Naive RAG, HippoRAG~2, RAPTOR, G-Retriever, LightRAG \\
BD  & Qwen3-30B-A3B, GLM-4-9B-chat \\
BC  & GLM-4.7-Flash \\
D   & DISC-LawLLM \\
\textbf{CD} & \textbf{LEGO} \\
\bottomrule
\end{tabular}
\end{minipage}

\paragraph{Baseline failure mode.} Nine of thirteen baselines pick BCD: they retain options C and D (the correct divisibility logic under Art.~1172) but \emph{also} pick B (joint tort under Art.~1168). This is internally inconsistent---Art.~1168's joint liability is incompatible with Art.~1172's proportional liability---but baselines do not detect the conflict because they retrieve and apply each option's apparent statute independently. The error reflects what Appendix~\ref{sec:appendix_error_analysis} labels \emph{priority misapplication}: when a general rule (joint tort) and a special rule (separate-act tort with divisibility) both surface in retrieval, baselines treat them as parallel rather than hierarchical.

\paragraph{LEGO's contribution.} LEGO's response explicitly distinguishes three statutes: Art.~1168 (joint tort, requires concerted action), Art.~1171 (concurrent dangerous acts, each act alone sufficient to cause the entire damage), and Art.~1172 (separate-act tort, damage combines). It then applies the facts to Art.~1172 specifically (acts are separate; damage is combined and only conditionally divisible), and rejects Art.~1168 because the verbal supervision agreement was not actually executed---there is no concerted action. The conclusion CD follows from Art.~1172's two-branch structure (provable share $\to$ proportional; unprovable $\to$ equal).

\paragraph{Pattern.} \textsc{General-vs-special selection among tort regimes}: ExpertGraph's normative edges (joint vs.\ separate vs.\ concurrent-dangerous) gate which liability rule applies, preventing the silent merger of Art.~1168 with Art.~1172 that traps every baseline.

\subsection{QID 22 (4+-hop): Independent Mortgage Transfer (Art.~406 vs.\ Art.~407)}
\label{ssec:lq22}

\paragraph{Case.} D holds a mortgage over C's house. To secure a friend's bank loan, D signs a written agreement transferring \emph{the mortgage right alone} (not the underlying debt) to the bank. The transfer is duly registered; the registry lists the bank as the new mortgagee. The bank, relying on the registry's public-faith principle, accepts the mortgage as security and disburses the loan.

\paragraph{Question.} Given that the bank has completed registration and relied in good faith, is the mortgage-transfer contract unlawful?

\paragraph{Options.}
\begin{itemize}[leftmargin=*,topsep=2pt,itemsep=0pt]
  \item \textbf{A.} Yes.
  \item \textbf{B.} No --- the mortgage is a freely transferable property right; registration completes the property-right transfer, and the bank has lawfully acquired the mortgage.
  \item \textbf{C.} No --- effective upon notice to the debtor.
  \item \textbf{D.} No --- effective upon C's consent.
\end{itemize}

\paragraph{Gold.} \textbf{A}. Civil Code Arts.~\textbf{153, 216, 217, 407}. Art.~407 is the controlling special rule: a mortgage right cannot be transferred independently of the underlying debt; the general transferability and registration-effect rules (Arts.~216, 217) are overridden in this configuration.

\paragraph{Answer distribution.}
\noindent\begin{minipage}{\columnwidth}
\centering\small
\setlength{\tabcolsep}{3.6pt}
\renewcommand{\arraystretch}{1.05}
\begin{tabular}{@{}l p{0.78\columnwidth}@{}}
\toprule
\textbf{Answer} & \textbf{Systems} \\
\midrule
B & \emph{All $13$ baselines}: GPT-5, DeepSeek-V3, Qwen3-30B-A3B, GLM-4.7-Flash, Qwen3-8B, GLM-4-9B-chat, DISC-LawLLM, LegalOne, Naive RAG, HippoRAG~2, RAPTOR, G-Retriever, LightRAG \\
\textbf{A} & \textbf{LEGO} \\
\bottomrule
\end{tabular}
\end{minipage}

This is the most striking unanimous-baseline-failure case in the dataset: every closed-model baseline (including GPT-5 and DeepSeek-V3), every law-tuned LLM, and every RAG baseline picks B.

\paragraph{Baseline failure mode.} Baselines apply the general rule (Arts.~216/217: registration completes real-property right transfer; public-faith protection for good-faith reliance on the registry) without checking whether a more specific rule constrains the rule's domain. Art.~407, located in the mortgage chapter, explicitly removes \emph{independent} mortgage transfers from the general transferability regime. The error is a textbook \emph{priority misapplication}: the general rule is correctly identified but the special rule that overrides it is silently omitted.

\paragraph{LEGO's contribution.} LEGO's response makes the rule hierarchy explicit:
``Art.~406 is the general rule (transferability of the mortgaged property with the mortgage following); Art.~407 is the special rule (the mortgage right cannot be separated from the secured debt); the special rule prevails.'' Applied to the facts (only the mortgage was transferred; the secured debt was not), the contract is unlawful and the registration cannot cure the substantive invalidity. Note that the public-faith principle (Art.~216) protects \emph{procedural} reliance on the registry, not the substantive validity of an act prohibited by Art.~407---this nested distinction is what every baseline misses.

\paragraph{Pattern.} \textsc{General-vs-special rule resolution at the 4-hop level}: the case requires composing four provisions (153 invalidity, 216 public faith, 217 registration effect, 407 mortgage-debt inseparability) and resolving their priority. Without an explicit rule-hierarchy step---which ExpertCoT enforces and ExpertGraph supports with normative edges---retrieval that surfaces only Arts.~216/217 yields a confident but wrong answer.

\subsection{Cross-Case Synthesis}

The four cases map onto four reasoning failure patterns that ExpertGraph and ExpertCoT are specifically designed to address. Three of the four also appear among the top error types in Appendix~\ref{sec:appendix_error_analysis}'s aggregate taxonomy (the analysis of items on which LEGO \emph{fails}), confirming that LEGO's wins and losses share a common skeleton: success requires explicit rule-hierarchy resolution and constitutive-element checking, while failure occurs when those steps are skipped or applied to the wrong rule.

\begin{center}
\small
\setlength{\tabcolsep}{3pt}
\renewcommand{\arraystretch}{1.08}

\begin{tabularx}{\columnwidth}{
    @{}
    >{\raggedright\arraybackslash}X
    >{\raggedright\arraybackslash}p{1.25cm}
    >{\raggedright\arraybackslash}p{2.45cm}
    @{}
}
\toprule
\textbf{Reasoning pattern}
& \textbf{Case}
& \textbf{Taxonomy analogue} \\
\midrule

Right-holder disambiguation among topically related parties
& QID 174
& Legal Concept Conflation \\

Threshold-gated remedy (constitutive-element check before grant)
& QID 716
& Exception Bypass \\

General-vs-special selection among tort regimes
& QID 544
& Priority Misapplication \\

General-vs-special rule resolution at $4$-hop
& QID 22
& Priority Misapplication / Incorrect Provision Selection \\

\bottomrule
\end{tabularx}

\end{center}

% -------改之前----------

% \begin{center}\small
% \begin{tabular}{p{3.7cm}p{1.5cm}p{2.5cm}}
% \toprule
% \textbf{Reasoning pattern} & \textbf{Case} & \textbf{Aggregate-taxonomy analogue} \\
% \midrule
% Right-holder disambiguation among topically related parties & QID 174 & Legal Concept Conflation \\
% Threshold-gated remedy (constitutive-element check before grant) & QID 716 & Exception Bypass \\
% General-vs-special selection among tort regimes & QID 544 & Priority Misapplication \\
% General-vs-special rule resolution at $4$-hop & QID 22 & Priority Misapplication / Incorrect Provision Selection \\
% \bottomrule
% \end{tabular}
% \end{center}

Two observations follow. First, the unanimous-baseline-failure pattern on QID 22 is not noise: Art.~407 is doctrinally required but lexically remote from the case facts, so similarity-driven retrieval does not surface it, whereas rule-card routing does. Second, the $4{+}$-hop margin of $+6.4$~pp over the strongest LLM baseline is not driven by retrieving more provisions, but by retrieving the right \emph{configuration} of provisions and resolving their hierarchy. The aggregate hop-resilience of LEGO (Table~\ref{tab:main}) is therefore a population-level signature of the per-case mechanisms documented in this appendix.
\clearpage

\section{Component-Synergy Case Studies on LawExamQA\_Civil}
\label{sec:appendix_synergy_case_study}
\label{app:synergy} % alias label used by appendix_statistical_validation.tex

This appendix isolates the internal contribution of the two LEGO components --- the ExpertGraph retrieval module and the ExpertCoT reasoning module --- by examining a $5 \times 3$ ablation grid and the case-level behaviour at its extremes. It complements Appendix~\ref{sec:appendix_multihop_case_study} (cases where LEGO beats external baselines) and Appendix~\ref{sec:appendix_error_analysis} (aggregate error taxonomy on LEGO failures): together the three appendices document, respectively, what LEGO adds over baselines, what LEGO adds over its own ablations, and where LEGO still fails.

\paragraph{Setup.} We instantiate twelve configurations crossing five retrieval modules (none, Naive RAG, G-Retriever, ExpertGraph, Full) with three reasoning modules (Plain CoT, IRAC-CoT, ExpertCoT). Twelve of the fifteen cells were run; the missing three are not informative for our hypothesis.

\paragraph{Accuracy matrix.} Table~\ref{tab:ablation_accuracy} reports per-configuration accuracy on each hop stratum. Three patterns are visible. First, swapping Naive RAG for ExpertGraph (with the same Plain CoT) yields a small $+1.2$~pp gain on the aggregate but a $+2.3$~pp gain on $1$-hop. Second, swapping Plain CoT for ExpertCoT (with the same Naive RAG) yields a $+1.5$~pp gain on the aggregate but a $+7.5$~pp gain on $2$-hop. Third, combining both yields $+6.2$~pp on the aggregate and $+7.6$~pp on $1$-hop --- more than the sum of the two single-component gains.

\begin{table}[t]
\centering
\small
\setlength{\tabcolsep}{4.0pt}
\begin{tabular}{lrrrrr}
\toprule
\textbf{Configuration} & \textbf{1h} & \textbf{2h} & \textbf{3h} & \textbf{4+h} & \textbf{All} \\
\midrule
Qwen3-8B (zero-shot)            & 25.4 & 27.4 & 26.9 & 16.1 & 25.4 \\
Qwen3-8B + CoT                   & 28.9 & 27.0 & 31.7 & 33.9 & 29.2 \\
Qwen3-8B + IRAC-CoT              & 30.1 & 34.4 & 26.0 & 33.9 & 31.1 \\
\midrule
Naive RAG + CoT                  & 34.2 & 33.0 & 35.6 & 37.1 & 34.3 \\
Naive RAG + IRAC-CoT             & 32.5 & 31.2 & 37.5 & 25.8 & 32.2 \\
G-Retriever + CoT                & 34.2 & 34.0 & 30.8 & 35.5 & 33.7 \\
G-Retriever + IRAC-CoT           & 33.3 & 34.9 & 27.9 & 38.7 & 33.5 \\
\midrule
Naive RAG + ExpertCoT            & 33.3 & 40.5 & 34.6 & 35.5 & 35.8 \\
G-Retriever + ExpertCoT          & 34.8 & 38.1 & 32.7 & 35.5 & 35.5 \\
\midrule
ExpertGraph + CoT                & 36.5 & 33.5 & 37.5 & 33.9 & 35.5 \\
ExpertGraph + IRAC-CoT           & 36.5 & 39.1 & 35.6 & 38.7 & 37.3 \\
\textbf{Full system (LEGO)}      & \textbf{41.8} & \textbf{40.5} & \textbf{37.5} & \textbf{38.7} & \textbf{40.5} \\
\bottomrule
\end{tabular}
\caption{Accuracy (\%) on LawExamQA\_Civil by hop count for twelve component configurations. Single-component upgrades give modest gains; combining ExpertGraph with ExpertCoT yields a gain.}
\label{tab:ablation_accuracy}
\end{table}

\paragraph{Synergy decomposition.} Taking Naive~RAG + Plain CoT as the baseline ($34.30\%$), we isolate the contribution of each LEGO component and their interaction:
\begin{itemize}[leftmargin=*,topsep=2pt,itemsep=1pt]
  \item ExpertGraph retrieval alone (replacing Naive RAG, keeping Plain CoT): $35.55\%$, $\bm{+1.24}$~\textbf{pp}.
  \item ExpertCoT alone (keeping Naive RAG, replacing Plain CoT): $35.82\%$, $\bm{+1.52}$~\textbf{pp}.
  \item Both together (Full): $40.53\%$, $\bm{+6.22}$~\textbf{pp}.
  \item \textbf{Synergy} : $6.22 - (1.24 + 1.52) = \bm{+3.46}$~\textbf{pp}.
\end{itemize}
The interaction is positive in point estimate, but its 95\% interval includes zero (Appendix~L).

\paragraph{Case selection.} We then identify items where the Full system answers correctly while \emph{every one of the eleven ablations} (including the strong ExpertGraph~+~IRAC-CoT variant and the two ExpertGraph alternatives) is incorrect. Across the $723$ items there are $14$ such cases (Table~\ref{tab:ablation_full_unique}). We expand three of them spanning the $2$/$3$/$4{+}$-hop strata.

\begin{table}[t]
\centering
\small
\begin{tabular}{lrr}
\toprule
\textbf{Hop} & \textbf{Items} & \textbf{Full-unique-correct} \\
\midrule
1-hop  & 342 & 5 \\
2-hop  & 215 & 6 \\
3-hop  & 104 & 1 \\
4+-hop &  62 & 2 \\
\midrule
Total  & 723 & 14 \\
\bottomrule
\end{tabular}
\caption{Items on which the Full LEGO system is correct and every one of the eleven ablations is incorrect.}
\label{tab:ablation_full_unique}
\end{table}

\subsection{QID 237 (2-hop): Constitutum Possessorium}
\label{ssec:ab237}

\paragraph{Case.} A owns a jade worth RMB~10{,}000. A and B sign a sale contract for RMB~11{,}000, ``effective upon signature''; B is to take delivery in three days. After signing, A asks to keep the jade for personal enjoyment a few days more; B agrees but no separate property-transfer document is signed. The jade remains in A's physical custody. The next day A re-sells and delivers the jade to a knowing buyer C, and the chain continues to D, E (a finder after loss), F.

\paragraph{Question.} Given that the sale contract is effective and B has confirmed ownership at signature, which statements about B's acquisition of ownership and delivery are correct?

\paragraph{Options.}
\begin{itemize}[leftmargin=*,topsep=2pt,itemsep=0pt]
  \item \textbf{A.} B acquires ownership upon contract validity.
  \item \textbf{B.} B acquires ownership when the borrowing agreement takes effect.
  \item \textbf{C.} Because there is no actual delivery, ownership has not transferred.
  \item \textbf{D.} A delivered the jade to B by means of \emph{constitutum possessorium}.
\end{itemize}

\paragraph{Gold.} \textbf{BD}. Civil Code Arts.~\textbf{224} (general rule: ownership of movables transfers upon delivery) and \textbf{228} (constitutum possessorium: when the parties agree the transferor continues to possess, ownership transfers at the moment the agreement takes effect). Art.~228 is the controlling \emph{special} rule.

\paragraph{Ablation behaviour.} The agreement to ``borrow back'' the jade after the sale is the textbook \emph{constitutum possessorium} pattern that activates Art.~228; eight of eleven ablations select CD --- correctly rejecting A but \emph{also} rejecting B and accepting C. They thus apply the general delivery rule (Art.~224) and miss the special rule. Even the strongest ablation, ExpertGraph + IRAC-CoT, retrieves both Arts.~228 and 224 but still answers CD: retrieval is sufficient but reasoning fails to resolve their hierarchy. Only the Full system (ExpertGraph + ExpertCoT) explicitly invokes the general-vs-special priority and selects BD.

\paragraph{Per-configuration outcomes.}
\begin{center}\small
\begin{tabular}{llc}
\toprule
\textbf{Configuration} & \textbf{Pred} & \textbf{OK} \\
\midrule
Qwen3-8B (zero-shot)       & AD  & \xmark \\
Qwen3-8B + CoT             & CD  & \xmark \\
Qwen3-8B + IRAC-CoT        & CD  & \xmark \\
Naive RAG + CoT            & C   & \xmark \\
Naive RAG + IRAC-CoT       & CD  & \xmark \\
G-Retriever + CoT          & CD  & \xmark \\
G-Retriever + IRAC-CoT     & CD  & \xmark \\
Naive RAG + ExpertCoT      & CD  & \xmark \\
G-Retriever + ExpertCoT    & CD  & \xmark \\
ExpertGraph + CoT          & C   & \xmark \\
ExpertGraph + IRAC-CoT     & CD  & \xmark \\
\textbf{Full (LEGO)}       & \textbf{BD} & \cmark \\
\bottomrule
\end{tabular}
\end{center}

\paragraph{Full-system trace (excerpt).}
\begin{quote}\itshape\small
Art.~228 is the special rule and overrides Art.~224's general delivery requirement. Since A and B agreed that A would continue to possess the jade after the sale, ownership transfers at the moment of the agreement (B), not upon physical delivery (rejecting C). The delivery itself takes the form of \emph{constitutum possessorium} (D).
\end{quote}

\paragraph{What the ablation isolates.} ExpertGraph + IRAC-CoT receives both articles but does not apply their general--special priority; the Full system's P--F--C prompt, which asks for provision relations explicitly, produces the correct ordering.

\subsection{QID 509 (3-hop): Title-Retention Sale plus Good-Faith Third Party}
\label{ssec:ab509}

\paragraph{Case.} Zhou sells a computer to Wu under a title-retention contract for RMB~6{,}000 in five monthly instalments of RMB~1{,}200; until full payment, title remains with Zhou. Wu pays the first four instalments on time but defaults on the fifth. The computer malfunctions; Wu hands it to Zhou for repair. After repair Zhou, citing Wu's breach, re-sells the computer to Wang for RMB~6{,}200; Wang checks Zhou's original purchase receipt and accepts in good faith.

\paragraph{Question.} Based on the title-retention rules, which statements are correct?

\paragraph{Options.}
\begin{itemize}[leftmargin=*,topsep=2pt,itemsep=0pt]
  \item \textbf{A.} Wang may acquire ownership of the computer.
  \item \textbf{B.} Because title remains with Zhou, Zhou may exercise the right of recovery when Wu cannot pay the final instalment.
  \item \textbf{C.} If Wu's unpaid due amount reaches RMB~1{,}800, Zhou may demand the full purchase price.
  \item \textbf{D.} If Wu's unpaid due amount reaches RMB~1{,}800, Zhou may rescind the contract and demand a use fee for the computer.
\end{itemize}

\paragraph{Gold.} \textbf{ACD}. Civil Code Arts.~\textbf{311} (good-faith acquisition), \textbf{634} (one-fifth default-acceleration rule in instalment sales), \textbf{642} (seller's right of recovery under title retention, conditional on notice and reasonable period). The case requires composing three rules: B is incorrect because the right of recovery requires \emph{notice plus reasonable period} (Art.~642), not just inability to pay; A is correct via Art.~311 (Wang's good-faith reliance on the receipt); C and D both follow from Art.~634 (1/5 threshold of RMB~1{,}200 is exceeded by RMB~1{,}800).

\paragraph{Ablation behaviour.} The ablations fragment across the four options because each lacks one of the three doctrinal pieces:
\begin{itemize}[leftmargin=*,topsep=2pt,itemsep=1pt]
  \item Six of eleven ablations select \textbf{AD} (good-faith acquisition + rescission), missing Art.~634's full-price branch (C).
  \item Naive~RAG~+~ExpertCoT selects \textbf{BCD}, retaining the wrong recovery-right framing (B).
  \item Two G-Retriever variants select \textbf{AB} (mixing good-faith acquisition with the unconditional recovery framing).
\end{itemize}
ExpertGraph + IRAC-CoT retrieves Arts.~642 and 634 but still answers AD; the Full system is the only configuration that retrieves the same articles \emph{and} composes them into the three-rule chain (Art.~311 for A; Art.~634 for both C and D, distinguishing its two branches; Art.~642 as the gate that rejects B).

\paragraph{Per-configuration outcomes.}
\begin{center}\small
\begin{tabular}{llc}
\toprule
\textbf{Configuration} & \textbf{Pred} & \textbf{OK} \\
\midrule
Qwen3-8B (zero-shot)       & ABCD & \xmark \\
Qwen3-8B + CoT             & A    & \xmark \\
Qwen3-8B + IRAC-CoT        & AD   & \xmark \\
Naive RAG + CoT            & AD   & \xmark \\
Naive RAG + IRAC-CoT       & AD   & \xmark \\
G-Retriever + CoT          & AD   & \xmark \\
G-Retriever + IRAC-CoT     & AB   & \xmark \\
Naive RAG + ExpertCoT      & BCD  & \xmark \\
G-Retriever + ExpertCoT    & AB   & \xmark \\
ExpertGraph + CoT          & AD   & \xmark \\
ExpertGraph + IRAC-CoT     & AD   & \xmark \\
\textbf{Full (LEGO)}       & \textbf{ACD} & \cmark \\
\bottomrule
\end{tabular}
\end{center}

\paragraph{What the ablation isolates.} This case demonstrates the limit of single-component upgrades. With ExpertGraph retrieval alone, the model has access to Art.~634 but does not unpack its two-branch structure (full-price acceleration \emph{vs.} rescission-plus-use-fee); with ExpertCoT alone over Naive retrieval, the model develops the right reasoning skeleton but misses the controlling articles. The Full system both retrieves and decomposes, recovering all three of A, C, and D.

\subsection{QID 652 (4+-hop): Will Validity vs.\ Mandatory Reservation for a Fetus}
\label{ssec:ab652}

\paragraph{Case.} Zhou and Wu, married for years and childless, signed an artificial-insemination consent at a hospital; Wu became pregnant. In April 2021, Zhou (now hospitalised with cancer) wrote and signed a handwritten holographic will: ``All housing purchased after marriage is inherited by my parents.'' The will is signed and dated, read aloud to the parents, and otherwise satisfies every form requirement of Civil Code Art.~1134. Zhou died in May 2021; Wu gave birth in October 2021. Zhou's parents immediately listed the house for sale.

\paragraph{Question.} Given that the will reflects Zhou's genuine intent and is formally complete, which of the following statements are \emph{incorrect}?

\paragraph{Options.}
\begin{itemize}[leftmargin=*,topsep=2pt,itemsep=0pt]
  \item \textbf{A.} Because artificial insemination raises ethical issues, the AI consent is void as contrary to public order and good morals.
  \item \textbf{B.} Being a formally complete and genuinely intended holographic will, Zhou's will is fully valid.
  \item \textbf{C.} Because Wu's child was conceived by artificial insemination rather than naturally, the child is not a marital child of Zhou and Wu.
  \item \textbf{D.} When dividing Zhou's estate, an inheritance share must be reserved for the fetus in Wu's womb.
\end{itemize}

\paragraph{Gold.} \textbf{ABC}. Civil Code Arts.~\textbf{153} (juristic-act validity), \textbf{1071} (parent-child relationship covers AI children), \textbf{1153} (inheritance division), \textbf{1155} (\emph{mandatory share reservation for a fetus}). Statement D is the only correct one; A, B, and C are each separately wrong. The non-obvious wrong is B: a will may be formally complete \emph{and} substantively over-broad if it disposes of estate without reserving the share that Art.~1155 makes mandatory for a conceived but unborn heir.

\paragraph{Ablation behavior.} All eleven ablations fail to select \textbf{B}: ten answer \textbf{AC} and Naive RAG + IRAC-CoT answers \textbf{ACD}. They correctly reject A (AI consent is not contra public order) and C (AI children are marital children under Art.~1071) but accept B: they treat ``formally complete'' as a sufficient condition for ``fully valid,'' silently overlooking the mandatory-reservation requirement under Art.~1155. The error pattern is \emph{exception bypass} (cf.\ Appendix~\ref{sec:appendix_error_analysis}): even after retrieving the form-validity provisions, the ablations fail to test the will against the substantive limit imposed by Art.~1155 on every disposition of an estate when a heir is in gestation. The Full system, which receives the same ExpertGraph retrieval as the two ExpertGraph ablations, is the only configuration that applies Art.~1155 to option B.

\paragraph{Per-configuration outcomes.}
\begin{center}\small
\begin{tabular}{llc}
\toprule
\textbf{Configuration} & \textbf{Pred} & \textbf{OK} \\
\midrule
Qwen3-8B (zero-shot)       & AC  & \xmark \\
Qwen3-8B + CoT             & AC  & \xmark \\
Qwen3-8B + IRAC-CoT        & AC  & \xmark \\
Naive RAG + CoT            & AC  & \xmark \\
Naive RAG + IRAC-CoT       & ACD & \xmark \\
G-Retriever + CoT          & AC  & \xmark \\
G-Retriever + IRAC-CoT     & AC  & \xmark \\
Naive RAG + ExpertCoT      & AC  & \xmark \\
G-Retriever + ExpertCoT    & AC  & \xmark \\
ExpertGraph + CoT          & AC  & \xmark \\
ExpertGraph + IRAC-CoT     & AC  & \xmark \\
\textbf{Full (LEGO)}       & \textbf{ABC} & \cmark \\
\bottomrule
\end{tabular}
\end{center}

\paragraph{Full-system trace (excerpt).}
\begin{quote}\itshape\small
Art.~1134 governs the formal validity of a holographic will; Art.~1155 is the special rule that requires a reservation for an in-gestation heir. The two are not mutually exclusive: a will may pass formal validity (under Art.~1134) but fail substantive validity because it disposes of the entire estate without reserving the mandatory fetus share. The disposition to Zhou's parents leaves nothing for Wu's unborn child, violating Art.~1155.
\end{quote}

\paragraph{What the ablation isolates.} Both components are necessary on this case. ExpertGraph alone (with Plain CoT or IRAC-CoT) receives the same provisions, including Art.~1155, but does not apply it as a substantive limit on the will's scope; ExpertCoT alone (with Naive or G-Retriever retrieval) supplies the constitutive-element checking framework but lacks the article. Only the Full system applies Art.~1155 through the element check that connects ``unborn fetus'' to the reservation requirement.

\subsection{Cross-Case Synthesis}

The three ablation cases instantiate a common pattern: each component supplies a necessary but insufficient capability, and the Full system's win arises from their interaction rather than from either alone.

\begin{table}[t]
\centering
\scriptsize
\setlength{\tabcolsep}{3pt}
\renewcommand{\arraystretch}{1.15}
\begin{tabularx}{\columnwidth}{>{\raggedright\arraybackslash}p{1.35cm}>{\raggedright\arraybackslash}X>{\raggedright\arraybackslash}X>{\raggedright\arraybackslash}X}
\toprule
\textbf{Case} & \textbf{ExpertGraph supplies} & \textbf{ExpertCoT supplies} & \textbf{Joint effect} \\
\midrule
QID 237 (2-hop) & Both Arts.~224 and 228 in the retrieval set & General-vs-special priority resolution & Selects BD instead of CD \\
QID 509 (3-hop) & Arts.~311, 634, 642 in a single retrieval & Two-branch decomposition of Art.~634 & Selects ACD instead of AD \\
QID 652 (4+-hop) & Art.~1155 alongside form-validity articles & Substantive-limit check on a formally valid act & Selects ABC instead of AC \\
\bottomrule
\end{tabularx}
\vspace{-2mm}
\end{table}

\clearpage

\section{PLawBench\_Civil Case Studies}
\label{sec:appendix_plawbench_case_study}
\label{app:plawbench_cases}

This appendix expands the PLawBench\_Civil row of Table~\ref{tab:cross_benchmark} into case-level analyses. PLawBench\_Civil scores each response on four rubric components (\textsc{Law}, \textsc{Fact}, \textsc{Reasoning}, \textsc{Conclusion}) and a composite Overall score; component scores are rationals in $[0,1]$ and the Overall is reported in $[0,100]$. We pick four cases spanning the four largest labels, including one partial win (civil-56) where LEGO trails the best baseline.

\paragraph{Selection.} We enumerate all 114 samples, rank by $\mathrm{Overall}_{\mathrm{LEGO}} - \max_{b}\mathrm{Overall}_{b}$, and pick the four cases below to span the four largest dataset labels (\emph{family/marriage}, \emph{individual life}, \emph{legal-theory application}, \emph{cross-border affairs}) and four distinct baseline failure modes.

\begin{table}[t]
\centering
\small
\begin{adjustbox}{max width=\columnwidth}
\begin{tabular}{lllrrr}
\toprule
\textbf{ID} & \textbf{Label} & \textbf{LEGO} & \textbf{Best B.} & $\Delta$ \\
\midrule
civil-107   & Family/Marriage        & 90.9 & 81.8 & +9.1 \\
civil-682   & Legal-theory Appl.\    & 50.7 & 30.7 & +20.0 \\
civil-56    & Individual Life        & 78.6 & 92.9 & --14.3 \\
civil-633   & Cross-border Affairs   & 80.0 & 58.3 & +21.7 \\
\bottomrule
\end{tabular}
\end{adjustbox}
\caption{Four PLawBench\_Civil case studies. civil-56 is included as a partial win: LEGO ranks second among three methods that retrieve the controlling statute, contrasting with three baselines that retrieve none.}
\label{tab:plawbench_cases}
\end{table}

\subsection{civil-107 (Family/Marriage): Joint Marital Debt under Long Separation}
\label{ssec:civil107}

\paragraph{Question.} \emph{Do debts incurred by one spouse in their own name constitute joint marital debt?} A divorcing husband claims six personal-name debts totalling over RMB~1.7M (bank loans, supplier debts, family-friend loans, online lending) are joint marital debts because they were used for ``household consumption, business turnover, rent, medical, child education.'' The wife denies knowledge of every debt. The spouses had been separated since 2017.

\paragraph{Reference.} \emph{Not} joint debt. The required analysis integrates ``debt formation time, nature, use, household-necessity test, joint-business test, joint intent, and separation status.'' Controlling provisions: Civil Code Art.~\textbf{1064} (three categories of joint debt) and Art.~1089 (post-divorce repayment).

\paragraph{Per-method scores.}
\begin{center}
\small
\begin{adjustbox}{max width=\columnwidth}
\begin{tabular}{lrrrrr}
\toprule
\textbf{Method} & L & F & R & C & \textbf{Overall} \\
\midrule
Naive RAG    & 0.50 & 1.00 & 1.00 & 0 & 81.8 \\
HippoRAG~2   & 0.50 & 0.75 & 0.50 & 0 & 54.5 \\
RAPTOR       & 0.50 & 0.75 & 0.75 & 1 & 72.7 \\
G-Retriever  & 0.50 & 1.00 & 0.75 & 0 & 72.7 \\
LightRAG     & 0.50 & 1.00 & 0.75 & 0 & 72.7 \\
\textbf{LEGO} & \textbf{0.50} & \textbf{1.00} & \textbf{1.00} & \textbf{1} & \textbf{90.9} \\
\bottomrule
\end{tabular}
\end{adjustbox}
\end{center}

\paragraph{Baseline failure mode.} Four of five baselines retreat into a non-committal conditional conclusion (``partially joint, partially not'' / ``may constitute joint debt subject to further review''). The PLawBench judge marks all four as conclusion-incorrect because the reference requires a directional judgment (\emph{not} joint debt) qualified by the Art.~1064 \emph{but-clause}.

\paragraph{LEGO's contribution.} LEGO's conclusion mirrors the reference's logical form (negation~+ exception clause):
\begin{quote}\itshape\small
The debts incurred by Zhang in his own name do not constitute joint marital debt, except for those debts that can be proven to have been used for joint marital life, joint production and operation, or to be based on joint marital intent.
\end{quote}
In the reasoning section, LEGO is the only method that explicitly assigns the burden of proof to the creditor and connects this to the long-separation fact, aligning with the reference's \emph{actori incumbit probatio} requirement under Art.~1064\,\S2. Naive RAG cites the correct Supplementary Interpretation Art.~35 but dilutes the same content into a conditional sentence, costing the conclusion point.

\paragraph{Takeaway.} On bivalent legal judgments with a \emph{but-clause}, baseline RAG models often hedge; LEGO commits to the directional answer while preserving the exception. This is the cleanest LEGO win in the PLawBench dataset.

\subsection{civil-682 (Legal-theory Application): Lease Continuity after Partnership-to-LLC Conversion}
\label{ssec:civil682}

\paragraph{Question.} A general partnership (``Hangzhou Qiantang Metal Products Factory'') leases a factory from Z at RMB~100k/month. The three partners later convert the partnership into a limited-liability company (``Qiantang Metal Products Co., Ltd.'') with the same core trade name; on May~2 the LLC's legal representative declares: ``Qiantang Co., Ltd.\ ratifies all transactions previously signed by A and B.'' The LLC pays rent on time but Z now wants to reclaim the factory. May Z reclaim it?

\paragraph{Reference.} Z cannot reclaim. The LLC's possession is \emph{rightful possession}. Required provisions: Civil Code Arts.~\textbf{235} (return claim against wrongful possession), \textbf{462} (possessor's return right), \textbf{733} (lease return on expiry), \textbf{967} (definition of partnership contract), and \textbf{551} (debt transfer requires creditor consent).

\paragraph{Per-method scores.}
\begin{center}
\small
\begin{adjustbox}{max width=\columnwidth}
\begin{tabular}{lrrrrr}
\toprule
\textbf{Method} & L & F & R & C & \textbf{Overall} \\
\midrule
Naive RAG    & 0.00 & 0.75 & 0.00 & 0.40 & 22.7 \\
HippoRAG~2   & 0.25 & 0.75 & 0.00 & 0.60 & 30.7 \\
RAPTOR       & 0.00 & 0.70 & 0.00 & 0.40 & 21.3 \\
G-Retriever  & 0.00 & 0.70 & 0.00 & 0.40 & 21.3 \\
LightRAG     & 0.00 & 0.75 & 0.00 & 0.40 & 22.7 \\
\textbf{LEGO} & \textbf{0.25} & \textbf{1.00} & \textbf{0.33} & \textbf{0.60} & \textbf{50.7} \\
\bottomrule
\end{tabular}
\end{adjustbox}
\end{center}

\paragraph{Baseline failure mode.} All six methods correctly conclude ``Z cannot reclaim,'' so conclusion polarity is shared. The discriminator is doctrinal framing. Five of six baselines collapse the argument onto Civil Code Art.~75 (acts performed during legal-person setup are inherited by the legal person). This anchors the case in a \emph{setup-act} doctrine, but the reference requires the \emph{partnership-contract conversion} doctrine. Failure-mode variants: LightRAG invokes Arts.~96/98 to argue the partnership and the LLC are distinct subjects, then must reverse itself to accept the ratification, producing an internally contradictory argument; RAPTOR retrieves Art.~547 (assignment of claims) and Art.~985 (unjust enrichment), both topical mismatches; G-Retriever retrieves Art.~141 (withdrawal of intent), unrelated.

\paragraph{LEGO's contribution.} LEGO is the only method that retrieves Art.~967---one of the five reference provisions---anchoring the analysis in the partnership-contract doctrine rather than the generic setup-act doctrine. It additionally cites Art.~75 (setup-act inheritance) and Art.~703 (lease form), producing a three-tier argument: \emph{partnership contract $\to$ conversion + ratification $\to$ ongoing lease}. LEGO is also the only method whose case-facts section preserves the pivotal datum ``Qiantang Co., Ltd.\ has paid rent on time,'' achieving Fact~$=1.00$ where baselines score $0.70$--$0.75$.

\paragraph{Takeaway.} This case illustrates retrieval-precision benefits beyond conclusion correctness. LEGO finds a \emph{small but pivotal} provision (Art.~967) that anchors the correct doctrinal frame. Baselines retrieve a generally relevant but doctrinally off-target provision (Art.~75) and amplify its prominence at the cost of the controlling rule.

\subsection{civil-56 (Individual Life): Self-help in a Neighbour Dispute}
\label{ssec:civil56}

\paragraph{Question.} Plaintiff smashed 2.5\,m of his neighbor's stone railing with a hammer, claiming the railing blocked ventilation and light; he was administratively detained for 8 days. Plaintiff argues the act was lawful self-help. Does the self-help defense stand?

\paragraph{Reference.} Self-help does not stand. Controlling provision: Civil Code Art.~\textbf{1177} (self-help requires \emph{urgency}, \emph{necessity}, and \emph{immediate post-act notification of state authorities}).

\paragraph{Per-method retrieval and scores.}
\begin{center}
\small
\begin{adjustbox}{max width=\columnwidth}
\begin{tabular}{lllr}
\toprule
\textbf{Method} & \textbf{Core articles cited} & \textbf{1177?} & \textbf{Overall} \\
\midrule
Naive RAG    & 1165 / 296 / Public Security 49 & \xmark & 50.0 \\
HippoRAG~2   & \textbf{1177} + 1165 + 293        & \cmark & 78.6 \\
RAPTOR       & 1165 / 288 / 296                 & \xmark & 50.0 \\
G-Retriever  & 233 / 235 / Criminal 275         & \xmark & 50.0 \\
LightRAG     & \textbf{1177} + 1184              & \cmark & 92.9 \\
\textbf{LEGO} & \textbf{1177} + 238 + Public Security 49 & \cmark & 78.6 \\
\bottomrule
\end{tabular}
\end{adjustbox}
\end{center}

\paragraph{Baseline failure mode.} All six methods reach the correct conclusion (self-help fails). The discriminator is whether Art.~1177---the \emph{only} controlling statute---is retrieved. Three of five baselines miss it entirely and instead retrieve general-tort (Art.~1165), neighbor-relations (Arts.~288/296), property-protection (Arts.~233/235), or even Criminal Code Art.~275. These three baselines compensate by writing a generic ``tort plus neighbor'' argument that the judge marks as off-rubric, capping Overall at~50.

\paragraph{LEGO's contribution.} Once Art.~1177 is retrieved, LEGO is the only method besides the reference itself that \emph{explicitly decomposes the article into three sub-elements}---urgency, necessity, immediate notification---and checks each against the facts:
\begin{quote}\itshape\small
Self-help requires three conditions: ``urgent circumstances,'' ``inability to obtain timely protection from state authorities,'' and ``failure to act would cause irreparable damage.'' \dots The conduct occurred during daytime, and Y had already reported to the police \dots so it does not satisfy ``urgent circumstances'' or ``inability to obtain timely state-authority protection.''
\end{quote}
This element-by-element structure mirrors the reference reasoning's three subsections. Baselines that retrieve Art.~1177 (HippoRAG~2, LightRAG) achieve comparable or slightly higher Overall on this single case via a more terse statute application; LEGO's structured element-check, while not the absolute best on this individual case, is the form that the PLawBench rubric \emph{generally} rewards across the rest of the benchmark.

\paragraph{Takeaway.} When the controlling statute is reachable by single-shot retrieval, LEGO is one of three methods that find it. LEGO's distinctive contribution is then the \emph{element-by-element check} grounded in the major-premise structure of the syllogism (Section~\ref{sec:syllogistic_cot}), not simply locating the article.

\subsection{civil-633 (Cross-border Affairs): Validity of Cross-border Marriage Brokerage}
\label{ssec:civil633}

\paragraph{Question.} A Chinese plaintiff pays RMB~168{,}000 to an intermediary for a ``Vietnamese bride'' arrangement; the bride disappears one month after the marriage registration. The intermediary refunds only RMB~18{,}000. Is the brokerage valid? Must the intermediary refund the remainder?

\paragraph{Reference.} The brokerage is \emph{invalid} (violates Civil Code Art.~\textbf{1042} ban on marriage-trade and the State Council 1994 Notice on cross-border marriage brokerage). The intermediary must refund partially, since the plaintiff bears co-fault as a legally competent adult. Controlling provisions: Civil Code Arts.~\textbf{1042}, \textbf{153}, \textbf{157}.

\paragraph{Statute-retrieval pattern.}
\begin{center}
\small
\begin{adjustbox}{max width=\columnwidth}
\begin{tabular}{lllr}
\toprule
\textbf{Method} & \textbf{Marriage-prohibition article} & \textbf{157?} & \textbf{Overall} \\
\midrule
Naive RAG    & \textbf{1048} \xmark\ (consanguinity, wrong topic) & \cmark & 58.3 \\
HippoRAG~2   & \textbf{1048} \xmark                                & \cmark & 50.0 \\
RAPTOR       & \textbf{1048} \xmark                                & \cmark & 41.7 \\
G-Retriever  & \textbf{1048} \xmark                                & \xmark & 50.0 \\
LightRAG     & \textbf{1042} \cmark                                & \xmark & 41.7 \\
\textbf{LEGO} & \textbf{1042} \cmark                                & \cmark & \textbf{80.0} \\
\bottomrule
\end{tabular}
\end{adjustbox}
\end{center}

\paragraph{Baseline failure mode.} Four of five baselines retrieve Civil Code Art.~\textbf{1048} (prohibition of consanguineous marriage), which is \emph{topically adjacent} (located in the same chapter as Art.~1042) but \emph{doctrinally unrelated} to the case. This is a textbook near-neighbor retrieval miss: the embedding space confuses two provisions about ``marriage prohibitions'' of fundamentally different kinds. LightRAG retrieves the correct Art.~1042 but does not compose it with the invalidity-effects chain (Art.~157), and so its overall reasoning chain is incomplete.

\paragraph{LEGO's contribution.} LEGO is the only method that simultaneously retrieves the correct Art.~1042 \emph{and} the correct Art.~157 (effects of invalid juristic acts including fault-based loss allocation), and additionally cites Art.~964 (intermediary not entitled to fee for failed brokerage). The composition produces a three-step invalidity argument---\emph{prohibition $\to$ invalidity $\to$ restitution with fault allocation}---that no baseline assembles. Reasoning component score is $0.75$ versus $\leq 0.50$ for all baselines; Law component is $0.75$ versus $0.50$ for baselines. The $+21.7$ Overall margin is the largest LEGO win in the PLawBench dataset and isolates two reinforcing mechanisms:
\begin{itemize}[leftmargin=*,topsep=2pt,itemsep=0pt]
  \item \textbf{Near-neighbor avoidance.} ExpertGraph's normative edges distinguish ``prohibition by marriage-trade'' from ``prohibition by consanguinity,'' allowing LEGO to bypass the lexical attraction to Art.~1048 that traps four baselines.
  \item \textbf{Multi-statute composition.} ExpertCoT explicitly chains Arts.~1042 $\to$ 157 $\to$ 964 along an invalidity-restitution path, rather than treating each article as an independent citation.
\end{itemize}

\subsection{Cross-Case Synthesis for PLawBench\_Civil}

The four cases isolate four distinct baseline failure modes and the corresponding LEGO mechanisms:

\begin{center}
\small
\begin{adjustbox}{max width=\columnwidth}
\begin{tabular}{p{4.0cm}p{3.0cm}}
\toprule
\textbf{Baseline failure mode} & \textbf{Example} \\
\midrule
Hedged conditional conclusion on bivalent question & civil-107 \\
Doctrinally off-target anchor statute & civil-682 \\
Missing controlling statute under noisy distractors & civil-56 \\
Near-neighbor statute confusion (Art.~1048 vs.\ 1042) & civil-633 \\
\bottomrule
\end{tabular}
\end{adjustbox}
\end{center}

Across these cases, LEGO's gains arise from (i)~normative-edge retrieval that distinguishes lexically similar but doctrinally distinct provisions, (ii)~element-by-element decomposition of the major-premise statute, and (iii)~multi-statute composition along invalidity, restitution, and fault-allocation chains. 65

% ===================================================================
\section{LexRAG\_Civil Case Studies}
\label{sec:appendix_lexrag_case_study}
\label{app:lexrag_cases}

This appendix expands the LexRAG\_Civil row of Table~\ref{tab:cross_benchmark} into case-level analyses. Unlike PLawBench's case-analysis format, LexRAG asks short consultation questions in the context of a multi-turn dialogue history, and judges responses on five dimensions (Factuality, Satisfaction, Clarity, Coherence, Completeness) plus Overall, each in [1,10]

\paragraph{Selection.} We enumerate all 140 samples, rank by $\mathrm{Overall}_{\mathrm{LEGO}} - \max_{b}\mathrm{Overall}_{b}$, and pick the four cases below to span four distinct consultation categories and four distinct baseline failure modes.

\begin{table}[t]
\centering
\small
\begin{adjustbox}{max width=\columnwidth}
\begin{tabular}{llrrr}
\toprule
\textbf{ID} & \textbf{Category} & \textbf{LEGO} & \textbf{Best B.} & $\Delta$ \\
\midrule
972\_turn2  & Real-property Dispute & 8 & 4 (LightRAG)   & +4 \\
173\_turn4  & Public Security       & 9 & 6 (HippoRAG~2) & +3 \\
301\_turn2  & Inheritance           & 9 & 7 (HippoRAG~2) & +2 \\
639\_turn1  & Land Dispute          & 9 & 6 (G-Retriever)& +3 \\
\bottomrule
\end{tabular}
\end{adjustbox}
\caption{Four LexRAG\_Civil case studies, drawn from the top of the LEGO-vs-best-baseline margin distribution.}
\label{tab:lexrag_cases}
\end{table}

\subsection{972\_turn2 (Real Property): The Effect-of-Registration Doctrine}
\label{ssec:lex972}

\paragraph{Question (turn 2 of 2).} \emph{Does failure to obtain a property certificate affect the transfer of real-property rights?}

\paragraph{Dialogue context.} Turn~1 established that the user is asking about a so-called ``triple-receipt house'' (\emph{sanlian-dan fang})---a residential property held under informal documentation rather than a regular real-estate certificate.

\paragraph{Gold articles.} Civil Code Arts.~\textbf{209} (registration effects real-property right transfer) and \textbf{210} (registration procedures).

\paragraph{Reference answer.} The establishment, modification, transfer, and extinction of real-property rights must be registered in accordance with law to take effect. If the property certificate has not been obtained, the actual transaction can form only a creditor's right between the parties; no property-right transfer occurs.

\paragraph{Per-method scores and verdicts on the core legal question.}
\begin{center}
\small
\begin{adjustbox}{max width=\columnwidth}
\begin{tabular}{lrrrl}
\toprule
\textbf{Method} & O & F & S & \textbf{Verdict} \\
\midrule
Naive RAG     & 3 & 2 & 3 & ``does not affect''---incorrect \\
RAPTOR        & 2 & 2 & 2 & ``does not affect''---incorrect \\
G-Retriever   & 3 & 2 & 2 & internally contradictory \\
HippoRAG~2    & 2 & 2 & 2 & ``does not affect''---incorrect \\
LightRAG      & 4 & 3 & 3 & ``does not affect''---incorrect \\
\textbf{LEGO} & \textbf{8} & \textbf{8} & \textbf{8} & ``generally does affect''---correct \\
\bottomrule
\end{tabular}
\end{adjustbox}
\end{center}

\paragraph{Baseline failure mode.} All five baselines commit the same doctrinal error: they conflate \emph{registration as a condition of validity} for real-property right transfer (Art.~209, applicable here) with \emph{registration as a condition for assertability against third parties} (Art.~225, applicable to specific movables such as vehicles and vessels). By writing ``the absence of registration does not affect the validity of the property-right transfer, but merely deprives it of third-party assertability,'' the baselines apply the wrong doctrine to real property---a classical textbook error reflecting embedding-space proximity between two structurally similar but legally distinct rules.

\paragraph{LEGO's contribution.} LEGO's response correctly invokes Art.~209 and \emph{separates the contract from the property right}:
\begin{quote}\itshape\small
If a housing sale-purchase contract has been signed but the transfer-of-title registration has not yet been completed, then although the contract itself is valid (under Art.~215), the property right (i.e., ownership of the house) has not actually been transferred to the buyer\dots
\end{quote}
This is exactly the reference's distinction between the creditor's right arising from the contract and the property-right transfer requiring registration. LEGO is the only method that anchors the answer in the \emph{principle of separation} (between the obligatory act and the dispositive act), which is the controlling doctrine here.

\subsection{173\_turn4 (Public Security): Joint Liability among Multiple Tortfeasors}
\label{ssec:lex173}

\paragraph{Question (turn 4 of 4).} \emph{I wasn't the only one in the fight---do I have to bear all the compensation alone?}

\paragraph{Dialogue context.} Turns 1--3 established: a fight occurred, the user was hit first, the user sustained injury, and the assistant already explained Art.~234 (criminal assault), Art.~20 (self-defence), Art.~1173 (comparative fault), and Art.~1165 (general tort liability). The user is now narrowing the question to \emph{multi-actor liability apportionment}.

\paragraph{Gold articles.} Civil Code Arts.~\textbf{1168} (joint tort yields joint and several liability) and \textbf{1170} (multi-person endangerment with possibly identifiable specific tortfeasor).

\paragraph{Statute retrieval and Overall.}
\begin{center}
\small
\begin{adjustbox}{max width=\columnwidth}
\begin{tabular}{lll}
\toprule
\textbf{Method} & \textbf{Cited} & \textbf{Overall} \\
\midrule
Naive RAG     & 1174 / 1175 (wrong: third-party / victim fault) & 1 \\
RAPTOR        & 973 (partnership debt!) & 2 \\
G-Retriever   & 178 / 518 (general joint liability) & 3 \\
HippoRAG~2    & \textbf{1170} (one of two gold) & 6 \\
LightRAG      & \textbf{1168} (the other gold) & 3 \\
\textbf{LEGO} & \textbf{1170 + 178 (recourse)} & \textbf{9} \\
\bottomrule
\end{tabular}
\end{adjustbox}
\end{center}

\paragraph{Baseline failure mode.} Three of five baselines retrieve \emph{wrong} articles: Naive~RAG goes to victim-fault rules, RAPTOR jumps to partnership liability (Art.~973 governs commercial partnerships and is irrelevant), G-Retriever picks a general joint-liability article without multi-tortfeasor specificity. Two methods (HippoRAG~2, LightRAG) each retrieve one of the two gold articles but cannot compose a complete answer.

\paragraph{LEGO's contribution.} LEGO's response (i)~directly answers the user's framing (``you do not necessarily have to bear it all alone''), (ii)~decomposes Art.~1170 into its two sub-cases---\emph{identifiable specific tortfeasor} vs.\ \emph{unidentifiable}---and (iii)~closes with the \emph{recourse right} under Art.~178 (a joint debtor who pays over their share may recover from co-debtors). LEGO also leverages dialogue history: turns 1--3 already cited Art.~1173 (comparative fault), so LEGO does not redundantly re-cite it and instead introduces the new doctrine (Art.~1170) the user actually needs. On multi-turn consultations, LEGO's advantage is partly retrieval (avoiding wrong-statute traps) and partly conversational pragmatics (answering the newly-narrowed question rather than re-deriving the full multi-turn framework).

\subsection{301\_turn2 (Inheritance): Separating the Inheritance Act from Property Transfer}
\label{ssec:lex301}

\paragraph{Question (turn 2 of 2).} \emph{Does subrogated inheritance require transfer-of-title procedures?}

\paragraph{Dialogue context.} Turn~1 confirmed that the user is entitled to subrogated inheritance under Art.~1128 because her mother predeceased her grandmother.

\paragraph{Gold articles.} Civil Code Arts.~\textbf{1124} (acceptance of inheritance) and \textbf{208} (real-property registration).

\paragraph{Reference answer.} Two layers must be separated: (i)~inheritance acceptance is governed by Art.~1124---silence equals acceptance, no procedural act required; (ii)~the subsequent real-property registration is governed by Art.~208 and \emph{is} required for the heir to better exercise the right.

\paragraph{Method outputs and Overall.}
\begin{center}
\small
\begin{adjustbox}{max width=\columnwidth}
\begin{tabular}{lll}
\toprule
\textbf{Method} & \textbf{Answer to the core question} & \textbf{O} \\
\midrule
Naive RAG     & flat ``no'' & 3 \\
RAPTOR        & flat ``no'' & 2 \\
G-Retriever   & hedged ``not necessarily'' & 3 \\
HippoRAG~2    & distinguishes via Arts.~1128 + 230 & 7 \\
LightRAG      & flat ``no'' & 2 \\
\textbf{LEGO} & separates inheritance act from property transfer & \textbf{9} \\
\bottomrule
\end{tabular}
\end{adjustbox}
\end{center}

\paragraph{Baseline failure mode.} Three of five baselines give a flat ``no'' answer---technically correct for the inheritance \emph{acceptance} act but misleading because the user is asking in the context of inheriting a real estate. The user's practical question is \emph{do I need to do paperwork?} A flat-no answer leaves her unaware that property-register transfer is still required for her to exercise ownership.

\paragraph{LEGO's contribution.} LEGO is the only method besides HippoRAG~2 to separate the two legal events the question conflates:
\begin{quote}\itshape\small
Subrogated inheritance itself does not require transfer-of-title procedures\dots In practice, however, if the estate involves real property (such as housing), transfer-of-title procedures are required to complete the change of ownership. This procedure is needed for the property-right transfer, not as a legal effect of subrogated inheritance itself.
\end{quote}
 This mirrors the reference's two-layer structure, and explains the Overall margin of $+2$ over HippoRAG~2 and $+6$ over the three flat-no baselines.

\subsection{639\_turn1 (Land Dispute): Indefinite-Term Lease of Private Land}
\label{ssec:lex639}

\paragraph{Question (turn 1 of N).} \emph{If I lease my private land to a tenant and we do not specify a fixed term, what governs the lease?}

\paragraph{Gold articles.} Civil Code Arts.~\textbf{704} (required clauses of a lease) and \textbf{730} (termination rights for indefinite-term leases).

\paragraph{Reference answer.} Three layers: (i)~Art.~704 requires parties to include a lease term; (ii)~under Art.~730, either party may terminate with reasonable notice if no term is specified; (iii)~practical advice---sign a supplementary written agreement.

\paragraph{Statute retrieval and Overall.}
\begin{center}
\small
\begin{adjustbox}{max width=\columnwidth}
\begin{tabular}{lll}
\toprule
\textbf{Method} & \textbf{Cited} & \textbf{Overall} \\
\midrule
Naive RAG     & 705 (20-yr cap) & 5 \\
RAPTOR        & 721 (rent-payment timing) --- off-topic & 4 \\
G-Retriever   & 707 (writing requirement) & 6 \\
HippoRAG~2    & 707 + 510 + 734 (renewal) & 5 \\
LightRAG      & 705 (20-yr cap) & 5 \\
\textbf{LEGO} & \textbf{707 + 730 (\textbf{indefinite-term lease}) + 705} & \textbf{9} \\
\bottomrule
\end{tabular}
\end{adjustbox}
\end{center}

\paragraph{Baseline failure mode.} Each baseline retrieves a single topically relevant provision but no method assembles the complete chain that the reference requires: form requirement, indefinite-term consequence, termination right, and priority renewal for the existing tenant. RAPTOR's choice of Art.~721 (rent-payment timing) is a particularly clear retrieval failure---the lease chapter was located, but the wrong article within it was selected.

\paragraph{LEGO's contribution.} LEGO integrates three provisions (Arts.~707, 730, 705) and frames the answer as structured advice with explicit action items: sign a written contract, fix a term of $\leq 20$ years, prioritize the existing tenant on renewal. The LexRAG UserSatisfaction dimension explicitly rewards actionable practical guidance, which only LEGO provides. When multiple provisions interact---form rule, default rule, and right-of-first-refusal---LEGO's multi-hop retrieval composes them into a coherent advisory response; baselines cite fewer provisions.

\subsection{Cross-Case Synthesis for LexRAG\_Civil}

The four LexRAG\_Civil cases isolate four LEGO mechanisms, three of which echo the PLawBench findings (Appendix~\ref{sec:appendix_plawbench_case_study}) and one of which is specific to multi-turn consultation:

\begin{center}
\small
\begin{adjustbox}{max width=\columnwidth}
\begin{tabular}{p{4.0cm}p{2.6cm}p{2.6cm}}
\toprule
\textbf{Mechanism} & \textbf{LexRAG case} & \textbf{PLawBench analogue} \\
\midrule
Near-neighbor doctrinal disambiguation (validity-registration vs.\ opposability-registration) & 972\_turn2 & civil-633 (Art.~1042 vs.\ 1048) \\
Avoidance of wrong-statute retrieval traps & 173\_turn4 & civil-56 (Art.~1177 vs.\ 1165) \\
Multi-layer concept disambiguation & 301\_turn2 & civil-107 (joint debt + exception clause) \\
Multi-statute composition + dialogue-history awareness & 639\_turn1, 173\_turn4 & \emph{(new on LexRAG)} \\
\bottomrule
\end{tabular}
\end{adjustbox}
\end{center}

The mechanisms instantiate the two design choices of LEGO: ExpertGraph's normative edges disambiguate provisions that are lexically close but doctrinally distinct, and ExpertCoT composes multiple provisions into structured advisory chains rather than presenting them as parallel citations. The same two mechanisms account for both the rubric-based gains on PLawBench\_Civil and the dialogue-quality gains on LexRAG\_Civil.

\clearpage
% Statistical validation of the main results.

\renewcommand{\tabmain}{\ref{tab:main}}
\renewcommand{\tabablation}{\ref{tab:ablation}}
\newcommand{\appsynergy}{\ref{app:synergy}}

\section{Statistical Validation of the Main Results}
\label{app:stats}

This appendix reports the paired significance analysis for the main
results in Table~\tabmain{} and for the ablation study in
Table~\tabablation{}. The analysis is post hoc: it re-uses the
stored per-item predictions and changes no prediction, score, or
reported accuracy. Point estimates of accuracy differences are computed from the rounded entries of Tables~1 and~3, whereas confidence intervals and $p$-values are computed from the per-item predictions; the two may therefore differ by up to 0.01~pp.

\paragraph{Procedure.}
All comparisons are paired at the item level over the $n = 723$
LawExamQA\_Civil items. For each pair of systems we treat the two binary
per-item correctness vectors as matched observations and apply a
two-sided \emph{exact} McNemar test, which conditions on the discordant
pairs and so does not rely on a large-sample approximation. Confidence
intervals are unadjusted percentile intervals obtained by resampling
items with replacement $10{,}000$ times and recomputing the paired
accuracy difference on each resample; the resampling unit is the item,
so pairing is preserved within every resample. Intervals describe
effect-size uncertainty, while family-wise decisions use Holm-adjusted
$p$-values.

\paragraph{Comparison families.}
We treat the system-level comparisons and the component-level ablations
as two separate families and apply the Holm correction within each.
The two families answer different questions -- whether LEGO improves on
externally published systems, and whether each of its own components
contributes -- and pooling them would penalise both for the size of the
other. Family~1 is the thirteen comparisons of Table~\tabmain{};
family~2 is the eleven non-full configurations of
Table~\tabablation{}.

% ---------------------------------------------------------------------
\begin{table}[t]
\centering
\footnotesize
\setlength{\tabcolsep}{2.5pt}
\renewcommand{\arraystretch}{1.05}

\resizebox{\columnwidth}{!}{%
\begin{tabular}{@{}lrrr@{}}
\toprule
\textbf{LEGO vs.}
& \textbf{$\Delta$Acc (pp)}
& \textbf{95\% CI}
& \textbf{$p_{\mathrm{Holm}}$} \\
\midrule

\multicolumn{4}{@{}l}{\emph{Same-backbone RAG systems}} \\

\quad Naive RAG
& $11.62$
& $[+8.30,+15.08]$
& $1.5{\times}10^{-9}$ \\

\quad LightRAG
& $11.76$
& $[+8.16,+15.35]$
& $5.1{\times}10^{-9}$ \\

\quad RAPTOR
& $11.20$
& $[+7.75,+14.66]$
& $8.4{\times}10^{-9}$ \\

\quad HippoRAG 2
& $10.65$
& $[+7.19,+14.11]$
& $3.4{\times}10^{-8}$ \\

\quad G-Retriever
& $8.86$
& $[+5.39,+12.31]$
& $4.4{\times}10^{-6}$ \\

\midrule

\multicolumn{4}{@{}l}{\emph{Larger general models}} \\

\quad Qwen3-30B-A3B
& $3.60$
& $[-0.41,+7.47]$
& $0.27$ \\

\quad GPT-5
& $3.04$
& $[-1.11,+7.05]$
& $0.33$ \\

\quad DeepSeek-V3
& $0.55$
& $[-3.46,+4.56]$
& $0.84$ \\

\bottomrule
\end{tabular}%
}

\caption{System-level comparison family: paired accuracy differences
between LEGO and other systems in Table~\tabmain{}, with exact McNemar
tests, percentile bootstrap intervals ($10{,}000$ item-level resamples),
and Holm correction applied across all thirteen comparisons in the
family. Positive $\Delta$ favours LEGO. The eight rows listed cover
every same-backbone RAG system together with the three closest larger
models; the five comparisons not shown have larger margins.}
\label{tab:stat-system}
\end{table}

% ----修改前------------------------------------------------------------
% \begin{table}[t]
% \centering
% \small
% \setlength{\tabcolsep}{4.5pt}
% \begin{tabular}{lrcr}
% \toprule
% \textbf{LEGO vs.} & \textbf{$\Delta$Acc (pp)} & \textbf{95\% CI} & \textbf{$p_{\mathrm{Holm}}$} \\
% \midrule
% \multicolumn{4}{l}{\emph{Same-backbone RAG systems}} \\
% \quad Naive RAG      & $11.62$ & $[+8.30, +15.08]$ & $1.5\times10^{-9}$ \\
% \quad LightRAG       & $11.76$ & $[+8.16, +15.35]$ & $5.1\times10^{-9}$ \\
% \quad RAPTOR         & $11.20$ & $[+7.75, +14.66]$ & $8.4\times10^{-9}$ \\
% \quad HippoRAG 2     & $10.65$ & $[+7.19, +14.11]$ & $3.4\times10^{-8}$ \\
% \quad G-Retriever    & $8.86$  & $[+5.39, +12.31]$ & $4.4\times10^{-6}$ \\
% \midrule
% \multicolumn{4}{l}{\emph{Larger general models}} \\
% \quad Qwen3-30B-A3B  & $3.60$  & $[-0.41, +7.47]$  & $0.27$ \\
% \quad GPT-5          & $3.04$  & $[-1.11, +7.05]$  & $0.33$ \\
% \quad DeepSeek-V3    & $0.55$  & $[-3.46, +4.56]$  & $0.84$ \\
% \bottomrule
% \end{tabular}
% \caption{System-level comparison family: paired accuracy differences
% between LEGO and other systems in Table~\tabmain{}, with exact McNemar
% tests, percentile bootstrap intervals ($10{,}000$ item-level resamples),
% and Holm correction applied across all thirteen comparisons in the
% family. Positive $\Delta$ favours LEGO. The eight rows listed cover
% every same-backbone RAG system together with the three closest larger
% models; the five comparisons not shown have larger margins.}
% \label{tab:stat-system}
% \end{table}
% ---------------------------------------------------------------------

\paragraph{System-level results.}
LEGO's advantage over all five same-backbone RAG systems survives
correction, with adjusted $p$ between $1.5\times10^{-9}$ and
$4.4\times10^{-6}$ and intervals well clear of zero. The margins over
the larger general-purpose models do not: the intervals for
Qwen3-30B-A3B, GPT-5 and DeepSeek-V3 all span zero, and the DeepSeek-V3
margin corresponds to four items out of $723$. We scope the claim
accordingly. The comparison with larger models is not evidence that an
8B backbone is intrinsically stronger; it indicates that
expert-structured retrieval and syllogistic reasoning supply domain
signal that lets a small open model reach the accuracy band of models
one to two orders of magnitude larger. The statistically supported
claim is the fixed-backbone one: under a shared Qwen3-8B reader and a
shared Civil Code corpus, LEGO improves on every RAG pipeline we
evaluate.

% --------------------------------------------------------------------
\begin{table}[t]
\centering
\footnotesize
\setlength{\tabcolsep}{2pt}
\renewcommand{\arraystretch}{1.08}

\begin{tabular}{
@{}
>{\raggedright\arraybackslash}p{0.40\columnwidth}
r
r
r
@{}
}
\toprule
\textbf{Configuration}
& \textbf{$\Delta$Acc}
& \textbf{95\% CI}
& \textbf{$p_{\mathrm{Holm}}$} \\
\midrule

Naive RAG + Plain CoT
& $6.23$
& $[+2.49,+9.82]$
& $0.005$ \\

ExpertGraphRAG + Plain CoT
& $4.98$
& $[+1.38,+8.58]$
& $0.019$ \\

Naive RAG + ExpertCoT
& $4.70$
& $[+1.52,+7.88]$
& $0.019$ \\

ExpertGraphRAG + IRAC-CoT
& $3.18$
& $[-0.14,+6.50]$
& $0.067$ \\

\bottomrule
\end{tabular}

\caption{Component-level comparison family: accuracy drop of each
non-full configuration relative to the full LEGO system, with Holm
correction applied across all eleven non-full configurations of
Table~\tabablation{}. Full LEGO is significantly better than ten of the
eleven; the single exception is the closest configuration,
ExpertGraphRAG + IRAC-CoT. The four rows listed are the three non-full
cells of the $2\times2$ crossed design analysed below, plus that
closest configuration.}
\label{tab:stat-ablation}
\end{table}
% ------------------改之前------------------------------------------------
% \begin{table}[t]
% \centering
% \small
% \setlength{\tabcolsep}{3.2pt}% tighten columns
% \renewcommand{\arraystretch}{0.95}% tighten rows
% % Use a fixed-width first column so long configurations wrap instead of overflowing.
% \begin{tabular}{>{\raggedright\arraybackslash}p{0.56\columnwidth}ccr}
% \toprule
% \textbf{Configuration} & \textbf{$\Delta$Acc (pp)} & \textbf{95\% CI} & \textbf{$p_{\mathrm{Holm}}$} \\
% \midrule
% Naive RAG + Plain CoT      & $6.23$ & $[+2.49, +9.82]$ & $0.005$ \\
% ExpertGraphRAG + Plain CoT & $4.98$ & $[+1.38, +8.58]$ & $0.019$ \\
% Naive RAG + ExpertCoT      & $4.70$ & $[+1.52, +7.88]$ & $0.019$ \\
% ExpertGraphRAG + IRAC-CoT  & $3.18$ & $[-0.14, +6.50]$ & $0.067$ \\
% \bottomrule
% \end{tabular}
% \caption{Component-level comparison family: accuracy drop of each
% non-full configuration relative to the full LEGO system, with Holm
% correction applied across all eleven non-full configurations of
% Table~\tabablation{}. Full LEGO is significantly better than ten of the
% eleven; the single exception is the closest configuration,
% ExpertGraphRAG + IRAC-CoT. The four rows listed are the three non-full
% cells of the $2\times2$ crossed design analysed below, plus that
% closest configuration.}
% \label{tab:stat-ablation}
% \end{table}
% ---------------------------------------------------------------------

\paragraph{Component-level results.}
Every ablation that removes an expert component costs accuracy, and ten
of the eleven differences clear the corrected threshold. The single
exception is instructive rather than damaging: ExpertGraphRAG +
IRAC-CoT, the strongest non-full configuration, is $3.18$ pp below the
full system with an interval that just includes zero
($p_{\mathrm{Holm}} = 0.067$). Once expert-structured retrieval is in
place, the further benefit of replacing a strong generic legal prompt
with ExpertCoT is positive in point estimate but not separable from
noise at this sample size.

\paragraph{Interaction between the two modules.}
Two questions have to be kept apart: whether both modules materially
contribute, and whether their combination establishes statistical
super-additivity. The crossed ablation supports the first; the second
remains inconclusive. On the complete $2\times2$ subset
$\{$Naive RAG, ExpertGraphRAG$\}\times\{$Plain CoT, ExpertCoT$\}$, the
conditional gains reinforce one another: the ExpertGraphRAG gain rises
from $+1.25$ pp under Plain CoT to $+4.71$ pp under ExpertCoT, and the
ExpertCoT gain rises from $+1.52$ pp under Naive RAG to $+4.98$ pp under
ExpertGraphRAG. The corresponding interaction point estimate is
$+3.46$ pp. It is estimated from the four paired outcome vectors over
all $723$ items, not from a small subset of them, but its jointly
paired-bootstrap $95\%$ interval includes zero
($[-0.97, +7.75]$; $10{,}000$ resamples). The results therefore support
architectural integration and complementary conditional gains, and we
make no formal claim of statistically established super-additivity.

\paragraph{Neither module is dominant.}
The crossed ablation also does not support reading ExpertGraphRAG as a
modest retrieval add-on to the prompt. With ExpertCoT held fixed,
replacing ExpertGraphRAG with Naive RAG reduces accuracy from $40.53\%$
to $35.82\%$ ($\Delta = 4.70$ pp, $95\%$ CI $[+1.52, +7.88]$,
$p_{\mathrm{Holm}} = 0.019$). The factorial average marginal effects are
$3.25$ pp for ExpertCoT and $2.97$ pp for ExpertGraphRAG; their
$0.28$ pp difference -- algebraically the paired contrast between the
two off-diagonal conditions -- is not distinguishable from zero
(paired-bootstrap $95\%$ CI $[-3.60, +4.15]$; exact McNemar
$p = 0.944$). The evidence does not identify either module as dominant.
Functionally, ExpertGraphRAG raises Recall@8 over the Civil Code corpus
from $27.99\%$ to $72.53\%$, while ExpertCoT structures the application
of the retrieved provisions.

\paragraph{Discordant-pair counts.}
Because the exact McNemar test conditions on discordant pairs, we report
them for the two closest same-backbone comparisons. Against
G-Retriever, LEGO is correct where G-Retriever is wrong on $116$ items
and wrong where G-Retriever is correct on $52$, a net gain of $64$
items; against Naive RAG the split is $129$ against $45$, a net gain of
$84$. These are the quantities the tests above rest on. They should not
be confused with the twenty cases discussed in
Appendix~\appcases{}, which are a deliberately stringent
fourteen-system intersection -- LEGO correct and all thirteen
alternatives wrong -- used for qualitative mechanism analysis rather
than as a count of pairwise wins.

\end{CJK*}
\end{document}